\documentclass[10pt,twocolumn,letterpaper]{article}

\usepackage{cvpr}

\definecolor{cvprblue}{rgb}{0.21,0.49,0.74}
\usepackage[pagebackref,breaklinks,colorlinks,citecolor=cvprblue]{hyperref}
\usepackage{newtxtext}

\usepackage{newtxmath}
\usepackage{orcidlink}
\usepackage[dvipsnames]{xcolor}

\title{PAWS: A Parametric Model of Canine Pose and Shape}

\author{
Theo Wieland\textsuperscript{1}\,\orcidlink{0009-0009-2808-5185} \quad
Zala \v{Z}gank\textsuperscript{2}\,\orcidlink{0009-0003-4315-865X} \quad
Silvia Zuffi\textsuperscript{3}\,\orcidlink{0000-0003-1358-0828} \quad \\
Ylva Mellbin\textsuperscript{2}\,\orcidlink{0009-0006-8801-343X} \quad
Elin Hernlund\textsuperscript{2}\,\orcidlink{0000-0002-5769-3958} \quad
Hedvig Kjellstr{\"o}m\textsuperscript{1}\,\orcidlink{0000-0002-5750-9655} \\[0.5em]
\textsuperscript{1}KTH, Sweden \quad
\textsuperscript{2}SLU, Sweden \quad
\textsuperscript{3}IMATI-CNR, Italy
}

\begin{document}

\maketitle
\begin{abstract}
We introduce PAWS, a parametric model of canine shape and pose, learned from 3D scans of real dogs.
Capturing the full diversity of dogs is challenging, as different breeds vary in size, proportion, and body shape.
To capture this range, we scanned 31 dogs across multiple breeds and sizes performing a range of natural movements.
From this dataset, we build a parametric model that combines an anatomically faithful skeleton with a linear shape space learned from the 3D scans.
Unlike previous models, our model is learned from real dogs in varying poses, allowing for realistic modelling of canine pose and shape.

We demonstrate that building on PAWS provides a robust framework for downstream tasks, such as surface reconstruction and image regression.
In particular, we show that a limited number of anatomically valid degrees of freedom is sufficient to accurately estimate dog pose and shape.
\end{abstract}
\section{Introduction}
\label{section:introduction}

Dogs are among the most diverse and well-studied domestic animals, with hundreds of recognized breeds\footnote{https://www.fci.be/en/Presentation-of-our-organisation-4.html} spanning an incredible range of appearance and behaviour.
Beyond companionship, they serve in working and service roles, from hunting and herding to search and rescue, protection, and olfactory detection.
Recognizing the range of human-dog interactions motivates a better understanding of their anatomy, locomotion, behaviour and communicative cues.

Despite this importance, building a realistic, anatomically plausible 3D model of dogs remains a challenge.
Existing models often fail to capture relevant regions, such as the ears, with desired realism, and limit realistic deformations, while other joints (wrist and elbow joints) are often modelled with excessive degrees of freedom, allowing for unnatural poses. 
Addressing these shortcomings requires balancing anatomical accuracy with controlled flexibility, constraining the model to abide by anatomical constraints while retaining sufficient expressiveness to capture relevant properties.

In this work, we present an approach for modelling the articulated 3D surface of dogs using surface scans of real dogs.
Using a multi-camera scanner, we captured 31 dogs spanning different breeds and body shapes, each performing a variety of natural movements.
We focused on dogs with short to medium fur length to capture body shape directly, avoiding interference from long coats.
Finally, we show that our model captures the relevant degrees of freedom, enabling a regression pipeline that estimates model parameters directly from images.

In summary, our key contributions are:
\begin{enumerate}
\item High-quality 4D surface scans of real dogs in static and dynamic poses.
\item An optimization-based pipeline for registering a template to 3D scans.
\item A realistic 3D articulated dog model learned from real scans, combining anatomical accuracy with expressive pose and shape modelling.
\item A dataset of real dog poses from fittings to the scans.
\end{enumerate}

\section{Related Work}
\label{section:related_work}
In this section, we review relevant literature, including studies on other animals beyond dogs.
A variety of dog models have been proposed for tasks such as 3D shape and pose reconstruction, animation, or biomechanical analysis.
Similarly, multiple regression architectures have been developed that regress dog-specific landmarks or model parameters from images.

Regression-based architectures are typically grouped into model-based and model-free approaches.
Model-based architectures leverage prior knowledge and biases from an underlying model to guide and constrain the regression, while model-free approaches rely solely on data-driven techniques without imposing prior assumptions.
The motivation behind incorporating suitable model priors is to improve regression results, particularly when limbs or parts of the recorded dog are occluded.

\subsection{Parametric Models}
\label{subsection:parametric_models}
A range of parametric and articulated models \cite{SMAL, BARC, BITE, SMAL+, hSMAL, VAREN} have been proposed for dogs, horses, and quadrupeds in general.
These models share a common design approach, combining a linear PCA-based shape space with pose parameters that articulate a template mesh.

Zuffi et al.~\cite{SMAL} introduced SMAL, the first articulated linear model of quadrupeds.
The model was learned from scanned toys, augmented with a kinematic tree that allowed pose articulation.
Despite its pioneering contributions, SMAL has several limitations: its shape space was learned solely from scans of toys, and its pose parameters did not include ear articulation. 

Focusing on dogs, R\"uegg et al.~\cite{BARC, BITE} introduced BARC and BITE, two sequential improvements on SMAL.
BARC used the multi-species SMAL model, and focused on exploiting side information, namely the breed, to learn better latent spaces for shape estimation, while BITE introduced a dog-only model learned from toys, with articulated ears.
In BITE, the authors also introduced a novel architecture to estimate the model parameters directly from images of dogs by leveraging contact with the ground. 
Despite these advances, some limitations remain: the joint design largely follows SMAL, modelling all joints as ball joints with over-parameterization in certain areas.
Moreover, the shape space was derived only from rigid toys that do not undergo realistic pose deformations.

Lastly, Zuffi and Black~\cite{SMAL+} introduced $\text{SMAL}^+$, an improvement on the original SMAL model.
 $\text{SMAL}^+$ follows the same modelling approach but incorporates an extensive shape space, combining templates from previous models such as SMAL, D-SMAL, and hSMAL~\cite{SMAL, BITE, hSMAL}. 
In total, the model's shape space comprises 145 templates from different quadrupeds.

Our work extends these approaches with a more diverse and realistic training set and an anatomically correct base model from the DeepMind Control Suite~\cite{DeepMindControlSuite}: a realistic model of a Pharaoh Dog created by leo3Dmodels\footnote{https://www.turbosquid.com/Search/Artists/leo3Dmodels}.
The model features a template surface mesh, an anatomical skeleton with defined joints, and 3D muscle and bone geometries. Building on this, La Barbera et al.~\cite{DeepMind_MuscleDogModel} developed a musculoskeletal model that leverages these 3D muscle geometries.

Beyond dogs, Li et al.~\cite{hSMAL} introduced hSMAL, a model inspired by SMAL~\cite{SMAL} and tailored for horses. 
Additionally, Zuffi et al.~\cite{VAREN} proposed VAREN, a highly realistic articulated model of horse shape and pose learned from scans of real horses.
Unlike previous models that use a linear function to model pose-dependent changes, the authors proposed to learn a non-linear function that predicts pose-dependent deformations conditioned on muscle activations.

Our model is mainly focused on accurate articulated pose and shape estimation. To address different applications and scientific questions, Stark et al.~\cite{OpenSimDogModel} developed a 3D musculoskeletal model of a dog. 
Unlike BARC and BITE, their model focuses on muscle dynamics rather than variations in body shape. 
However, the model is highly anatomically accurate, with a skeleton extracted from a real dog using a computed tomography (CT) scan.

Finally, multiple regression architectures have been proposed that regress parametric model parameters from images.
Recent state-of-the-art methods include AniMer~\cite{AniMer} by Lyu et al.\ and GenZoo~\cite{GenZoo} by Niewiadomski et al., which predict SMAL parameters, and AniMer+~\cite{AniMerPlus} by An et al., which predicts both SMAL and AVES parameters.

\subsection{Model-free Approaches}
Model-free approaches estimate animal pose and shape directly from images without relying on a predefined parametric model.
These approaches have gained attention recently due to their flexibility and strong generalization to body types not seen during training.

Sinha et al.~\cite{CommonPets3D} proposed Tracker-NeRF, a NeRF-based framework that constructs animatable 4D representations of cats and dogs from only a small number of given reference images by inferring 3D trajectories of points across time.
In addition, they curated a large-scale crowd-sourced dataset of around 4200 dog and cat videos.
Yao et al.~\cite{LASSIE, Hi-LASSIE} estimate 3D pose and shape of animals given only a few images of an animal species, leveraging a proposed part consistency of simpler 3D parts that together build the whole animal.
Moreover, multiple works by Shooter et al.~\cite{Shooter_SyDog, Shooter_3DDogs} leverage different synthetic datasets to estimate dog-related landmarks from both images and videos.
Finally, Kaye et al.~\cite{dual-point-maps} propose dual point maps, a data representation linking image pixels to 3D coordinates and back to the object vertices, improving 3D pose and shape reconstruction. 

\section{Method}
\label{section:method}

\subsection{Model Formulation}
\label{subsection:model_formulation}
We build our model on the same framework introduced in SMPL~\cite{SMPL} and SMAL~\cite{SMAL}.
Similar to prior models, our model employs Linear Blend Skinning (LBS) augmented with additive shape- and pose-dependent template offsets, parameterized by shape coefficients and pose parameters, respectively.
The model is defined by several learned parameters: a template mesh $\bar{T} \in \mathbb{R}^{15981 \by 3}$ representing the surface of the model in its rest pose, comprising $15981$ vertices and $31958$ faces; the skinning weights $\mathcal{W} \in \mathbb{R}^{15981 \by K}$; a joint regressor $\mathcal{J} \in \mathbb{R}^{K \by 15981}$ that predicts the positions of the $K=62$ joints in the model's kinematic tree; and two functions, $B_S$ and $B_P$, that define the shape- and pose-dependent blend deformations.
Formally, we define our model as a function $M_{\texttt{PAWS}}$, as shown in \cref{equation:paws_model_equation}.
\begin{align}
    \label{equation:paws_model_equation}
    M_{\texttt{PAWS}}(\theta, \beta) &= \text{LBS}(T_P(\theta, \beta), J(\beta), \theta, \mathcal{W}) \\
    T_P(\theta, \beta) &= \bar{T} + B_S(\beta) + B_P(\theta)
\end{align}

\begin{figure}
    \centering
    
    \includegraphics[width=0.25\linewidth]{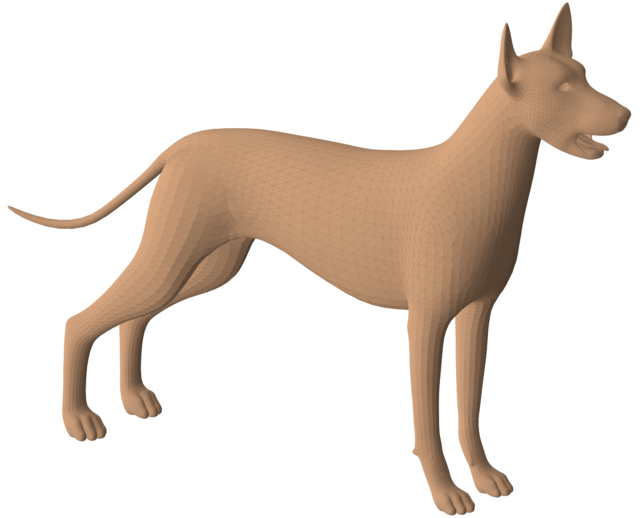}%
    \includegraphics[width=0.25\linewidth]{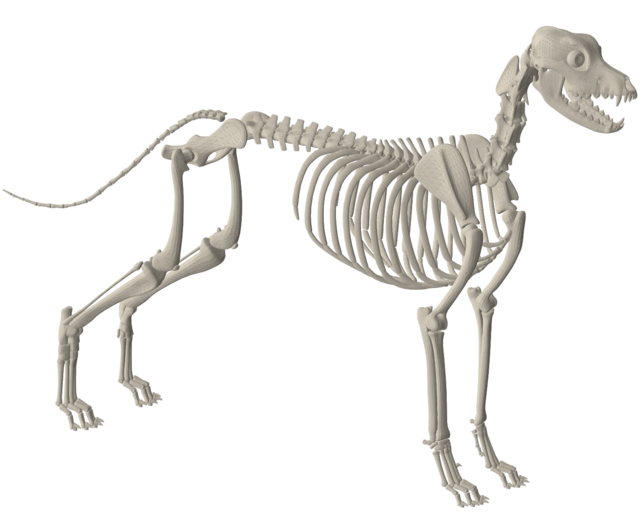}%
    \includegraphics[width=0.25\linewidth]{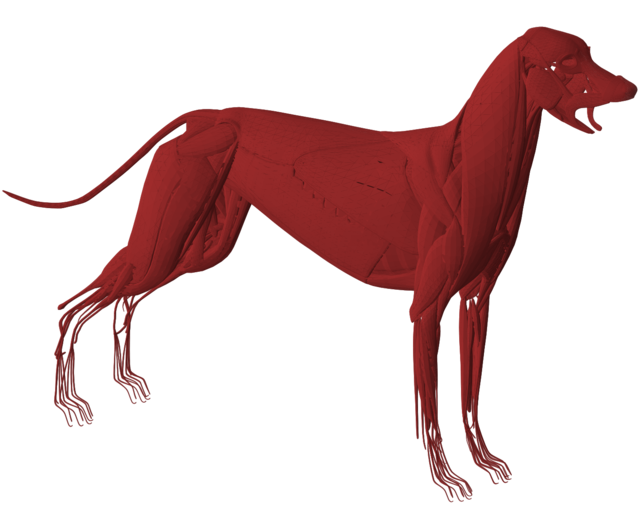}%
    \includegraphics[width=0.25\linewidth]{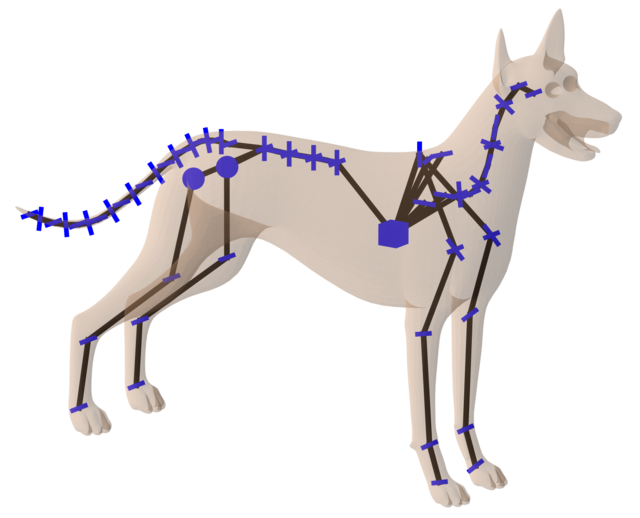}
    \caption{Artist-made Pharaoh Dog model from the \texttt{DeepMind Control Suite} (left to right): skin template, bone and muscle geometries, and kinematic tree. Joint types: free (blue boxes), ball (blue spheres), and hinge (rotation axes).}
    \label{figure:dm_control_original_model_definition}
\end{figure}

We initialize the model parameters using an artist-created Pharaoh Dog model, originally published by Tassa et al.~\cite{DeepMindControlSuite} as part of the DeepMind Control Suite.
In its original form, the model consists of a skin surface, bone and muscle geometries, and a kinematic tree with precise joint locations and joints with limited degrees of freedom.
We modify the model in two ways.
(i) We add ear articulation, representing each ear with three connected rigid bones.
(ii) We modify the template to match the geometry captured by our scanner.
Given that the scans do not contain the interior of the mouth or the eye sockets, we remove both.
Specifically, we remove the tongue and close the eye cavities, leaving a watertight surface mesh consistent with the observed data.
Moreover, we initialize the skinning weights using Blender's automatic weighting, followed by manual refinement.
Subsequently, we initialize the joint regressor to regress joint locations from a subset of the surface vertices, selected around each joint using the skinning weights.
Lastly, we initialize the shape space by transferring the shape space from R\"uegg et al.~\cite{BITE}.
To do so, we manually define a set of per-vertex correspondences between the BITE~\cite{BITE} template mesh and our template.
From these coarse correspondences, we follow the approach as described by Sumner and Popovi\'c~\cite{sumner2004deformation} to establish dense correspondences and transfer the shape space.
Finally, following Zuffi et al.~\cite{VAREN}, we augment the shape space with synthetic dimensions acting as per-part scaling controls.
We spread each scaling deformation across the surface using the skinning weights, ensuring smooth blending across neighbouring vertices.
In total, we define scaling controls for the neck, spine, tail base, tail, ears, face, paws, and front and back legs, resulting in a total of 17 additional parameters.
We refer to \cref{figure:dm_control_original_model_definition,figure:PAWS_model_components} for an overview of the model definition and initialization.

\subsection{Dataset}
\label{subsection:dog_body_surface_scans}
A convenience sample of 31 dogs of various breeds (11 Labradors, 4 Jack Russells, 2 Belgian Shepherds, 2 Dalmatians, 1 Yorkshire Terrier, 1 Whippet, 1 Vizsla, 1 English Setter, 1 Kromfohrl\"ander, 1 Husky, 1 Lagotto Romagnolo, 2 German Shepherds, 1 Golden Retriever, 1 Flat-coated Retriever, and 1 Great Dane) were recruited based on fur type (short to medium length) and ear position (erect and pendulous). 
The dogs' heights (measured from the ground to the withers) ranged from \SI{30}{\centi\metre} to \SI{81}{\centi\metre}, and their weights ranged from \SI{7.5}{\kilogram} to \SI{62}{\kilogram}.
Owners gave their written informed consent for their dogs' participation in the study.
During the recording sessions, owners were instructed to have their dogs perform various activities, including sitting, standing, lying down, walking, and running. 
Throughout all recording sessions, the dogs were allowed to move freely within the scanner's observable area, either on or off a leash, depending on the owner's preference.

The scanner used for this study is capable of capturing dynamic 4D scenes as a sequence of triangular meshes without texture.
The scanner is made up of a camera rig equipped with 16 multi-camera units, each containing two cameras, along with 22 speckle projectors mounted on the rig.
The scanner was initially set up and designed to record horses.
In total, we recorded 31 individual dogs, with 31192 total frames, accounting for roughly 52 minutes of dog observations.
From these, we manually selected 13683 frames during which each recorded dog was clearly visible, excluding segments where the current dog was outside the scanner's field of view.
All selected frames originate from 166 consecutive sequences of uninterrupted dog observations.
Finally, we manually annotated the first and last frame of all 166 sequences with 42 keypoints each.
In addition, we annotated a reduced set of keypoints at regular intervals throughout each sequence, to capture changes in the dog's pose.
In total, across the 166 sequences, we annotated 934 frames, which we use to learn our parametric model.
We refer to the appendix for an overview of the recorded dogs and additional visualizations.

\subsection{Registration}
\label{subsection:registration}
We jointly optimize the model-defining parameters and align the model to the recorded scans in a registration pipeline, commonly referred to as coregistration~\cite{coregistration}.
The registration model is initialized as described in \cref{subsection:model_formulation}.

\textbf{Subject Initialization}:
\label{subsection:subject_initialization}
We initiate the registration pipeline by learning subject-specific shape identities, one for each recorded dog.
For each identity, we register the initialized model to a subset of scans (two selected scans each, where the dog is close to the rest pose).
During initialization, we jointly optimize per-subject shape and per-frame pose parameters.
We use manual keypoint annotations to initialize the model's root translation, rotation, and isometric scale via shared-scale Procrustes alignment.
Finally, from this rough alignment, we minimize the energy function in \cref{equation:registration_subject_initialization}, which consists of a data term and several regularization terms.
We minimize with respect to the pose $\theta \in \mathbb{R}^{K \times 3 \times 3}$, shape $\beta \in \mathbb{R}^{78 + 17}$, and isometric scale $\kappa$.
\begin{align}
    \label{equation:registration_subject_initialization}
    E_\text{init}(\theta, \beta, \kappa) &= E_{\text{data}} + \mathcal{L}_{\text{pose-reg}} + \mathcal{L}_{\text{shape-reg}} \\
    \label{equation:registration_data_term}
    E_\text{data} &= \lambda_{\text{kp}} \mathcal{L}_{\text{kp}} + \lambda_{\text{s2m}} \mathcal{L}_{\text{s2m}} + \lambda_{\text{m2s}} \mathcal{L}_{\text{m2s}} \\
    \label{equation:registration_scan_to_model}
    \mathcal{L}_{\text{s2m}} &= \smashoperator{\sum_{v_s \in \mathcal{S}}} \rho \bigl( \min\nolimits_{v_m \in M_{\text{PAWS}}} \| v_s - v_m \|_2^2 \bigr) \\
    \label{equation:registration_model_to_scan}
    \mathcal{L}_{\text{m2s}} &= \smashoperator{\sum_{v_m \in M_{\text{PAWS}}}} \rho \bigl( \min\nolimits_{v_s \in \mathcal{S}} \| v_s - v_m \|_2^2 \bigr)
\end{align}
In the data term (\cref{equation:registration_data_term}), the keypoint loss $\mathcal{L}_\text{kp}$ measures the mean squared Euclidean distance between annotated keypoints and corresponding model keypoints, masked by keypoint visibility.
Here, visibility is determined by the annotations, as some keypoints may be occluded due to self-occlusions or noise during the recordings.
The scan-to-model distance $\mathcal{L}_{\text{s2m}}$ (\cref{equation:registration_scan_to_model}) measures, for each sampled vertex on the scan surface, the distance to the closest point on the model surface.
Analogously, the model-to-scan term $\mathcal{L}_{\text{m2s}}$ (\cref{equation:registration_model_to_scan}) measures, for each sampled point on the model surface, the distance to the closest point on the scan surface.
The pose regularizer $\mathcal{L}_\text{pose-reg}$ combines a rest-pose prior, a joint-limit penalty, and a manifold term.
The rest-pose prior penalizes the Frobenius distance of each joint rotation from the identity. 
The joint-limit penalty is a barrier that activates when a hinge angle leaves its predefined range.
The manifold term keeps the over-parameterized rotation parameters on their manifolds.
Additionally, we regularize the shape parameters $\beta$ to be close to zero via the shape regularizer $\mathcal{L}_\text{shape-reg}$.
The isometric scale $\kappa$ remains unregularized.
\begin{align}
    \mathcal{L}_{\text{pose-reg}} &= \lambda_\theta \|\theta - I\|_F + \lambda_\text{lim} \mathcal{L}_\text{lim} + \lambda_\text{man} \mathcal{L}_\text{man} \\
    \mathcal{L}_{\text{shape-reg}} &= \lambda_\beta \|\beta\|^2_2
\end{align}

\textbf{Pose and Shape Parameters}:
\label{subsection:pose_and_shape_parameters}
Following the subject initialization, we alternate between optimizing pose and shape parameters, respectively in \cref{equation:registration_individual_pose} and \cref{equation:registration_subject_shape} in a coordinate descent optimization scheme, gradually lowering the scale of the Geman-McClure robustifier over multiple iterations. 
\begin{align}
    \label{equation:registration_individual_pose}
    E_\text{pose} &= E_{\text{data}} + \mathcal{L}_{\text{pose-reg}} \\
    \label{equation:registration_subject_shape}
    E_\text{shape} &= E_{\text{data}} + \mathcal{L}_{\text{shape-reg}}
\end{align}

\textbf{Template Parameters}:
\label{subsection:template_parameters}
As a final step during the per-subject shape initialization, we go beyond the parametric shape space to capture finer surface details.
We introduce a set of per-vertex displacements $\Delta V^d$, which we optimize directly.
Optimizing all displacements freely would however allow the registration to overfit to scan noise and introduce one-sided artifacts that break the model's sagittal symmetry.
To prevent this, we constrain the per-subject templates to remain symmetric across the sagittal plane.
We achieve this by optimizing a reduced set of coupled proxy parameters $\Delta V^d_\text{opt} \in \mathbb{R}^{315 \times 2 + 7833 \times 3}$ instead of the full displacements.
The 315 on-plane vertices are restricted to in-plane motion (2 DoF) and the 7833 vertices on one side can move freely (3 DoF).
We recover the full displacements by mirroring displacements from one side to the other.
During this step, optimization is done only over the per-vertex displacements $\Delta V^d$.
As before, we alternate between pose parameter and template parameter optimization.
\begin{align}
    E_\text{template}(\Delta V^d) &= E_{\text{data}} + \mathcal{L}_{\text{template-reg}} \\
    \mathcal{L}_{\text{template-reg}} &= \lambda_\text{ARAP} \mathcal{L}_\text{ARAP} + \lambda_{\text{normal}} \mathcal{L}_{\text{normal}} \\
    & + \lambda_{\text{lap}} \mathcal{L}_{\text{lap}} + \lambda_{\Delta V^d} \mathcal{L}_{\text{aniso}}
\end{align}
To keep the displacements well-behaved, we regularize the displacements with a combination of ARAP~\cite{arap}, normal consistency, Laplacian smoothing, and anisotropic regularization terms.
The anisotropic regularization term penalizes tangential vertex sliding while allowing vertices to move along their template normals.
During template parameter optimization, we remove the keypoint loss from the data term to prevent overfitting to individual vertices.

\textbf{Subject Refinement}:
\label{subsection:subject_refinement}
After initializing subject-specific templates, we repeat the pose and shape optimization steps as described above (\cref{subsection:subject_initialization}) for all annotated frames.
Unlike before, we now initialize each pose from the corresponding subject's template rather than the common template.
During this step, we keep the shape and scale parameters fixed and alternate between optimizing the pose parameters and the per-vertex template offsets.

\textbf{Model Parameters}:
\label{subsection:model_parameters}
We optimize the skinning weights $\mathcal{W} \in \mathbb{R}^{15981 \by 62}$, joint regressor $\mathcal{J} \in \mathbb{R}^{62 \by 15981}$, and blend pose functions $B_P \in \mathbb{R}^{15981 \by 3 \by 549}$, whose feature dimension $549 = 61 \by 9$ stacks the $3 \by 3$ rotation matrices of the $61$ non-root joints.
As with the template offsets (\cref{subsection:template_parameters}), we optimize a set of proxy parameters that we project to the full parameters to enforce sagittal symmetry.
Let $\pi(\cdot)$ map each template vertex $v$ and each joint $j$ to its symmetric counterpart across the sagittal plane, with on-plane elements mapped to themselves.

For the skinning weights we enforce $\mathcal{W}_{v,j} = \mathcal{W}_{\pi(v),\pi(j)}$.
For each off-plane joint we optimize the full set of weights and recover its mirror-side counterpart by projection, and for each on-plane joint (where $\pi(j)=j$) we optimize one side only and mirror it.
This gives $\mathcal{W}_{\text{opt}} \in \mathbb{R}^{d_{\mathcal{W}}}$, with $d_{\mathcal{W}} = 12 \by 15981 + 38 \by (315 + 7833)$, where 12 and 38 are the numbers of off-plane (per side) and on-plane joints, and 15981, 315, and 7833 the total, on-plane, and per-side off-plane vertex counts.
The joint regressor is parameterized identically, giving $\mathcal{J}_\text{opt} \in \mathbb{R}^{d_{\mathcal{J}}}$ with $d_{\mathcal{J}} = d_{\mathcal{W}}$.

Lastly, we introduce a set of proxy parameters for the blend pose functions $B_P$, which linearly map pose features to per-vertex displacements.
Let $M=\mathrm{diag}(s_0, s_1, s_2)$ be the reflection across the sagittal plane, with $s_a=-1$ for the orthogonal axis and $s_a=+1$ otherwise. 
Applying $M$, a rotation transforms as $R \mapsto MRM$, and a displacement as $d \mapsto Md$.
Requiring that a mirrored pose results in mirrored per-vertex displacements gives ${B_P}_{v, i, (j, p, q)} = s_i s_p s_q {B_P}_{\pi(v), i, (\pi(j), p, q)}$, where $v$ indexes vertices, $i$ vertex coordinates, $j$ joints, $(p, q)$ the feature entries of joint $j$.
Each entry is thus tied to its mirrored counterpart with the factor $s_i s_p s_q \in \{\pm 1\}$.
For on-plane vertices and on-plane joints ($\pi(v)=v$, $\pi(j)=j$) the constraint simplifies to ${B_P} = s_i s_p s_q {B_P}$, forcing every entry with $s_i s_p s_q = -1$ to zero.
Intuitively, an entry is kept when an even number of its three indices  $(i, p, q)$ point along the mirrored axis.
In total, this holds for $14$ of the $27$ entries.
There are $2^3=8$ entries with no mirrored index, given that each of the three slots $i, p, q$ independently picks one of the $2$ non-mirrored axes, and $\tbinom{3}{2} \times 2 = 6$ entries with exactly two mirrored indices, where $\tbinom{3}{2}$ selects which two of the three slots are mirrored and the factor $2$ lets the remaining slot take either non-mirrored axis.
The remaining $27 - 14 = 13$ entries are forced to zero.
This yields three cases.
(i) For $7833$ off-plane vertices (on one side) we keep the full feature dimension.
(ii) For the 315 on-plane vertices we keep the full dimension of the 12 off-plane joints.
(iii) For the 315 on-plane vertices paired with the 37 on-plane joints (excluding the root), only 14 of every 27 block entries remain ($315 \times 518$, with $518 = 37 \times 14$).
We optimize ${B_P}_{\text{opt}} \in \mathbb{R}^{d_{B_P}}$, with $d_{B_P} = 7833 \by 3 \times 549 + 315 \by 3 \by 108 + 315 \by 518$.

\begin{figure}[H]
    \centering
    \resizebox{\linewidth}{!}{%
        \includegraphics[width=0.2\linewidth]{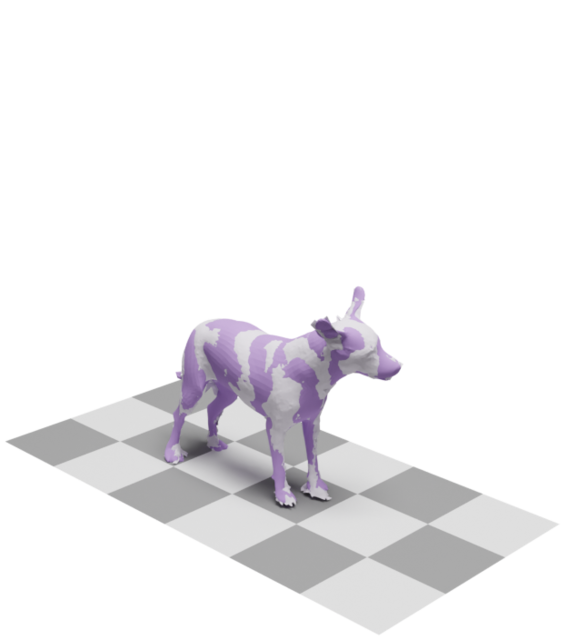}%
        \hspace{-5mm}
        \includegraphics[width=0.2\linewidth]{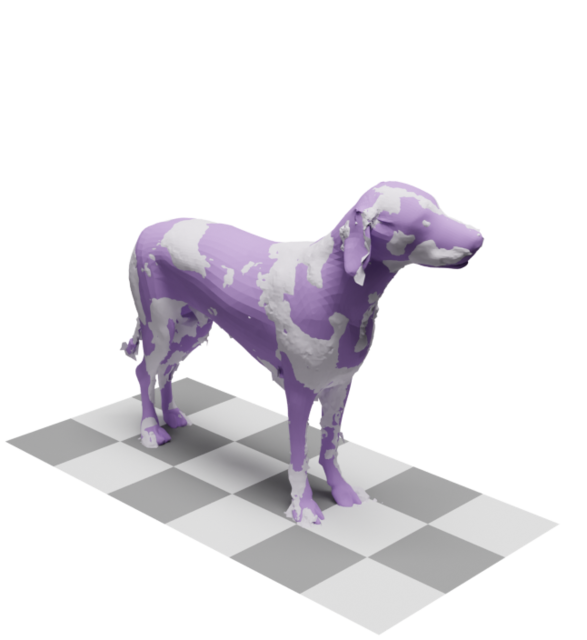}%
        \hspace{-5mm}
        \includegraphics[width=0.2\linewidth]{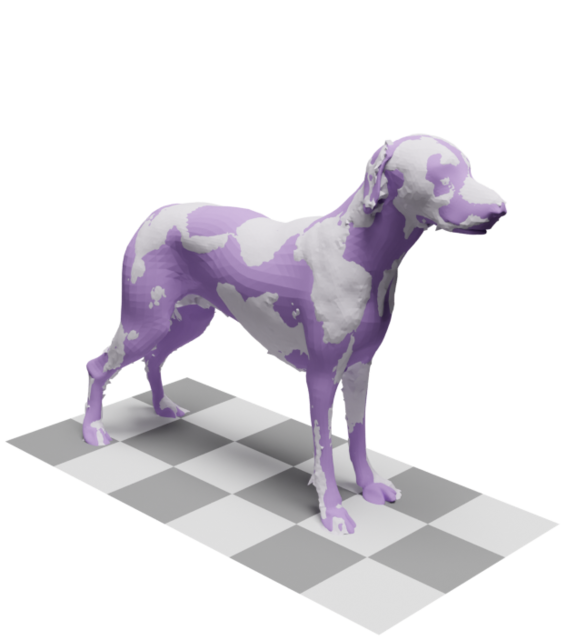}%
        \hspace{-5mm}
        \includegraphics[width=0.2\linewidth]{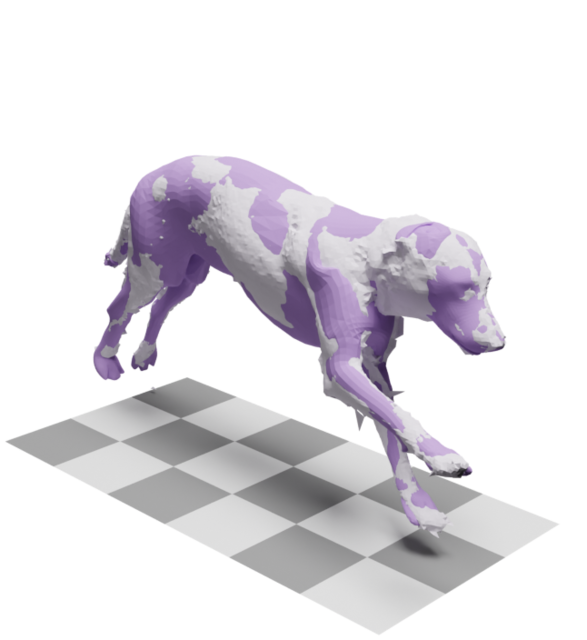}%
        \hspace{-5mm}
        \includegraphics[width=0.2\linewidth]{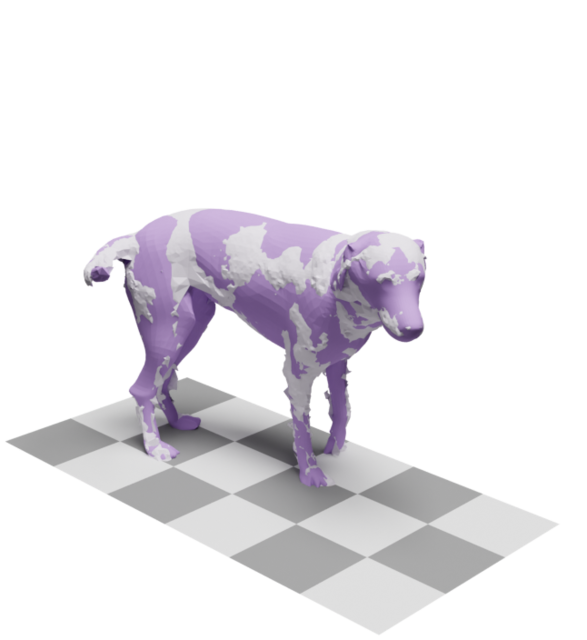}%
    }
    
    \vspace{-12mm}
    \resizebox{\linewidth}{!}{%
        \includegraphics[width=0.2\linewidth]{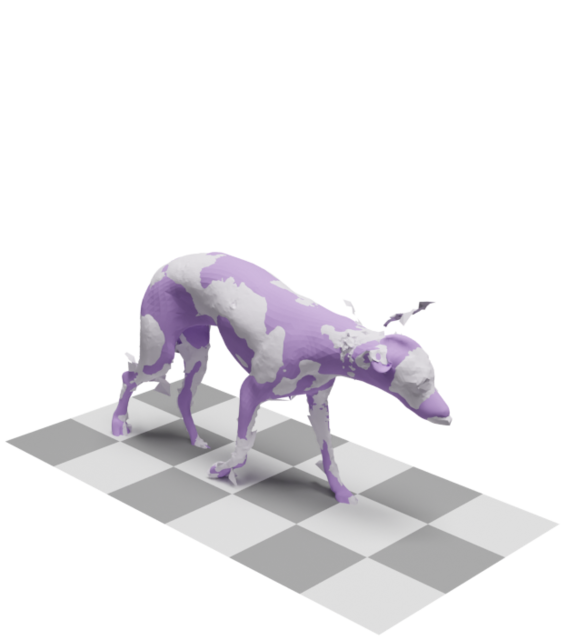}%
        \hspace{-5mm}
        \includegraphics[width=0.2\linewidth]{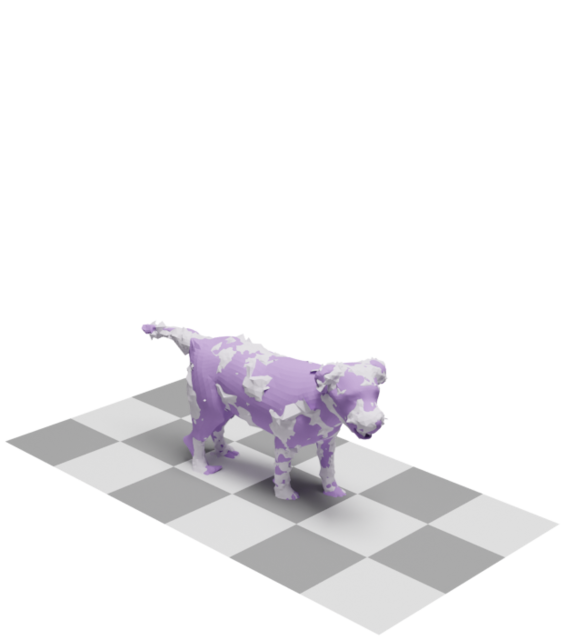}%
        \hspace{-5mm}
        \includegraphics[width=0.2\linewidth]{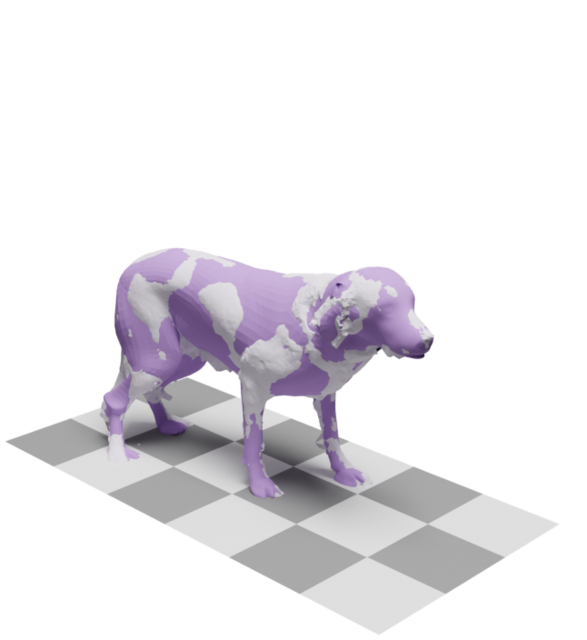}%
        \hspace{-5mm}
        \includegraphics[width=0.2\linewidth]{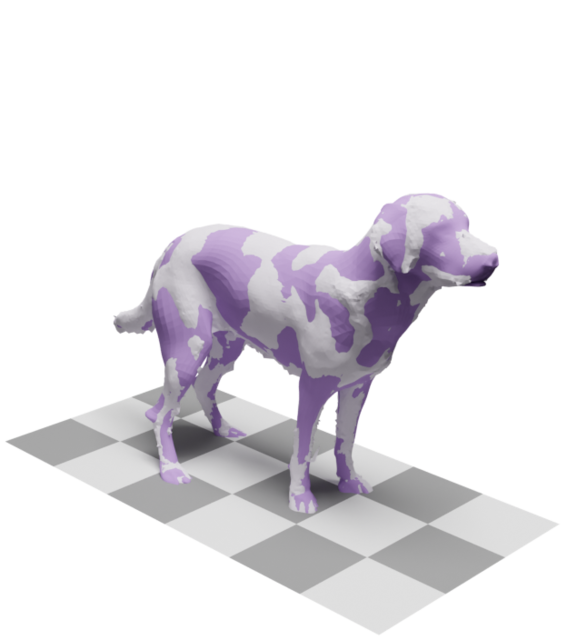}%
        \hspace{-5mm}
        \includegraphics[width=0.2\linewidth]{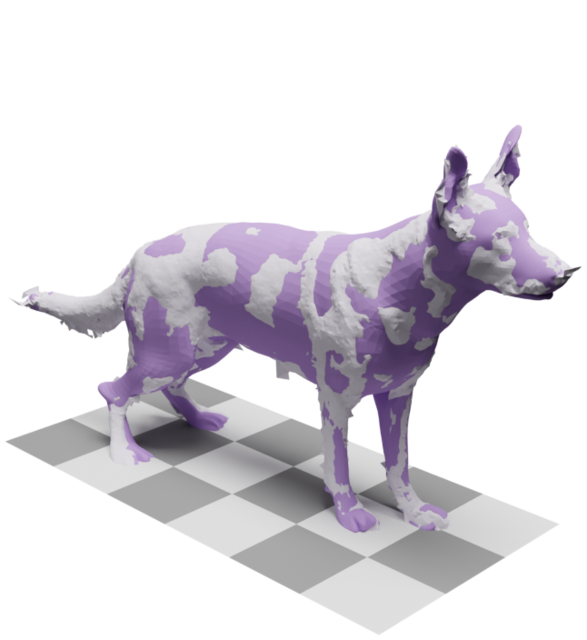}
    }
    
    \vspace{-12mm}
    \resizebox{\linewidth}{!}{%
        \includegraphics[width=0.2\linewidth]{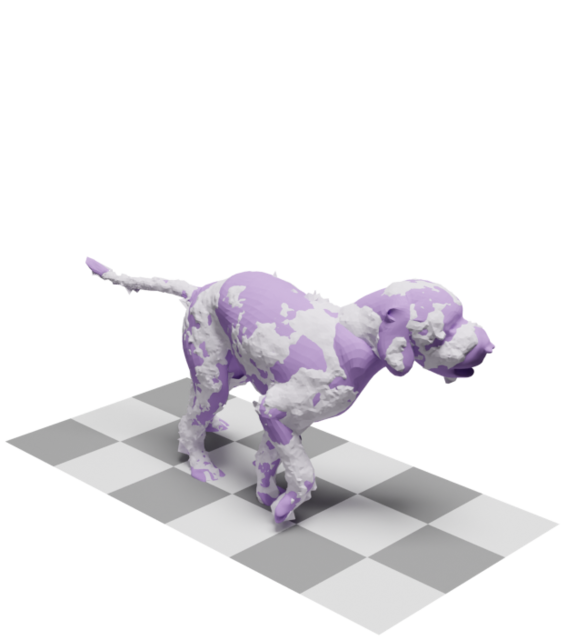}%
        \hspace{-5mm}
        \includegraphics[width=0.2\linewidth]{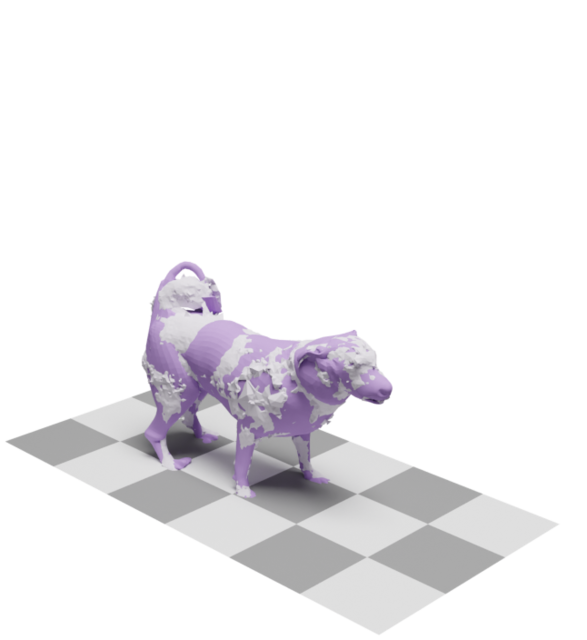}%
        \hspace{-5mm}
        \includegraphics[width=0.2\linewidth]{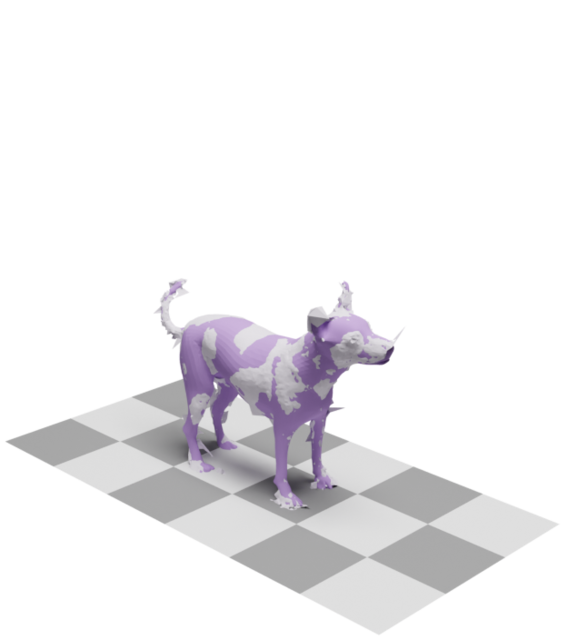}%
        \hspace{-5mm}
        \includegraphics[width=0.2\linewidth]{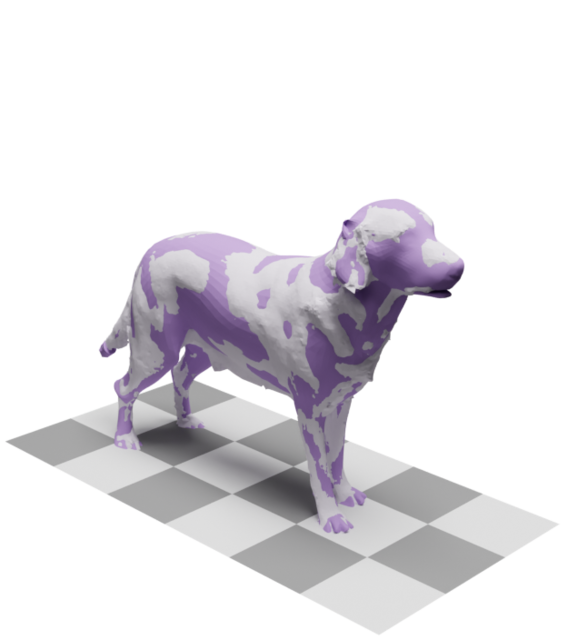}%
        \hspace{-5mm}
        \includegraphics[width=0.2\linewidth]{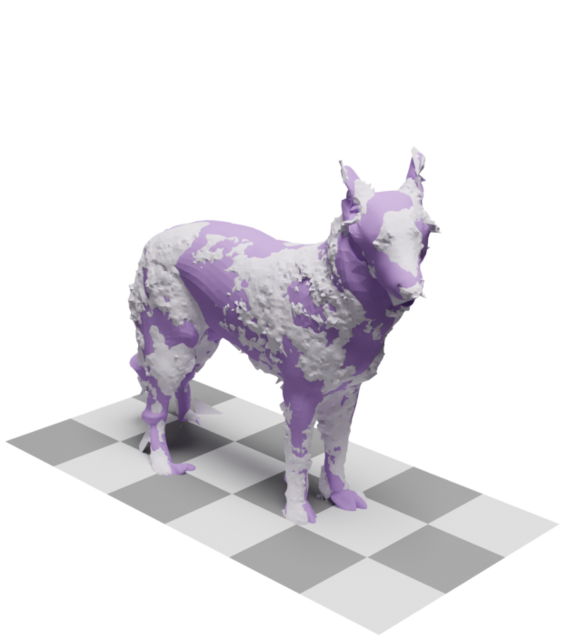}
    }
    \caption{Selected scans from the registration pipeline. We present an overlay of the model (purple) with the scan (white).}
    \label{fig:registration_scan_overlay}
\end{figure}

Finally, we minimize a composite energy of a data term and several regularization terms (\cref{equation:fine-registration-model}):
\begin{equation}
    \begin{split}
        E_\text{model} &= E_{\text{data}} + \mathcal{L}_{\text{model-reg}} \\
        \mathcal{L}_{\text{model-reg}} &= \lambda_{\mathcal{W}_\text{init}} \mathcal{L}_{\mathcal{W}_\text{init}} + \lambda_{\mathcal{W}_\text{smooth}} \mathcal{L}_{\mathcal{W}_\text{smooth}} \\
        & + \lambda_{\mathcal{J}_\text{init}} \mathcal{L}_{\mathcal{J}_\text{init}} \\
        & + \lambda_{{B_P}_\text{smooth}} \mathcal{L}_{{B_P}_\text{smooth}} + \lambda_{B_P}\|B_P\|_F \\
        & + \lambda_{\text{normal}} \mathcal{L}_{\text{normal}} + \lambda_{\text{lap}} \mathcal{L}_{\text{lap}}
    \end{split}
    \label{equation:fine-registration-model}
\end{equation}
Here, $\mathcal{L}_{\mathcal{W}_\text{init}}$ and $\mathcal{L}_{\mathcal{J}_\text{init}}$ penalize deviations from the initial skinning weights and joint regressor, $\mathcal{L}_{\mathcal{W}_\text{smooth}}$ and $\mathcal{L}_{{B_P}_\text{smooth}}$ encourage similar skinning weights and blend shapes across connected vertices, $\lambda_{B_P}\|B_P\|_F$ keeps the pose-dependent blend shapes small, and $\mathcal{L}_{\text{normal}}$ and $\mathcal{L}_{\text{lap}}$ keep the posed and shaped surface smooth.
We refer to \cref{fig:registration_scan_overlay} for an overview of selected subject registrations.

\subsection{Model Construction}
\label{subsection:model_construction}
After registration, we have learned a subject-specific dog template for each recorded subject.
We apply principal component analysis (PCA) on all subjects to initialize the blend shape functions $B_S$ of our model.
The mean template serves as the learned template mesh $\bar{\mathbf{T}}$.
We refer to \cref{figure:learned_model_overview} for an overview of the learned model components.
We present the first eight principal components, each varied by $\pm 3$ standard deviations, illustrating the dominant modes of dog shape variation.
\newcommand{\vc}[1]{$\vcenter{\hbox{#1}}$}
\newcommand{\blockht}{1.75cm}
\newcommand{\rowlab}[1]{\makebox[0pt][r]{\vc{\rotatebox{90}{\makebox[1.6cm][c]{\small\strut #1}}}\hspace{3pt}}}

\begin{figure*}[h]
   \begin{subfigure}[t]{0.3\linewidth}
        \centering
        \resizebox{!}{\blockht}{%
            \begin{minipage}{\linewidth}\centering
                \rowlab{Initialized}%
                \vc{\includegraphics[width=0.31\linewidth]{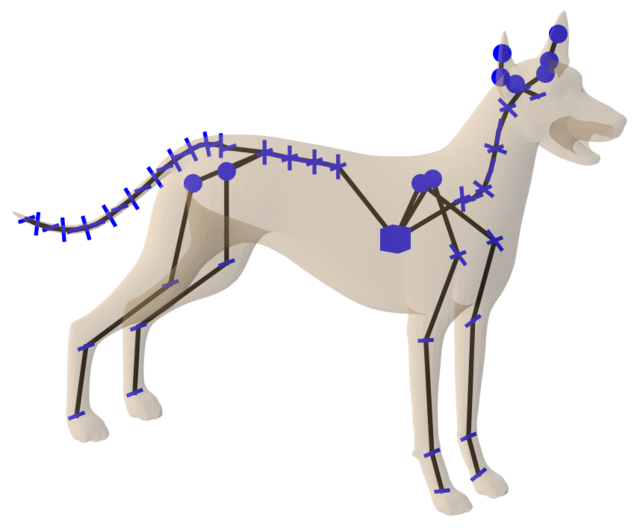}}%
                \vc{\includegraphics[width=0.31\linewidth]{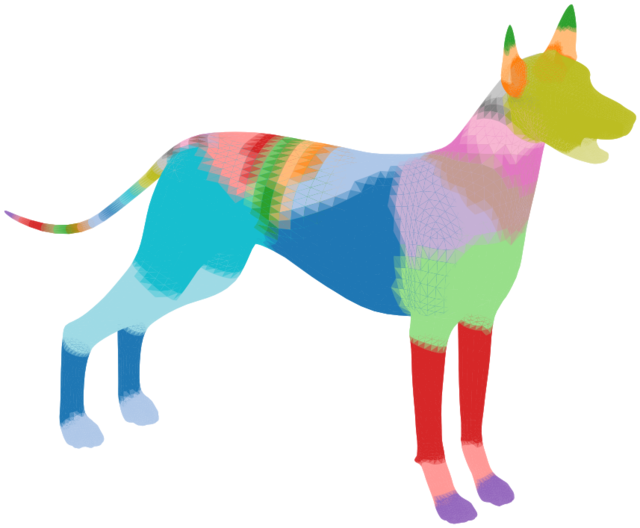}}%
                \vc{\includegraphics[width=0.31\linewidth]{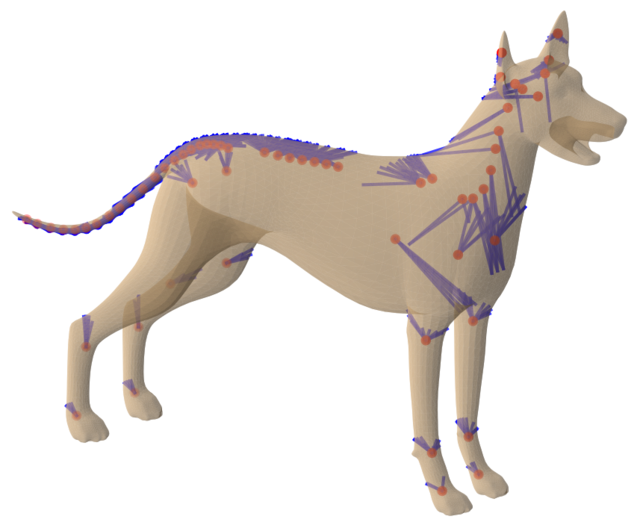}}%
                \vspace{-0.2cm}
                \rowlab{Optimized}%
                \vc{\includegraphics[width=0.31\linewidth]{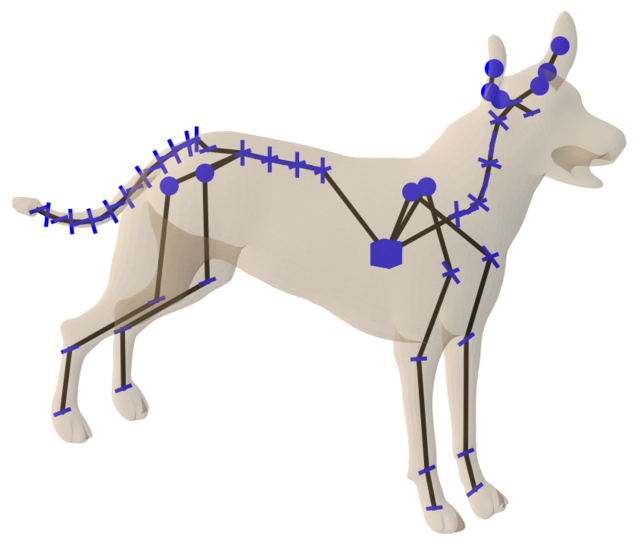}}%
                \vc{\includegraphics[width=0.31\linewidth]{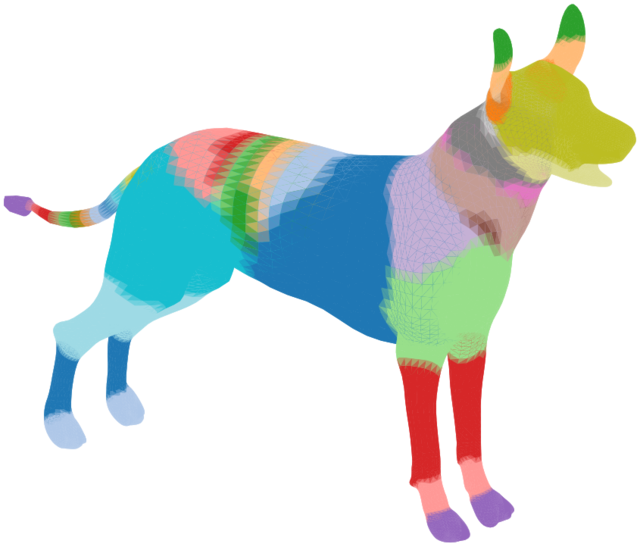}}%
                \vc{\includegraphics[width=0.31\linewidth]{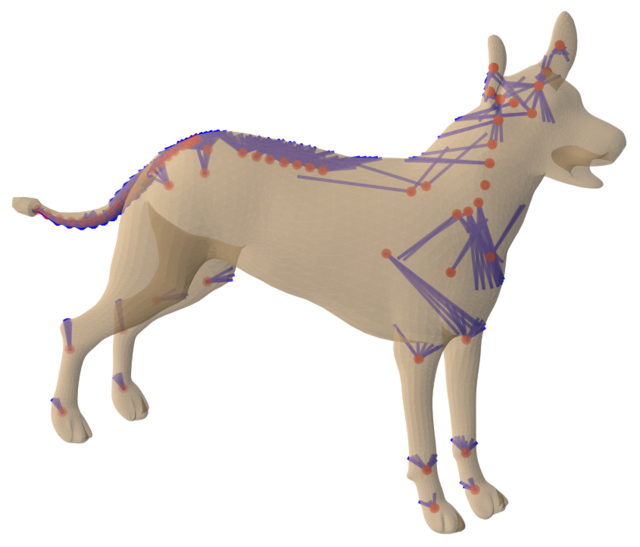}}%
            \end{minipage}%
        }%
        \caption{Model components (left to right): kinematic tree, skinning weights (visualized via argmax over the joint dimension), and joint regressor, each overlaid on the template mesh, before (top) and after (bottom) optimization.}
        \label{figure:PAWS_model_components}
    \end{subfigure}
    \hfill
    \begin{subfigure}[t]{0.65\linewidth}
        \centering
        \resizebox{!}{\blockht}{
            \begin{minipage}{\linewidth}
            \centering
            \vc{\includegraphics[width=0.125\linewidth]{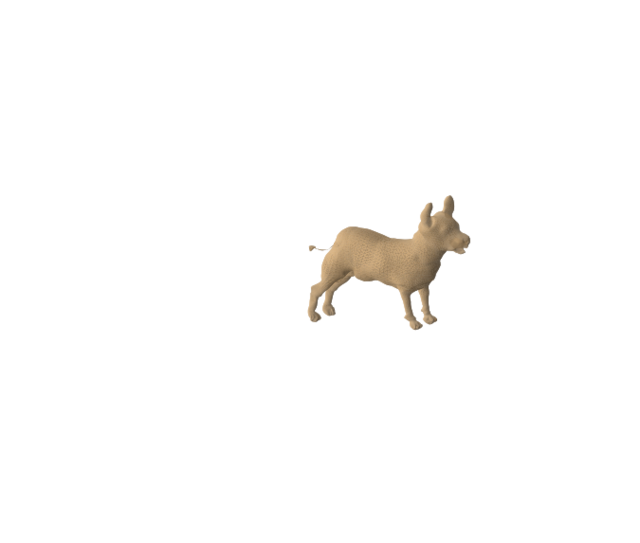}}%
            \vc{\includegraphics[width=0.125\linewidth]{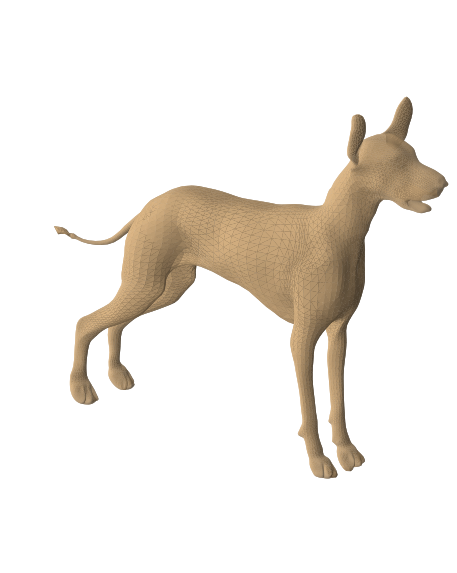}}%
            \vc{\includegraphics[width=0.125\linewidth]{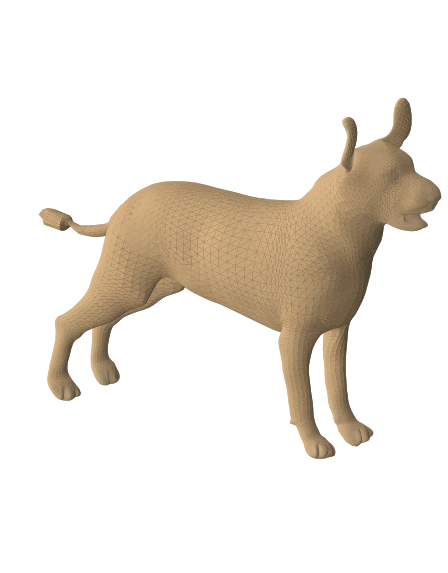}}%
            \vc{\includegraphics[width=0.125\linewidth]{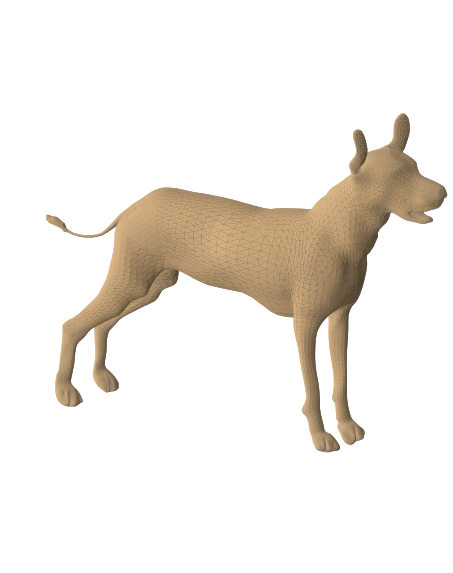}}%
            \vc{\includegraphics[width=0.125\linewidth]{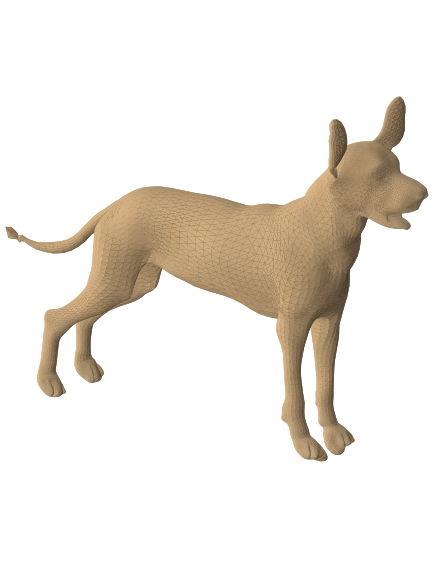}}%
            \vc{\includegraphics[width=0.125\linewidth]{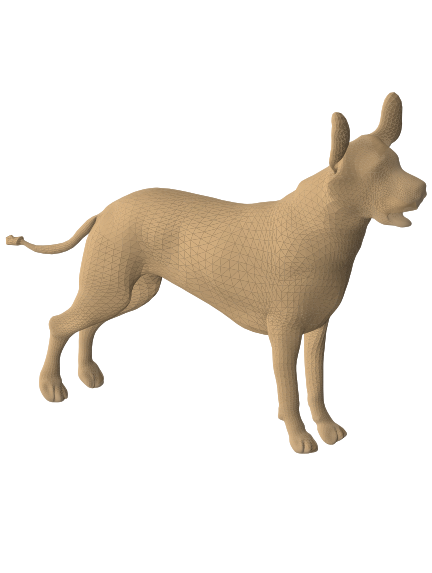}}%
            \vc{\includegraphics[width=0.125\linewidth]{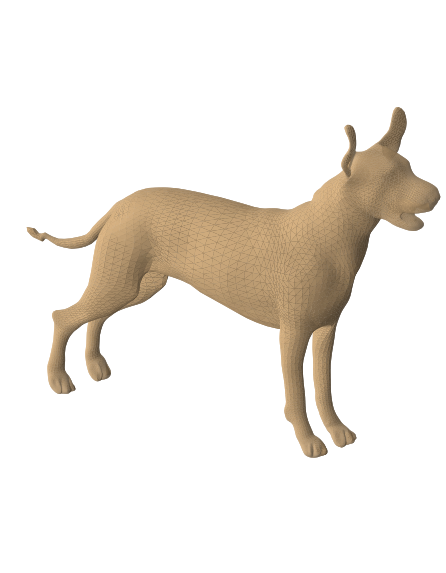}}%
            \vc{\includegraphics[width=0.125\linewidth]{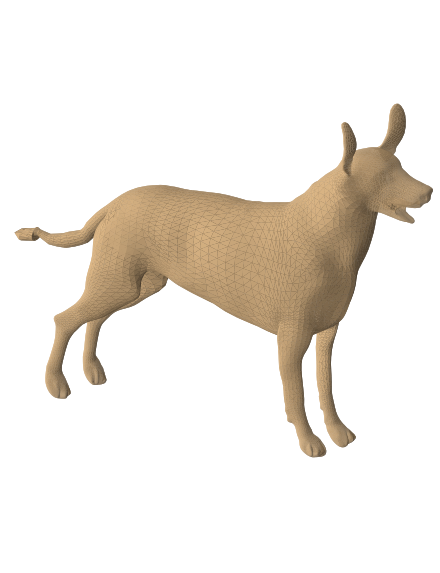}}%
            \makebox[0pt][l]{\hspace{6pt}\vc{\rotatebox{90}{\small$+3\sigma$}}}%
            \par\smallskip
            \smash{%
                \makebox[0.125\linewidth][c]{\small PC1}%
                \makebox[0.125\linewidth][c]{\small PC2}%
                \makebox[0.125\linewidth][c]{\small PC3}%
                \makebox[0.125\linewidth][c]{\small PC4}%
                \makebox[0.125\linewidth][c]{\small PC5}%
                \makebox[0.125\linewidth][c]{\small PC6}%
                \makebox[0.125\linewidth][c]{\small PC7}%
                \makebox[0.125\linewidth][c]{\small PC8}%
            }%
            \vspace{-0.3cm}
            \par\smallskip
            \vc{\includegraphics[width=0.125\linewidth]{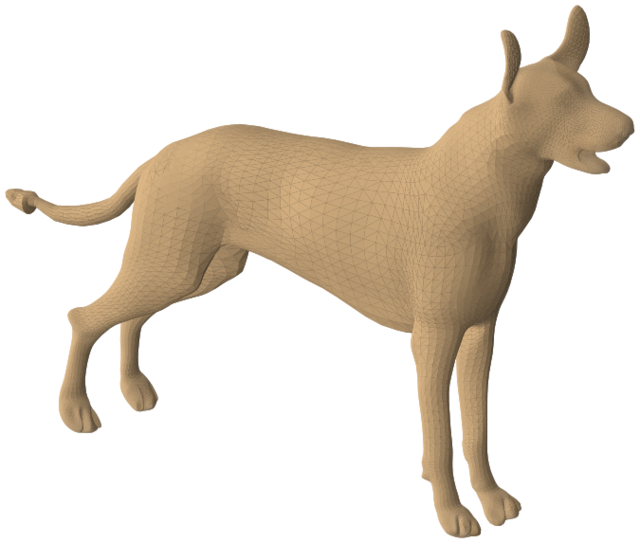}}%
            \vc{\includegraphics[width=0.125\linewidth]{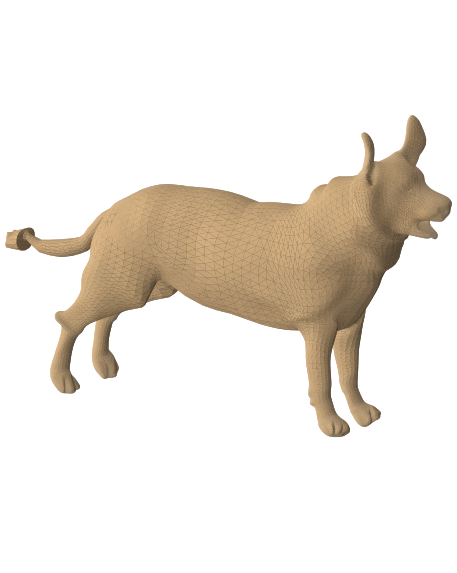}}%
            \vc{\includegraphics[width=0.125\linewidth]{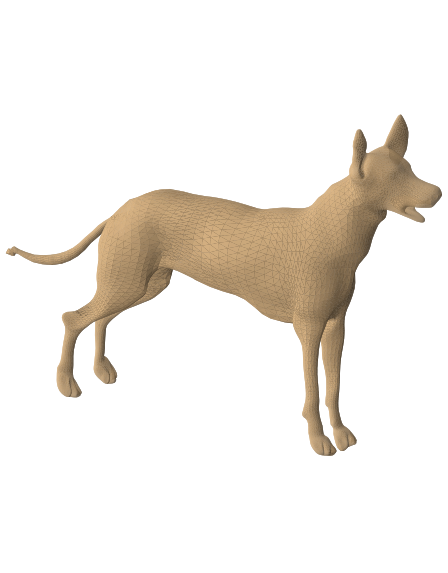}}%
            \vc{\includegraphics[width=0.125\linewidth]{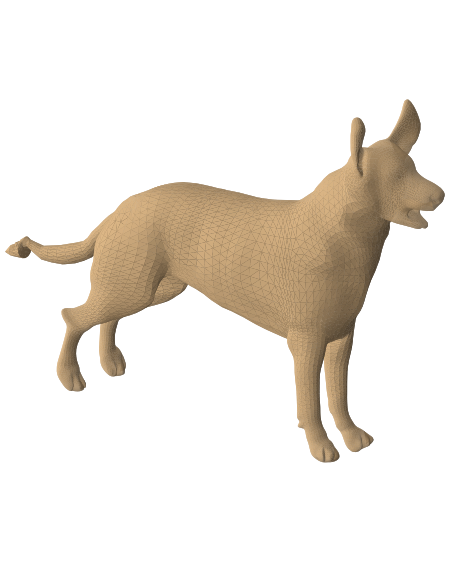}}%
            \vc{\includegraphics[width=0.125\linewidth]{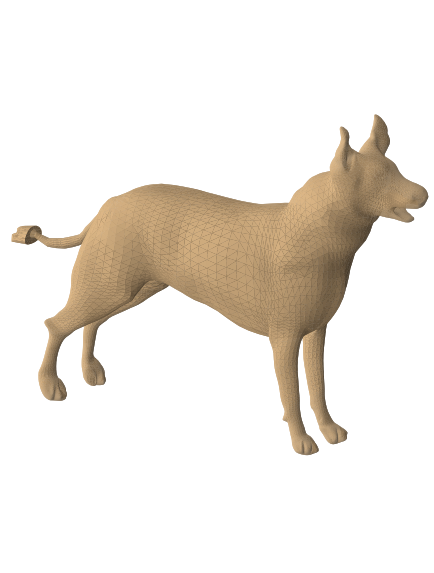}}%
            \vc{\includegraphics[width=0.125\linewidth]{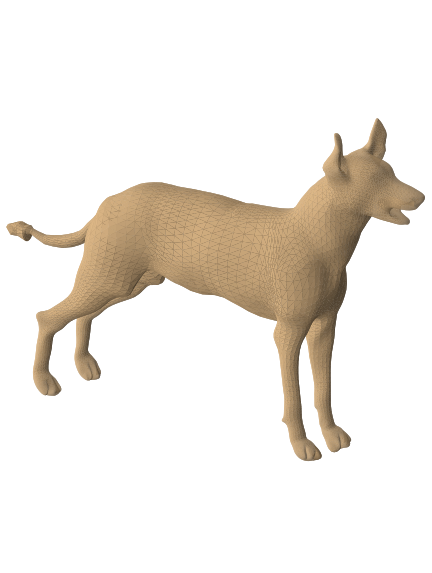}}%
            \vc{\includegraphics[width=0.125\linewidth]{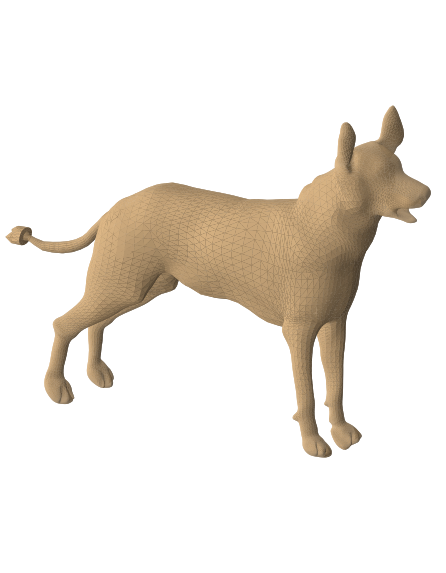}}%
            \vc{\includegraphics[width=0.125\linewidth]{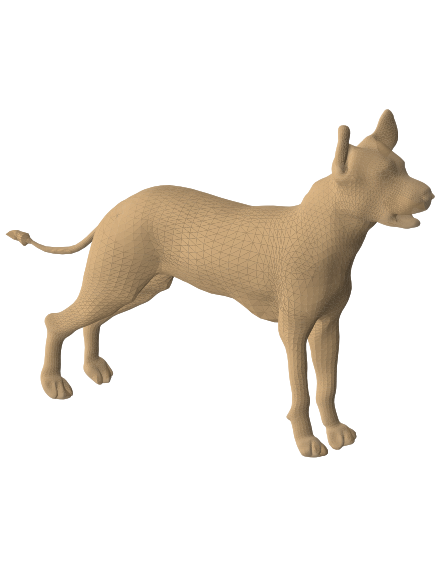}}%
            \makebox[0pt][l]{\hspace{6pt}\vc{\rotatebox{90}{\small$-3\sigma$}}}%
            \end{minipage}
        }
        \caption{The first eight principal components (PC1--PC8) of the learned shape space, obtained by PCA over the per-subject templates. Each column displays the mean displaced by $\pm3\sigma$ along one component (top and bottom rows). The leading mode captures general size variation, while later modes capture finer details such as variations in head shape.}
        \label{figure:learned_model_shape_space_overview}
    \end{subfigure}
    \caption{Overview of the PAWS model: the initialized and optimized model parameters (\cref{figure:PAWS_model_components}) and the learned shape space (\cref{figure:learned_model_shape_space_overview}).}
    \label{figure:learned_model_overview}
\end{figure*}

\subsection{Implementation Details}
\label{subsection:registration_implementation_details}
The registration pipeline is implemented in Python, using PyTorch for automatic differentiation. 

The data and regularization weights are set empirically. 
We use $\lambda_{\text{kp}}{=}25$, $\lambda_{\text{s2m}}{=}1$, $\lambda_{\text{m2s}}{=}0.45$, $\lambda_\beta{=}0.1$, $\lambda_\theta{=}0.001$, $\lambda_\text{lim}{=}0.5$, $\lambda_{\Delta V^d}{=}10$, $\lambda_{\text{ARAP}}{=}0.005$, $\lambda_\text{lap}{=}500$, $\lambda_\text{normal}{=}0.05$.
For model optimization (\cref{subsection:model_parameters}), we additionally set $\lambda_{\mathcal{W}_\text{init}}=0.05$, $\lambda_{\mathcal{W}_\text{smooth}}=1$, $\lambda_{\mathcal{J}_\text{init}}=1$, $\lambda_{{B_P}_\text{smooth}}=0.5$, and $\lambda_{B_P}=0.01$.

We optimize pose parameters using an over-parameterized encoding that we map to the rotation manifold at each iteration to avoid discontinuities. 
For ball joints we use unconstrained $3\times3$ matrices, projected onto $SO(3)$ via special orthogonal Procrustes. 
For hinge joints we use $(\sin,\cos)$ pairs, converted to an angle via $\operatorname{atan2}$ and mapped to $SO(3)$ via Rodrigues' formula using the fixed rotation axis.
All parameters are initialized to the rest pose.
This leaves the kinematic tree with 62 joints represented with 221 parameters. 
We refer to the appendix for further details on the pose parameterization and implementation details.

\section{Experiments}
\label{section:experiments}
We evaluate our model both quantitatively and qualitatively against relevant baselines.
First, we present model fitting results by registering our model to a held-out set of scans in \cref{subsection:experiments_model_fitting}.
Second, we perform parameter regression by predicting model parameters directly from images in \cref{subsection:experiments_regression}.

\subsection{Model Fitting}
\label{subsection:experiments_model_fitting}
\begin{figure*}
    \centering

    \newlength{\wscan}\setlength{\wscan}{0.1818\textwidth}
    \newlength{\wmeth}\setlength{\wmeth}{0.2727\textwidth}

    \newlength{\gap}\setlength{\gap}{-4mm}
    \newlength{\wimg}\setlength{\wimg}{0.325\wmeth}
    \newlength{\wref}\setlength{\wref}{0.49\wscan}
    \newlength{\zoomsize}\setlength{\zoomsize}{0.75cm}

    \newcommand{\imcrop}[2]{\includegraphics[width=#1]{#2}}

    \newcommand{\distzoom}[4]{%
      \begin{tikzpicture}[spy using outlines={rectangle, red!85!black, very thick, magnification=3, size=\zoomsize, connect spies}]
        \node[anchor=south west, inner sep=0pt] (im) at (0,0) {\includegraphics[width=#1]{#2}};
        \coordinate (sp) at ($(im.south west)!#3!(im.south east)$);
        \coordinate (sp) at ($(sp)!#4!(im.north west -| sp)$);
        \spy on (sp) in node [anchor=north east] at (im.north east);
      \end{tikzpicture}%
    }

    \newcommand{\refcell}[1]{%
      \resizebox{\wscan}{!}{%
        \imcrop{\wref}{figures/selected_registrations_comparison/BARC/#1/scan_render_0000_view+036deg.png}\hspace{\gap}%
        \imcrop{\wref}{figures/selected_registrations_comparison/BARC/#1/scan_render_0000_view-144deg.png}}%
    }

    \newcommand{\methodcell}[3]{%
      \resizebox{\wmeth}{!}{%
        \imcrop{\wimg}{#1/distance_render_0000_view+036deg.png}\hspace{\gap}%
        \imcrop{\wimg}{#1/model_render_0000_view-144deg.png}\hspace{\gap}%
        \distzoom{\wimg}{#1/model_render_0000_view+036deg.png}{#2}{#3}}%
    }

    \newcommand{\comprow}[3]{%
      \makebox[\wscan]{\refcell{#1}}%
      \makebox[\wmeth]{\methodcell{figures/selected_registrations_comparison/BARC/#1}{#2}{#3}}%
      \makebox[\wmeth]{\methodcell{figures/selected_registrations_comparison/BITE/#1}{#2}{#3}}%
      \makebox[\wmeth]{\methodcell{figures/selected_registrations_comparison/PAWS/#1}{#2}{#3}}%
    }

    \makebox[\wscan]{Reference Scan}%
    \makebox[\wmeth]{BARC}%
    \makebox[\wmeth]{BITE}%
    \makebox[\wmeth]{Ours}\\[-5pt]
    \makebox[\wscan]{\rule{0.9\wscan}{0.1pt}}%
    \makebox[\wmeth]{\rule{0.9\wmeth}{0.1pt}}%
    \makebox[\wmeth]{\rule{0.9\wmeth}{0.1pt}}%
    \makebox[\wmeth]{\rule{0.9\wmeth}{0.1pt}}\\[2pt]

    \comprow{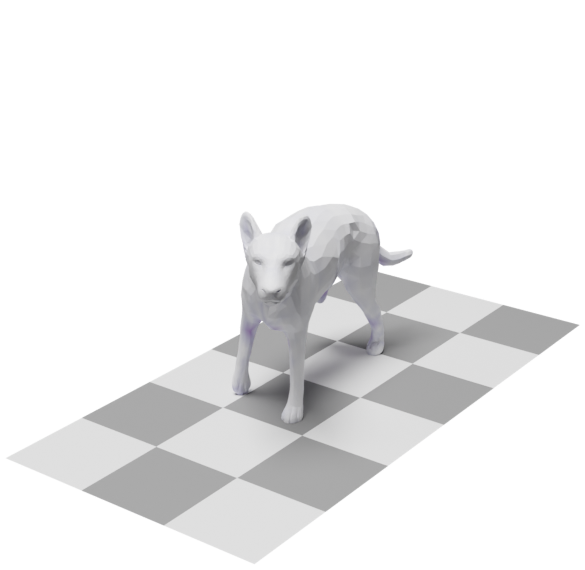}{0.6}{0.45}\\[-10pt]
    \comprow{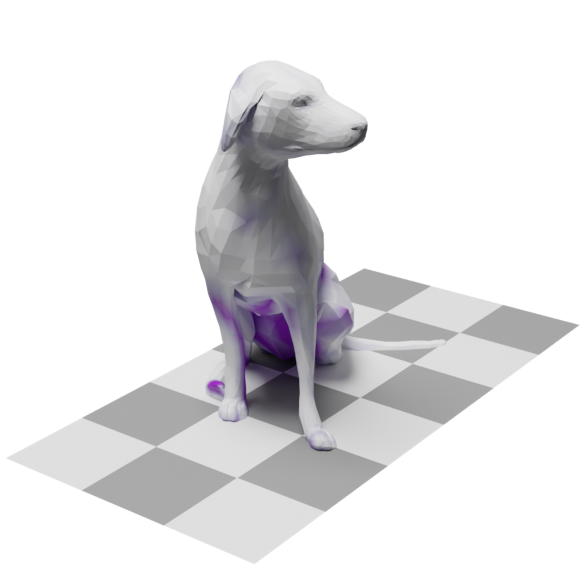}{0.55}{0.35}\\[-10pt]
    \comprow{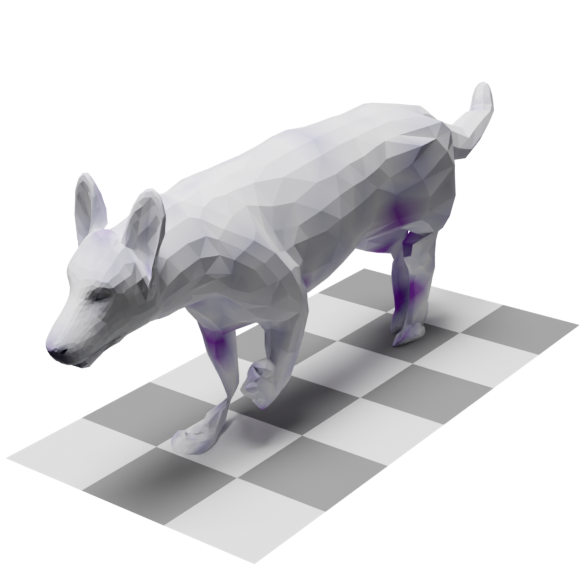}{0.4}{0.25}

    \includegraphics[width=0.5\textwidth]{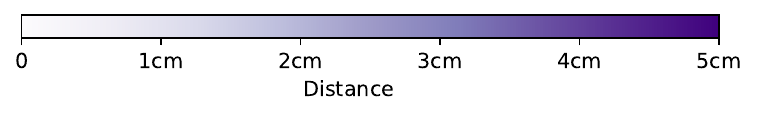}\\[-4pt]
    \caption{Qualitative comparison of model fittings. From left to right, we show the target scan (two views), followed by the fitting obtained with BARC~\cite{BARC}, BITE~\cite{BITE}, and our model, each rendered from two viewpoints. The first column visualizes per-vertex surface-to-scan distances using a purple colormap. The remaining two views show the shaped and posed model. We additionally present red zoomed-in views, highlighting cases where our model correctly resolves joint articulation where both previous models produce twisted joint configurations.}
    \label{fig:scan_registration_overview}
\end{figure*}

We evaluate the surface reconstruction accuracy by fitting our model to a held-out set of scans.
We follow the same approach as our registration pipeline (see \cref{subsection:registration}).
During these experiments we limit all optimization steps to remain within the model's parametric pose and shape space.
The per-vertex displacements, as well as the model-defining parameters remain unoptimized.
We compare against relevant parametric dog models, namely BARC~\cite{BARC} and BITE~\cite{BITE}.
To ensure a fair evaluation, we use leave-one-out cross-validation (LOOCV). 
For each of the 31 recorded subjects, we construct a limited shape space from the remaining 30 dogs and fit the resulting model to the scans of the left-out dog.
All metrics are computed using uniformly sampled points on both surfaces with a fixed density, excluding the inner mouth of each model's template.
Moreover, normal consistency is determined as the mean absolute cosine similarity between nearest-neighbour matched points, averaged over both directions.
We report quantitative results in \cref{tabel:registration_scan_distance} and qualitative results in \cref{fig:scan_registration_overview}.
\begin{table}[h]
    \centering
    \sisetup{separate-uncertainty=true, table-align-uncertainty=true, detect-all=true}
    \caption{Surface reconstruction accuracy against prior parametric models (BARC~\cite{BARC}, BITE~\cite{BITE}, and ours). We report Chamfer (CD), scan-to-mesh (S2M), and mesh-to-scan (M2S) distances (mm), normal consistency (NC), and F-scores~\cite{tatarchenko2019single} at distance thresholds $\tau\in\{1,5,10\} \mathrm{mm}$, as mean $\pm$ std across subjects (averaged over poses within subjects first). Best method highlighted in bold per column when it is significantly better than all other methods (paired Wilcoxon signed-rank test, Holm-corrected, $p<0.05$).}
    \label{tabel:registration_scan_distance} 
    \resizebox{\linewidth}{!}{%
        \begin{tabular}{l
            S[table-format=2.2(1.2)]
            S[table-format=2.2(1.2)]
            S[table-format=2.2(1.2)]
            S[table-format=2.2(1.2)]
            S[table-format=2.2(1.2)]
            S[table-format=2.2(1.2)]
            S[table-format=2.2(1.2)]}
            \toprule[1.5pt]
            {Method} & {CD$\downarrow$} & {S2M$\downarrow$} & {M2S$\downarrow$} & {NC$\uparrow$} & {F@1mm$\uparrow$} & {F@5mm$\uparrow$} & {F@10mm$\uparrow$} \\
            \midrule
            BARC & 9.38 +- 2.25 & 5.92 +- 3.36 & 8.66 +- 2.12 & 0.85 +- 0.03 & 0.15 +- 0.04 & 0.59 +- 0.09 & 0.80 +- 0.06 \\
BITE & 9.37 +- 2.24 & 6.04 +- 3.48 & 8.52 +- 1.99 & 0.85 +- 0.03 & 0.15 +- 0.03 & 0.58 +- 0.08 & 0.80 +- 0.06 \\
Ours & \bfseries 8.74 +- 2.21 & \bfseries 5.08 +- 3.48 & \bfseries 7.75 +- 1.93 & \bfseries 0.86 +- 0.03 & \bfseries 0.21 +- 0.04 & \bfseries 0.70 +- 0.07 & \bfseries 0.85 +- 0.04 \\
            \bottomrule[1.5pt]
        \end{tabular}%
    }
\end{table}

\subsection{Regression}
\label{subsection:experiments_regression}
\begin{figure}[h]
    \centering

    \includegraphics[width=0.2\linewidth]{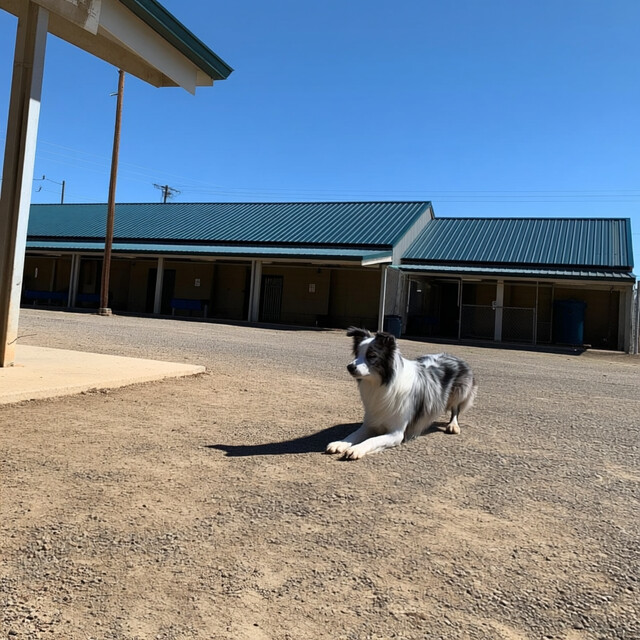}%
    \includegraphics[width=0.2\linewidth]{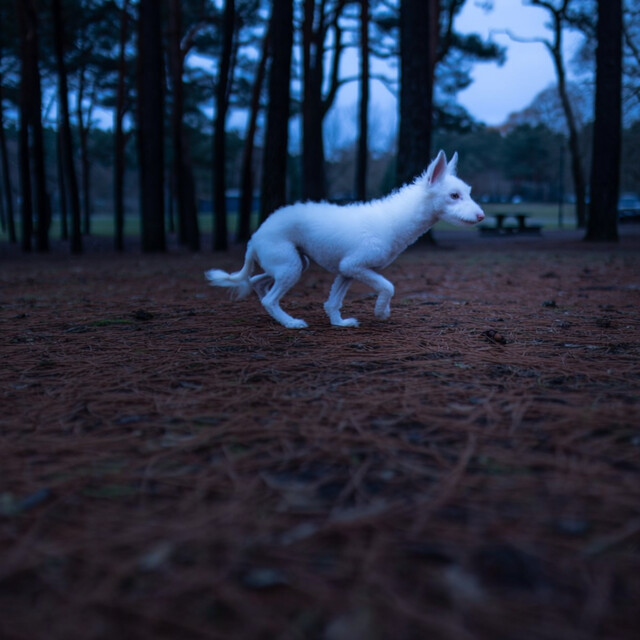}%
    \includegraphics[width=0.2\linewidth]{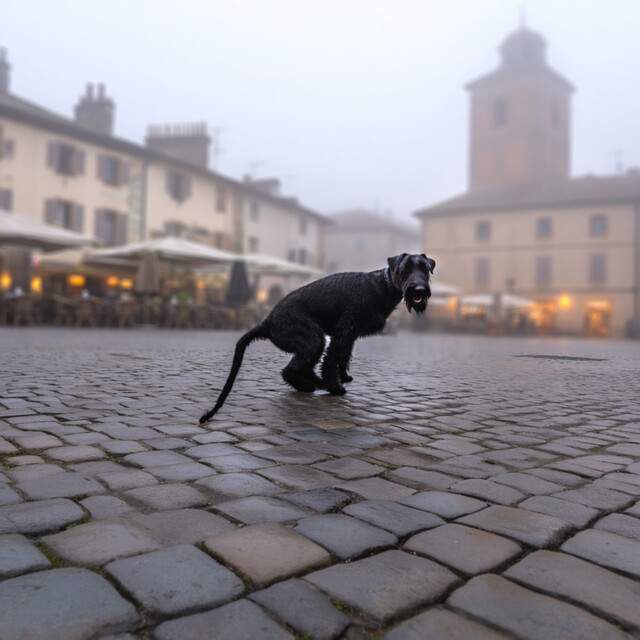}%
    \includegraphics[width=0.2\linewidth]{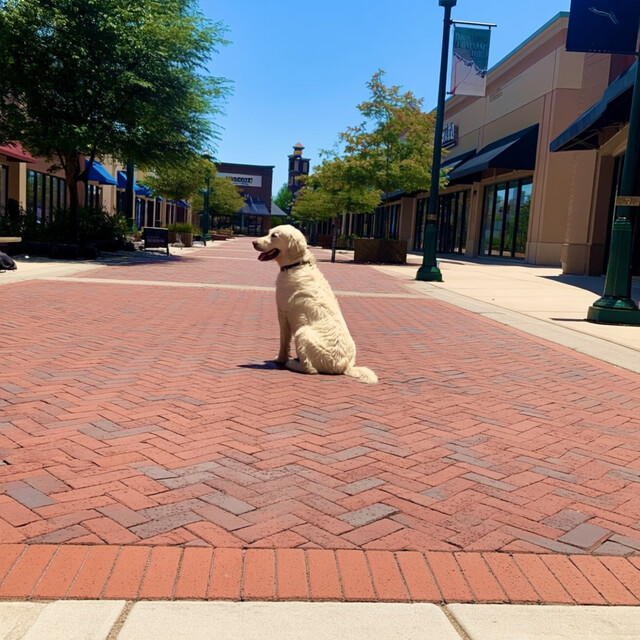}%
    \includegraphics[width=0.2\linewidth]{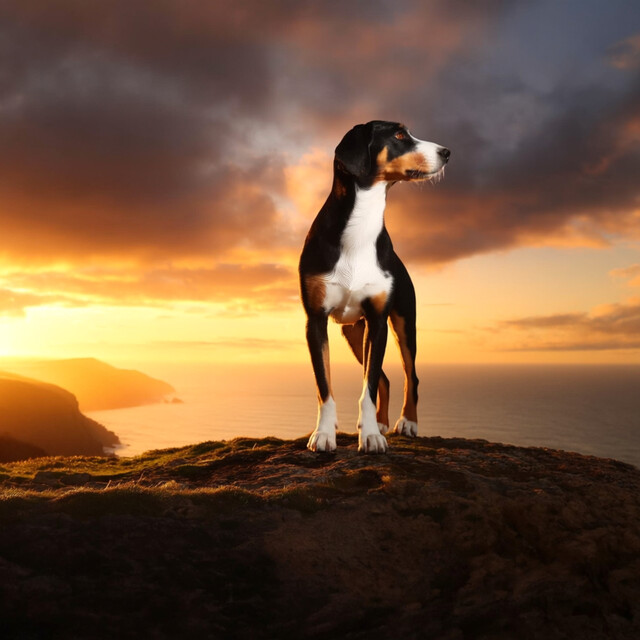}%
    \vspace{-1pt}
    \includegraphics[width=0.2\linewidth]{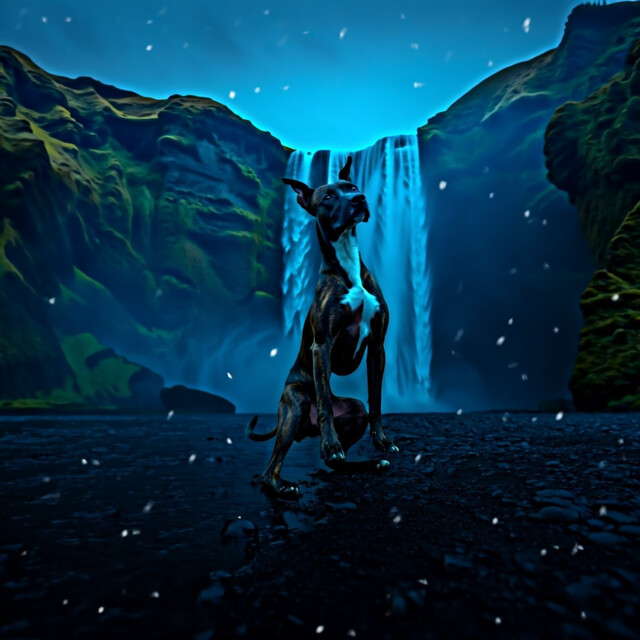}%
    \includegraphics[width=0.2\linewidth]{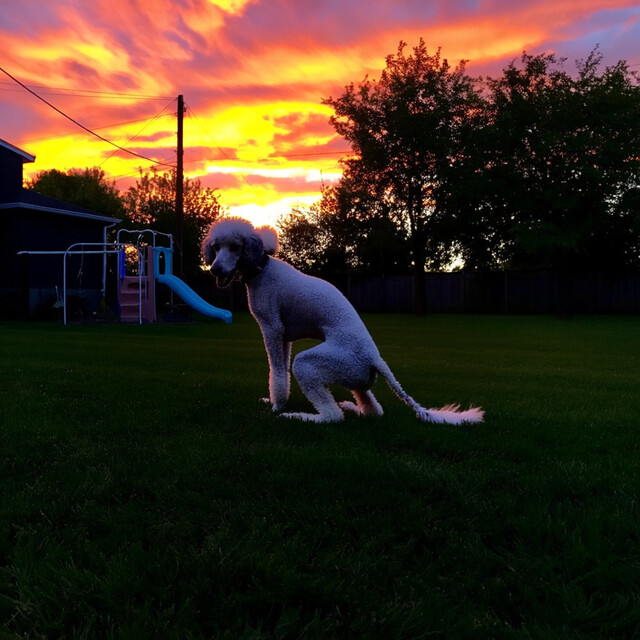}%
    \includegraphics[width=0.2\linewidth]{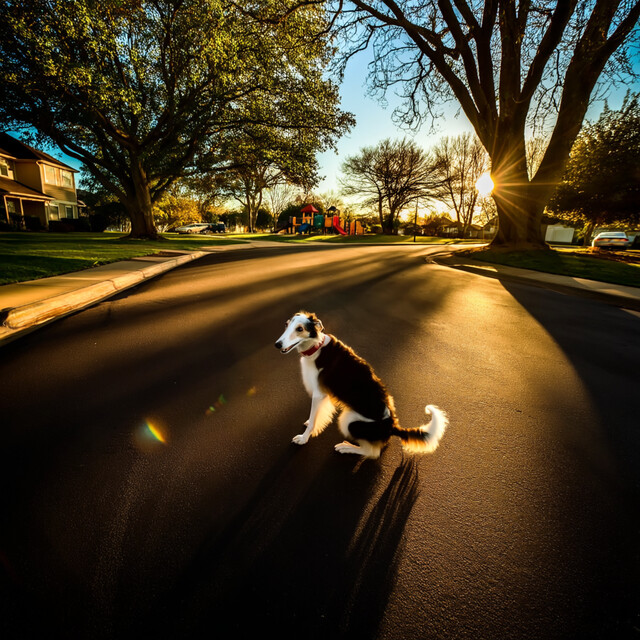}%
    \includegraphics[width=0.2\linewidth]{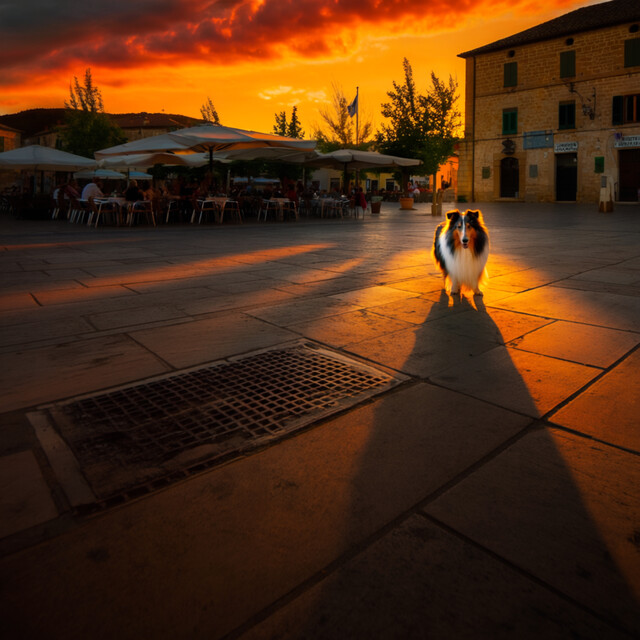}%
    \includegraphics[width=0.2\linewidth]{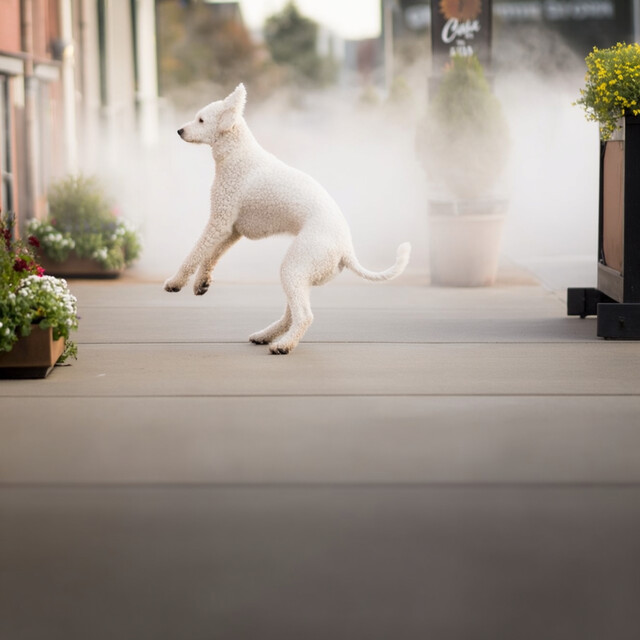}%
    \caption{Randomly sampled images from the test set of our synthetic dataset. In total the dataset comprises 12288 images of natural dog appearances.}
    \label{fig:synthetic_dog_images_overview}
\end{figure}

To train a regressor that predicts dog pose and shape parameters directly from images, we synthesize a dataset of natural dog appearances in varying poses, following Niewiadomski et al.~\cite{GenZoo}.
We randomly sample poses from the pool of registered poses produced by our registration pipeline, augmenting them with random ground-plane rotations and sagittal joint flips.
Shape parameters are randomly sampled from the PCA shape basis within a fixed radius.
For each sampled pair, we render a Canny edge map and depth map, and combine them with a text prompt assembled from randomly sampled attributes (breed, coat type, and scene) and the dog's orientation derived from the pose parameters. 
We generate a natural text prompt using Qwen3-VL-8B-Instruct~\cite{Qwen3}.
In total, we synthesize 12288 images, reserving a small percentage for validation and testing (see \cref{fig:synthetic_dog_images_overview}).
We augment the training set with StanfordExtra~\cite{WLDO}, Animal3D~\cite{Animal3D}, CtrlAni3D~\cite{AniMer}, and AnimalPose~\cite{AnimalPoseDataset}, resulting in 31685 training images.
Using this dataset, we train the architecture with a full-perspective camera model.
Depending on the available annotations, we supervise pose, shape, and translation parameters, as well as 2D reprojection and reference-aligned 3D losses on keypoints and joints.
We also provide cropped bounding-box information to the regressor, following Li et al.~\cite{CLIFF}.
We finetune a pretrained ResNet~\cite{resnet} with an HMR~\cite{HMR} regressor head, with four pose parameterizations:
axis-angle (AA), 6D with Gram-Schmidt orthogonalization ($\mathbb{R}^6$+GS), 9D with SVD orthogonalization ($\mathbb{R}^9$+SVD), and our anatomical degrees of freedom (ADoF).
Orthogonalization is applied during both training and inference for $\mathbb{R}^6$+GS and $\mathbb{R}^9$+SVD.
Our 62-joint kinematic tree is represented with 186, 372, and 558 parameters using AA, $\mathbb{R}^6$+GS, and $\mathbb{R}^9$+SVD, respectively, while our ADoF parameterization uses 221 parameters (\cref{subsection:registration_implementation_details}).

\subsection{Regression Evaluation}
We evaluate the regression architecture using three datasets: StanfordExtra~\cite{WLDO}, Animal3D~\cite{Animal3D}, and CtrlAni3D~\cite{AniMer}.
Given our goal of estimating dog pose and shape parameters from images, we restrict the evaluation to dog images, thus excluding evaluation on any other quadrupeds. 
For 2D alignment we report the Percentage of Correct Keypoints (PCK) and the silhouette Intersection over Union (IoU).
A keypoint is considered correct when its distance to the ground truth is below a fraction of the ground-truth head-to-tail distance.
We report $\mathrm{PCK@HTH}$ (with the threshold set to half the head-to-tail distance), together with $\textrm{PCK@0.1}$ and $\textrm{PCK@0.15}$.
For 3D evaluation, we report the scale-aligned mean per-joint position error (S-MPJPE), the Procrustes-aligned mean per-joint position error (PA-MPJPE), and the Procrustes-aligned mean per-vertex position error (PA-MPVPE).
The per-joint errors act as a measure of general pose alignment, whereas the PA-MPVPE measures shape and surface reconstruction accuracy.
We use a pre-computed vertex-to-vertex mapping between our model's template mesh and the ground-truth annotations of each dataset.
Moreover, we define a per-joint mapping between the kinematic tree of our model and each dataset's annotations. 
For evaluation on the real datasets, namely StanfordExtra and Animal3D, we use all available ground-truth annotations, namely the 24 and 26 keypoints respectively.
Animal3D additionally provides 3D ground-truth annotations that we utilize.
Finally, when evaluating on CtrlAni3D we compute evaluation metrics against the synthetic ground-truth annotations.

\subsection{Discussion}
We report quantitative results in \cref{table:regression_results_overview} and qualitative comparisons in \cref{figure:regression_results_overview}.
Our model trails the strongest methods, particularly in IoU and on Animal3D, while reaching comparable PCK.
Animal3D and CtrlAni3D provide ground truth as SMAL meshes, which SMAL-based methods predict directly, whereas our predictions must be mapped to the SMAL topology via a vertex correspondence.
Moreover, AniMer and AniMer+ use a larger transformer backbone, whereas we use a ResNet-50 with an HMR head.
Our ADoF parameterization matches AA, $\mathbb{R}^6$+GS, and $\mathbb{R}^9$+SVD in accuracy while yielding anatomically plausible joints even under self-occlusion (\cref{figure:regression_results_overview}).
Finally, when fitted to real scans, our model improves on BARC and BITE in surface reconstruction accuracy (\cref{tabel:registration_scan_distance}).

\begin{table*}[t]
    \centering
    \caption{Quantitative comparison of monocular 3D dog reconstruction methods. Each row contains the evaluation results for a given method across different datasets and metrics. From top to bottom, WLDO~\cite{WLDO}, BARC~\cite{BARC}, BITE~\cite{BITE}, GenZoo~\cite{GenZoo}, AniMer~\cite{AniMer}, AniMer+~\cite{AniMerPlus}, and ours.}
    \vspace{-0.2cm}
    \label{table:regression_results_overview} 
    \sisetup{detect-weight=true, detect-family=true}
    \resizebox{\textwidth}{!}{%
        \begin{tabular}{l
                *{4}{S[table-format=2.2]}
                *{4}{S[table-format=2.2]}
                *{3}{S[table-format=3.2]}
                *{4}{S[table-format=2.2]} 
                *{3}{S[table-format=3.2]}}
            \toprule[1.5pt]
            & \multicolumn{4}{c}{StanfordExtra} & \multicolumn{7}{c}{Animal3D [dogs]} & \multicolumn{7}{c}{CtrlAni3D [dogs]} \\
            \cmidrule(lr){2-5} \cmidrule(lr){6-12} \cmidrule(lr){13-19}
            & \multicolumn{4}{c}{2D} 
            & \multicolumn{4}{c}{2D} & \multicolumn{3}{c}{3D} 
            & \multicolumn{4}{c}{2D} & \multicolumn{3}{c}{3D} \\
            \cmidrule(lr){2-5} \cmidrule(lr){6-9} \cmidrule(lr){10-12} \cmidrule(lr){13-16} \cmidrule(lr){17-19}
            {Method} & 
            {P@0.1$\uparrow$} & {P@0.15$\uparrow$} & {P@HTH$\uparrow$} & {IoU$\uparrow$} & 
            {P@0.1$\uparrow$} & {P@0.15$\uparrow$} & {P@HTH$\uparrow$} & {IoU$\uparrow$} & {S-MPJPE$\downarrow$} & {PA-MPJPE$\downarrow$} & {PA-MPVPE$\downarrow$} &
            {P@0.1$\uparrow$} & {P@0.15$\uparrow$} & {P@HTH$\uparrow$} & {IoU$\uparrow$} & {S-MPJPE$\downarrow$} & {PA-MPJPE$\downarrow$} & {PA-MPVPE$\downarrow$} \\
            \midrule
            WLDO & \bfseries 61.8864 & 76.2193 & 95.6330 & 72.8428 & 25.6016 & 40.8788 & 83.3627 & 58.5912 & 483.3004 & 208.1101 & 210.2291 & 48.2231 & 65.8497 & 94.1412 & 61.4149 & 476.4134 & 206.6354 & 185.3897 \\
            BARC & 61.1337 & \bfseries 78.9622 & 96.4758 & 74.4689 & 34.5448 & 52.9026 & 87.2649 & 61.5532 & 323.4863 & 131.2839 & 174.0069 & 64.1913 & 79.7495 & 95.0672 & 67.9049 & 178.4322 & 108.6765 & 133.3256 \\
            BITE & 55.8028 & 77.0986 & 97.9253 & 75.2667 & 34.3393 & 53.2080 & 90.3781 & 65.0230 & 318.7101 & 137.8167 & 157.5004 & 69.3409 & 82.3839 & 96.4843 & 71.9186 & 175.9601 & 113.8521 & 119.5849 \\
            GenZoo & 56.2562 & 75.2619 & 97.9633 & \bfseries 77.7768 & 51.8369 & 70.8659 & 96.9893 & \bfseries 72.6003 & 197.1815 & 124.6991 & 138.6311 & 86.0900 & 93.4022 & 99.4573 & \bfseries 78.3769 & 145.2144 & 102.2802 & 110.5134 \\
            AniMer & 57.5578 & 77.1163 & 97.4214 & 69.2301 & \bfseries 64.2336 & \bfseries 81.3948 & \bfseries 98.6465 & 67.9734 & \bfseries 123.8970 & \bfseries 74.4674 & \bfseries 87.1905 & \bfseries 87.4325 & \bfseries 94.2194 & \bfseries 99.7770 & 76.7855 & \bfseries 58.1090 & \bfseries 40.5651 & \bfseries 47.0759 \\
            AniMer+ & 52.0063 & 73.9177 & 97.2794 & 68.1390 & 60.1218 & 78.7233 & 98.4871 & 66.6205 & 124.8101 & 75.8398 & 89.6014 & 86.3593 & 93.6532 & 99.7628 & 76.9733 & 64.5860 & 45.0539 & 52.5805 \\
            \hline
            Ours [AA] & 58.3926 & 77.2874 & 98.0881 & 57.9137 & 40.0000 & 60.4628 & 94.3112 & 52.8415 & 196.8268 & 133.0252 & 134.5709 & 72.7632 & 86.0580 & 96.3740 & 60.1867 & 152.0263 & 124.4150 & 133.6072 \\
            Ours [$\mathbb{R}^6$ + GS] & 58.1730 & 77.0563 & 98.1976 & 59.3637 & 41.5992 & 60.3674 & 94.2556 & 54.2287 & 183.0635 & 129.6507 & 132.2147 & 77.0192 & 87.2464 & 96.0383 & 62.0921 & 159.5785 & 128.2392 & 139.7101 \\
            Ours [$\mathbb{R}^9$ + SVD] & 57.9410 & 76.6215 & \bfseries 98.2259 & 59.0928 & 40.0309 & 60.0058 & 93.2335 & 54.0153 & 178.4857 & 126.5280 & 129.9181 & 76.1896 & 87.3255 & 96.2916 & 61.0347 & 156.9399 & 127.0804 & 141.5543 \\
            Ours [ADoF] & 57.8207 & 76.3786 & 98.1622 & 60.2124 & 38.0459 & 58.2500 & 94.0080 & 52.8324 & 175.0730 & 125.9089 & 126.2338 & 71.1438 & 84.9072 & 96.2668 & 62.2706 & 162.1402 & 124.6930 & 132.2253 \\  
            \bottomrule[1.5pt]
        \end{tabular}%
    }
\end{table*}

\newcommand{\rowimgw}{0.125\textwidth}

\newcommand{\resultrow}[1]{%
  \includegraphics[width=\rowimgw]{figures/amr_regression_overview/reference_images/full_input_#1.jpg}%
  \includegraphics[width=\rowimgw]{figures/amr_regression_overview/WLDO/full_overlay_#1.jpg}%
  \includegraphics[width=\rowimgw]{figures/amr_regression_overview/BARC/full_overlay_#1.jpg}%
  \includegraphics[width=\rowimgw]{figures/amr_regression_overview/BITE/full_overlay_#1.jpg}%
  \includegraphics[width=\rowimgw]{figures/amr_regression_overview/GenZoo/full_overlay_#1.jpg}%
  \includegraphics[width=\rowimgw]{figures/amr_regression_overview/AniMer/full_overlay_#1.jpg}%
  \includegraphics[width=\rowimgw]{figures/amr_regression_overview/AniMerPlus/full_overlay_#1.jpg}%
  \includegraphics[width=\rowimgw]{figures/amr_regression_overview/PAWS/full_overlay_#1.jpg}%
}

\newcommand{\resultrowBITE}[1]{%
  \includegraphics[width=\rowimgw]{figures/amr_regression_overview_BITE/reference_images/full_input_#1.jpg}%
  \includegraphics[width=\rowimgw]{figures/amr_regression_overview_BITE/WLDO/full_overlay_#1.jpg}%
  \includegraphics[width=\rowimgw]{figures/amr_regression_overview_BITE/BARC/full_overlay_#1.jpg}%
  \includegraphics[width=\rowimgw]{figures/amr_regression_overview_BITE/BITE/full_overlay_#1.jpg}%
  \includegraphics[width=\rowimgw]{figures/amr_regression_overview_BITE/GenZoo/full_overlay_#1.jpg}%
  \includegraphics[width=\rowimgw]{figures/amr_regression_overview_BITE/AniMer/full_overlay_#1.jpg}%
  \includegraphics[width=\rowimgw]{figures/amr_regression_overview_BITE/AniMerPlus/full_overlay_#1.jpg}%
  \includegraphics[width=\rowimgw]{figures/amr_regression_overview_BITE/PAWS/full_overlay_#1.jpg}%
}

\begin{figure*}
    \centering

    \begin{minipage}{0.125\linewidth}\centering\textbf{Image}\end{minipage}%
    \begin{minipage}{0.125\linewidth}\centering\textbf{WLDO}\end{minipage}%
    \begin{minipage}{0.125\linewidth}\centering\textbf{BARC}\end{minipage}%
    \begin{minipage}{0.125\linewidth}\centering\textbf{BITE}\end{minipage}%
    \begin{minipage}{0.125\linewidth}\centering\textbf{GenZoo}\end{minipage}%
    \begin{minipage}{0.125\linewidth}\centering\textbf{AniMer}\end{minipage}%
    \begin{minipage}{0.125\linewidth}\centering\textbf{AniMer+}\end{minipage}%
    \begin{minipage}{0.125\linewidth}\centering\textbf{Ours}\end{minipage}%

    \resultrow{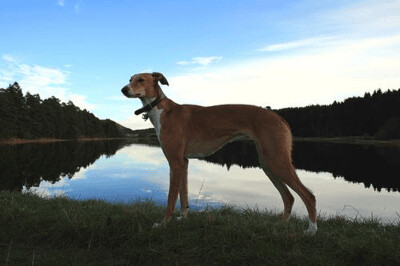}\vspace{-1pt}
    \resultrow{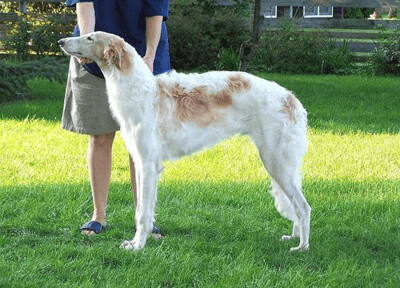}\vspace{-1pt}
    \resultrow{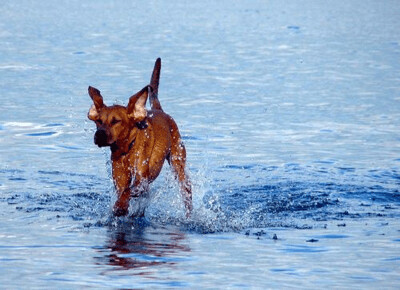}\vspace{-3pt}
    \resultrowBITE{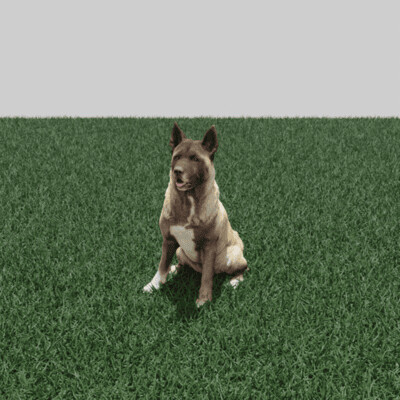}\vspace{-5mm}
    \resultrowBITE{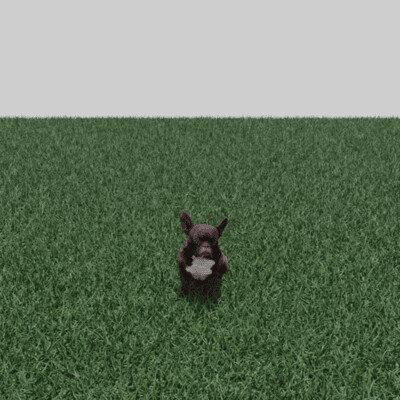}
    \caption{Qualitative comparison of monocular 3D dog reconstruction methods. Each column shows the reconstruction overlaid on the input image produced by a different approach. From left to right, WLDO~\cite{WLDO}, BARC~\cite{BARC}, BITE~\cite{BITE}, GenZoo~\cite{GenZoo}, AniMer~\cite{AniMer}, AniMer+~\cite{AniMerPlus}, and ours. The input image is shown in the first column. All methods regress from a bounding-box crop around the dog with the reconstruction transformed onto the full image. Images sourced from StanfordExtra~\cite{WLDO} (top three rows) and synthetic renders from BITE~\cite{BITE} (bottom two rows). Despite using only limited pose parameters and degrees of freedom, our model produces robustly posed dogs with anatomically plausible joint configurations that capture the observed pose faithfully. In particular, in the last two rows our model produces a plausible sitting configuration for challenging self-occluded dog appearances.}
    \label{figure:regression_results_overview}
\end{figure*}

\newpage
\section{Conclusion}
\label{section:conclusion}
In this work, we introduced PAWS, a parametric model of dog shape and pose. 
Unlike previous models, our model combines a shape space learned from real dogs with an anatomically faithful kinematic tree containing a total of 97 (including 3 for translating the model) degrees of freedom (DoFs) spread over 62 joints.
As presented in our experiments, limiting the degrees of freedom achieves regression accuracy comparable to unconstrained parameterizations while producing anatomically plausible poses.
Moreover, our model's shape space allows for accurate surface reconstruction of real dogs as presented during model fitting.
Despite being able to capture the overall body shapes of different dogs, accurately modelling nuanced and detailed facial expressions and finer ear deformations remains future work.
To better understand dogs' communicative cues, it is important to not only capture their body shape and pose but also subtle facial signals.
Future work could therefore investigate methods for capturing fine-grained facial deformations caused by facial muscle movements. 
Moreover, collecting a dataset with a wider range of breeds, dog sizes, movements and postures could further expand the diversity of the shape and pose space. 
\clearpage

\section*{Acknowledgments} 
We gratefully acknowledge access to the Berzelius computing resource at the National Supercomputer Centre, provided by the Knut and Alice Wallenberg Foundation. 
The 4D scanning was done with equipment owned by the Optics and Sensing Lab, Max Planck Institute for Intelligent Systems, Germany. 
This work was supported by the Swedish Research Council, project 2023-04693, the Swedish e-Science Research Centre (SeRC), and the Agria-SKK Research Fund, project N2025-0006.

{
    \small
    \bibliographystyle{ieeenat_fullname}
    \bibliography{main}
}
\setcounter{page}{1}
\maketitlesupplementary

\setcounter{section}{0}
\setcounter{figure}{0}
\setcounter{table}{0}
\setcounter{equation}{0}

\section{Introduction}
\begin{figure*}
    \centering

    \resizebox{0.8\textwidth}{!}{%
        \includegraphics[]{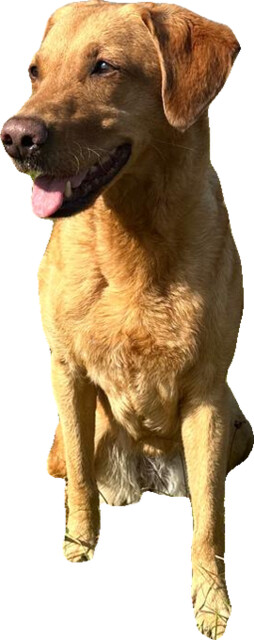}%
        \includegraphics[]{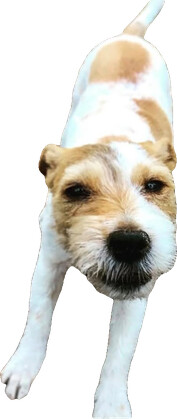}%
        \includegraphics[]{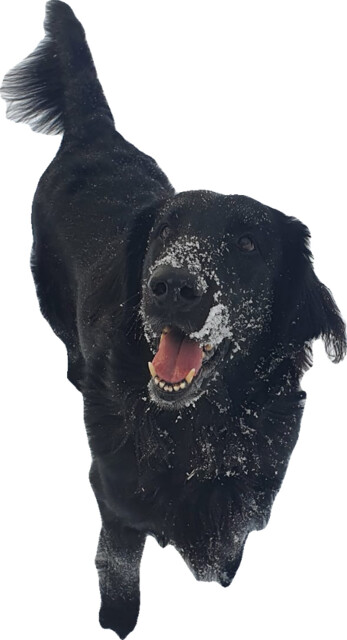}%
        \includegraphics[]{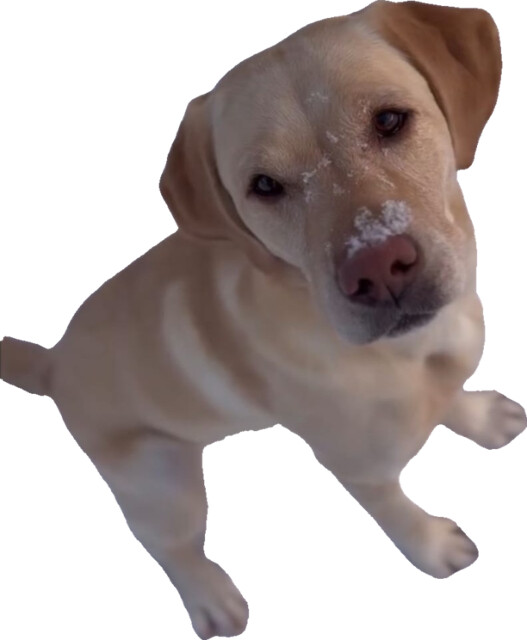}%
        \includegraphics[]{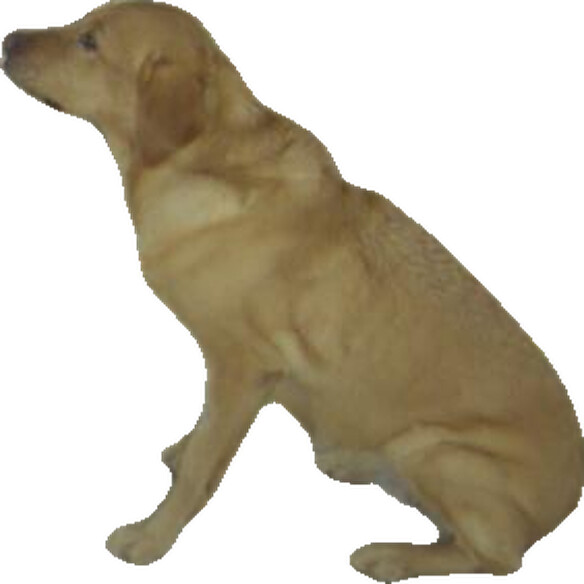}%
        \includegraphics[]{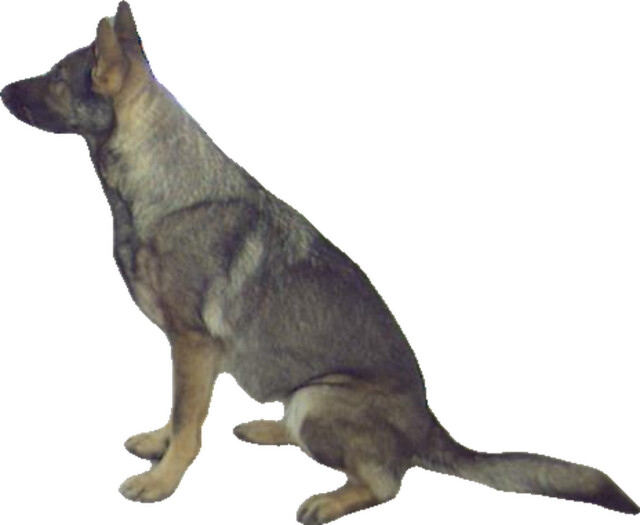}%
        \includegraphics[]{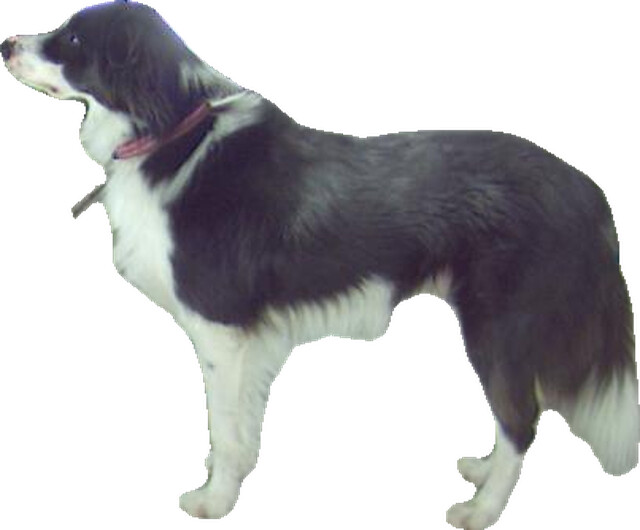}%
        \makebox[0.125\textwidth]{}
    }

    \resizebox{0.8\textwidth}{!}{%
        \includegraphics[]{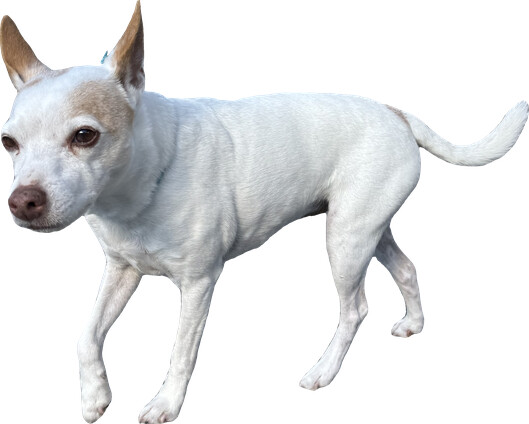}%
        \includegraphics[]{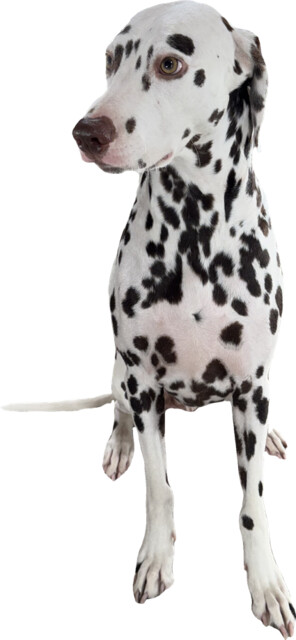}%
        \includegraphics[]{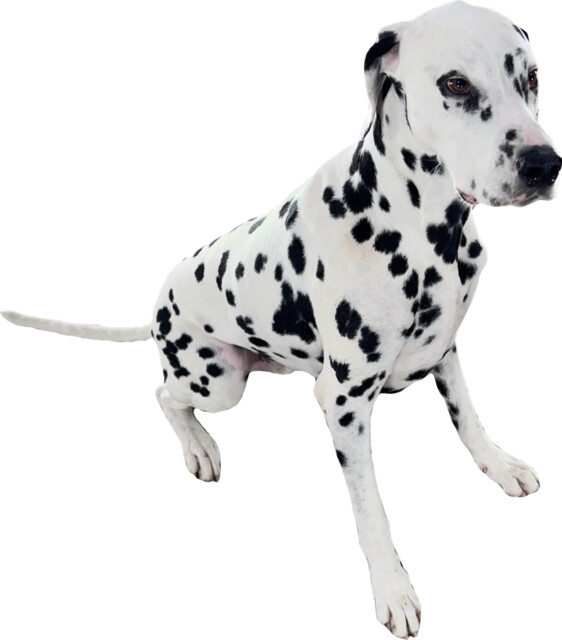}%
        \includegraphics[]{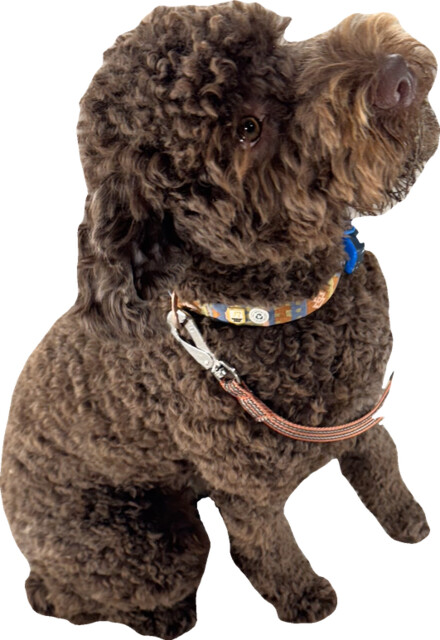}%
        \includegraphics[]{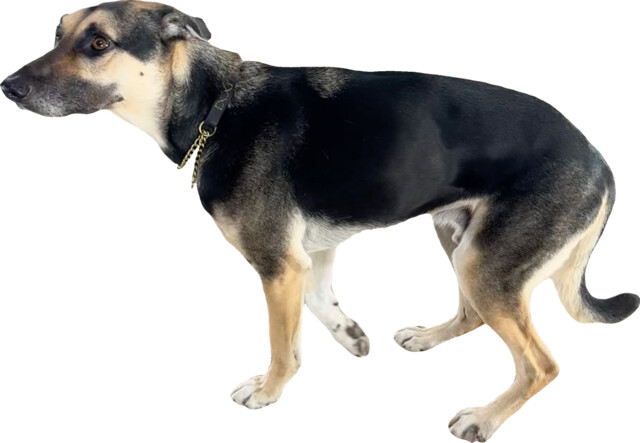}%
        \includegraphics[]{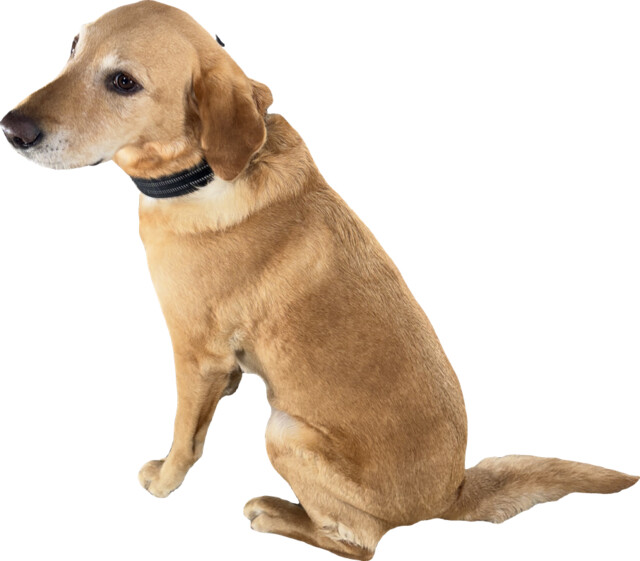}%
        \includegraphics[]{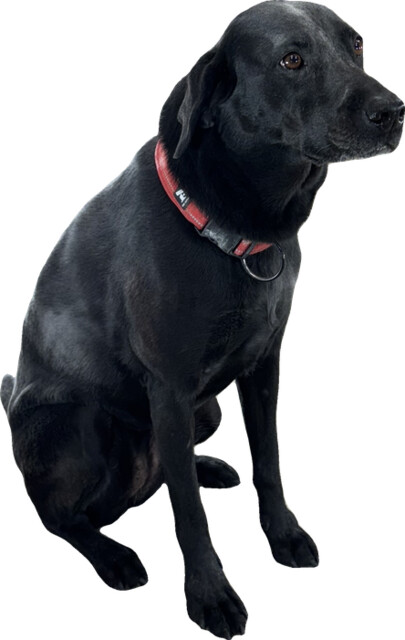}%
        \includegraphics[]{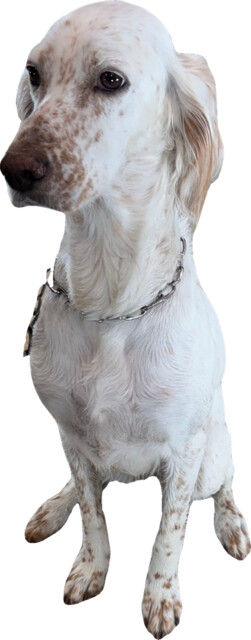}
    }

    \resizebox{0.8\textwidth}{!}{%
        \includegraphics[]{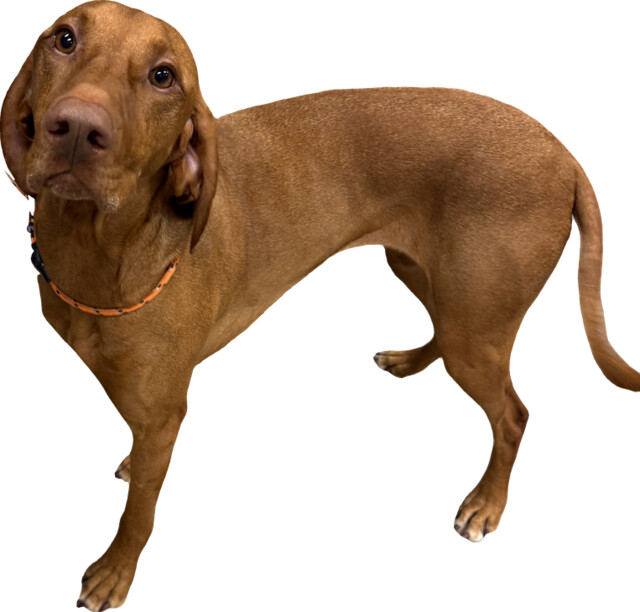}%
        \includegraphics[]{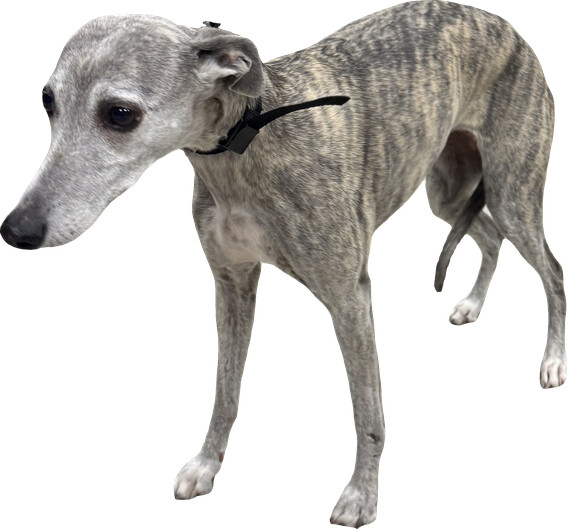}%
        \includegraphics[]{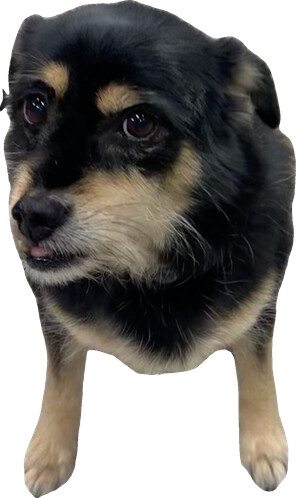}%
        \includegraphics[]{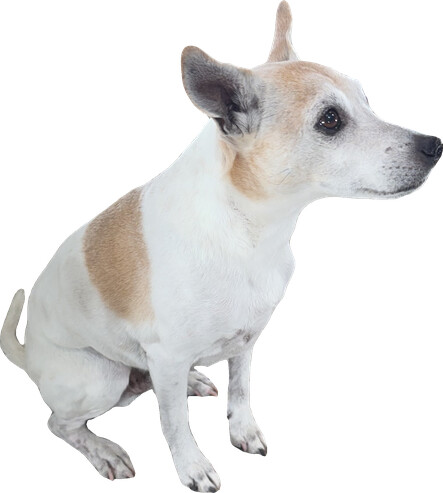}%
        \includegraphics[]{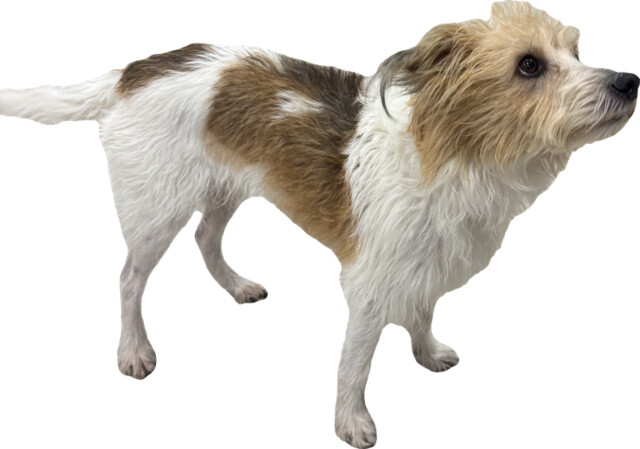}%
        \includegraphics[]{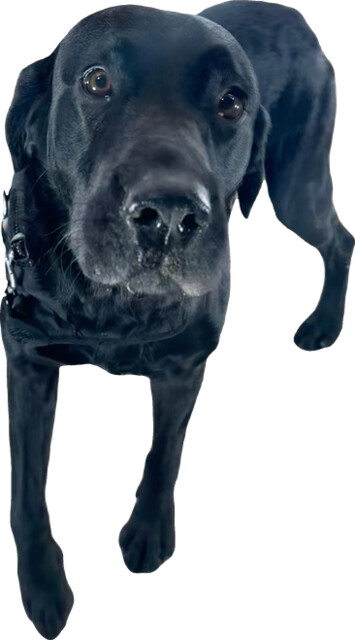}%
        \includegraphics[]{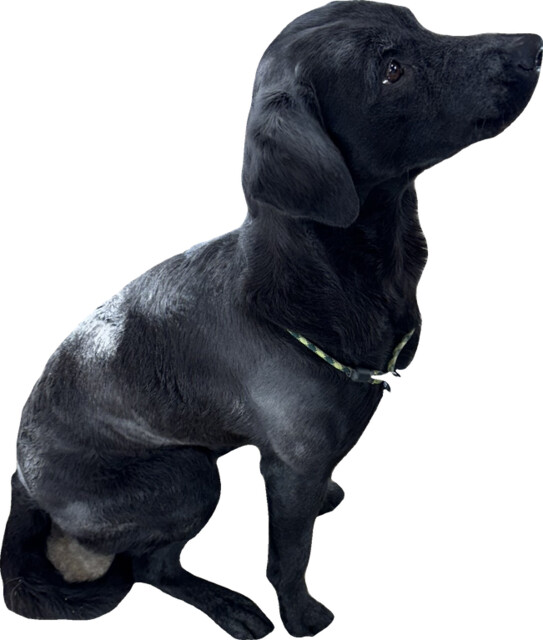}%
        \includegraphics[]{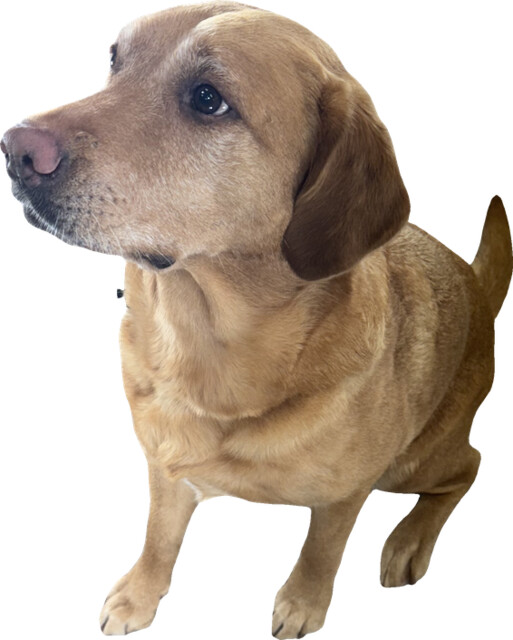}
    }

    \resizebox{0.8\textwidth}{!}{%
        \includegraphics[]{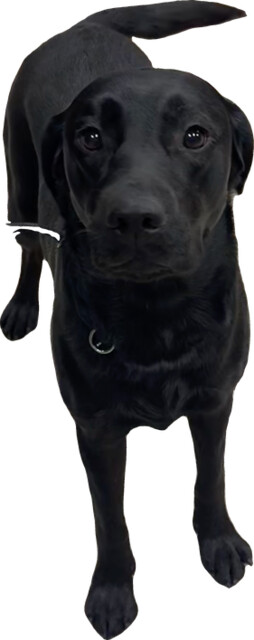}%
        \includegraphics[]{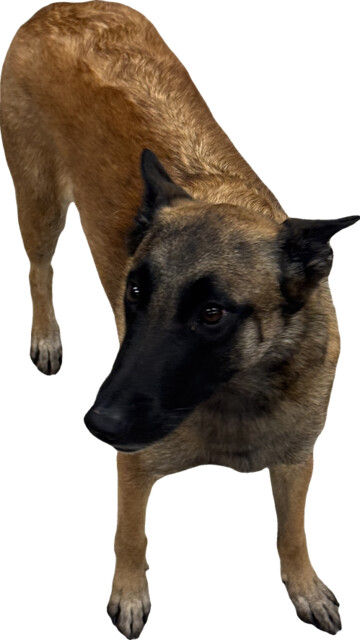}%
        \includegraphics[]{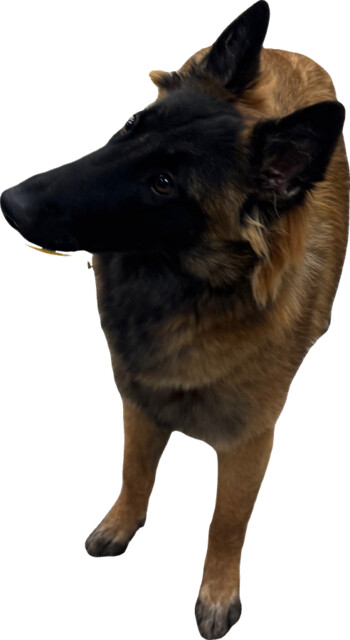}%
        \includegraphics[]{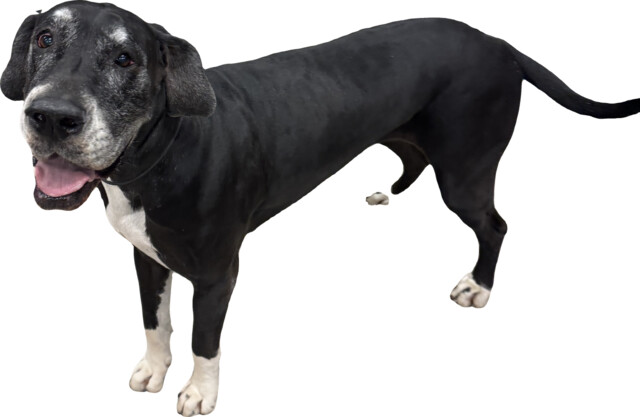}%
        \includegraphics[]{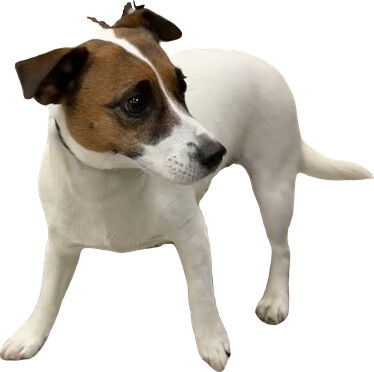}%
        \includegraphics[]{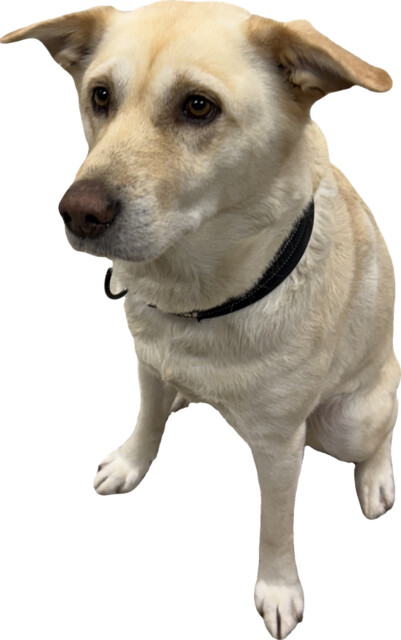}%
        \includegraphics[]{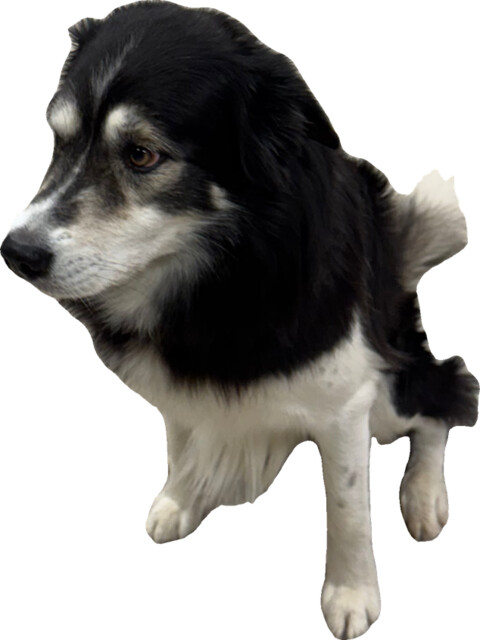}%
        \includegraphics[]{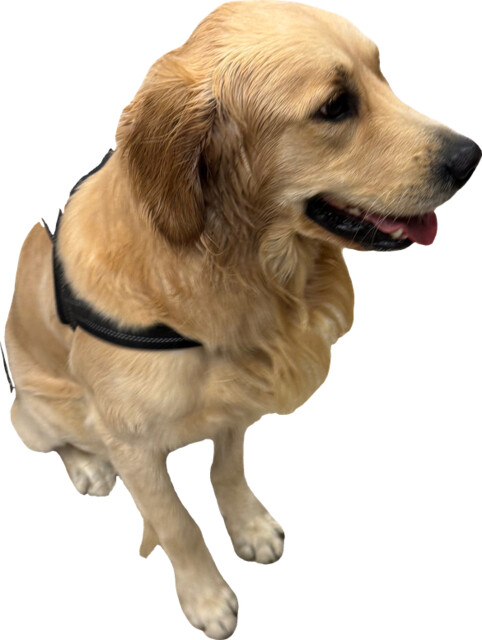}
    }
    \caption{Overview of the recorded dog subjects in our dataset. We present one representative picture for each of the 31 dog subjects.}
    \label{figure:subjects_overview}
\end{figure*}

\begin{figure*}[]
    \centering

    \resizebox{\textwidth}{!}{%
        \includegraphics[width=0.125\textwidth]{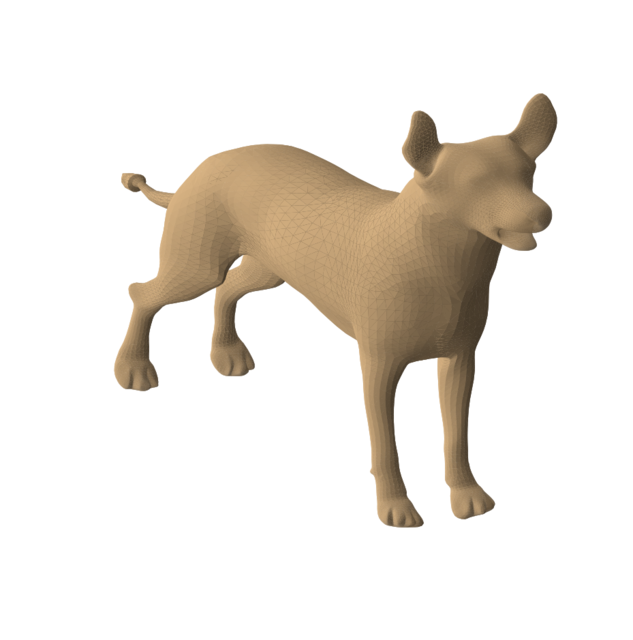}%
        \hspace{-4mm}%
        \includegraphics[width=0.125\textwidth]{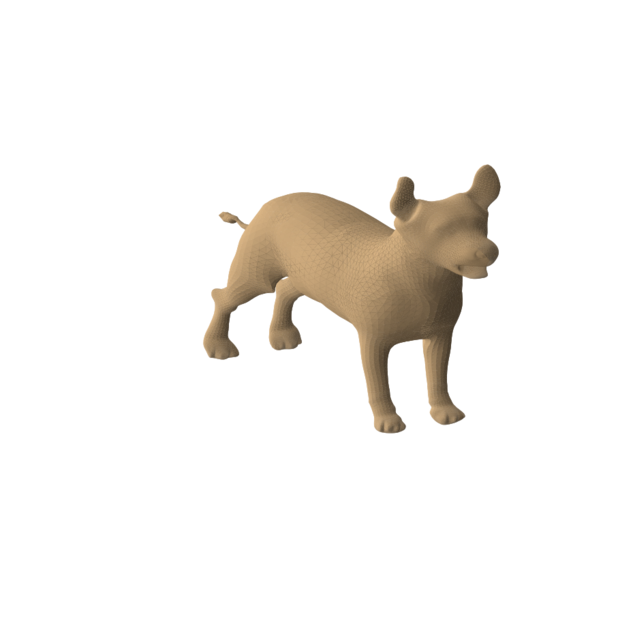}%
        \hspace{-4mm}%
        \includegraphics[width=0.125\textwidth]{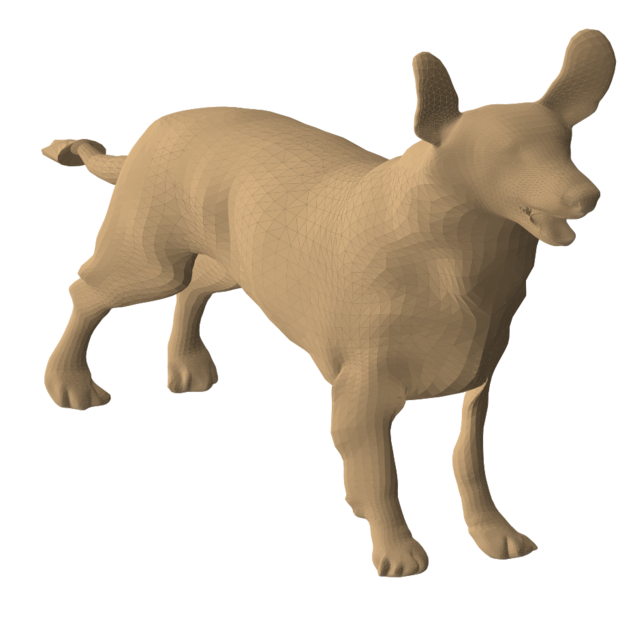}%
        \hspace{-4mm}%
        \includegraphics[width=0.125\textwidth]{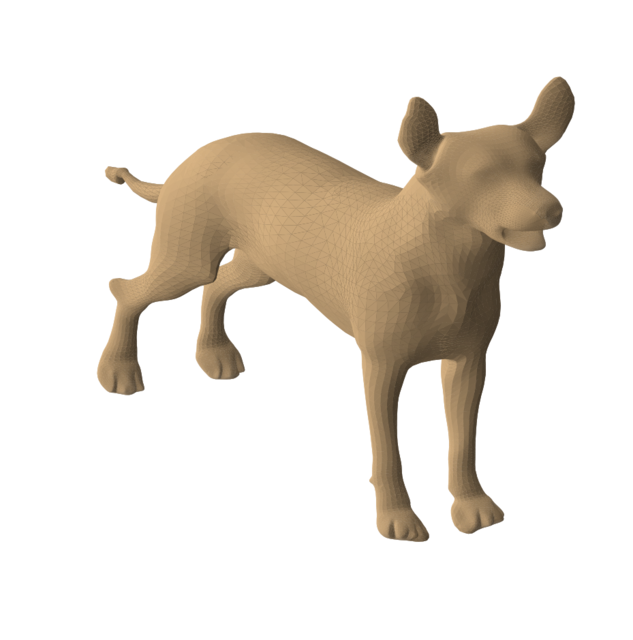}%
        \hspace{-4mm}%
        \includegraphics[width=0.125\textwidth]{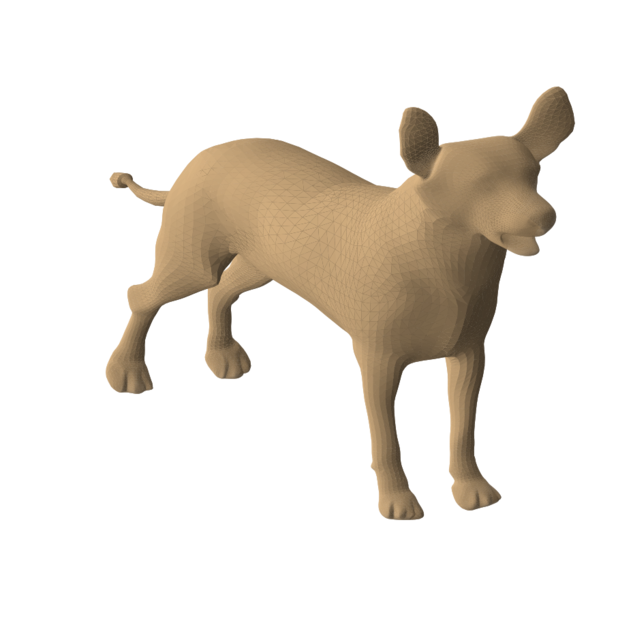}%
        \hspace{-4mm}%
        \includegraphics[width=0.125\textwidth]{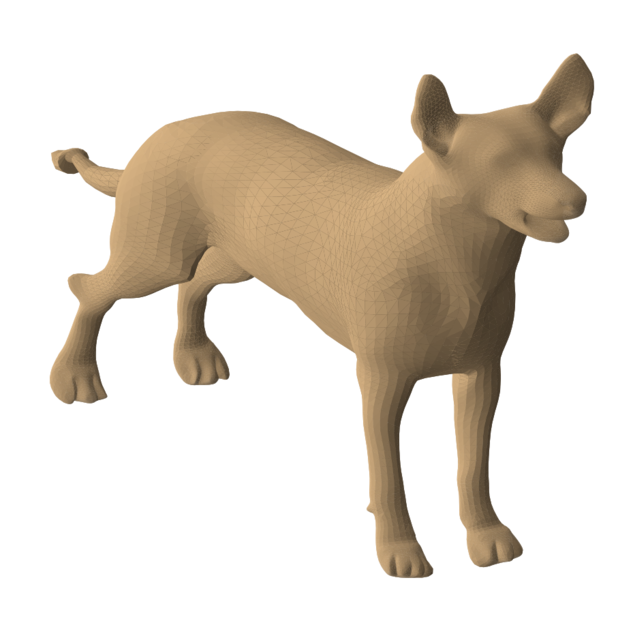}%
        \hspace{-4mm}%
        \includegraphics[width=0.125\textwidth]{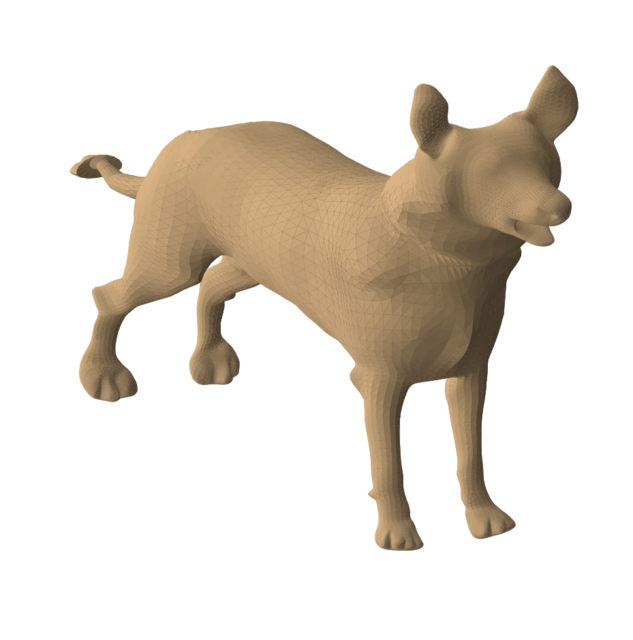}%
        \hspace{-4mm}%
        \makebox[0.125\textwidth]{}
    }

    \vspace{-8mm}
    \resizebox{\textwidth}{!}{%
        \includegraphics[width=0.125\textwidth]{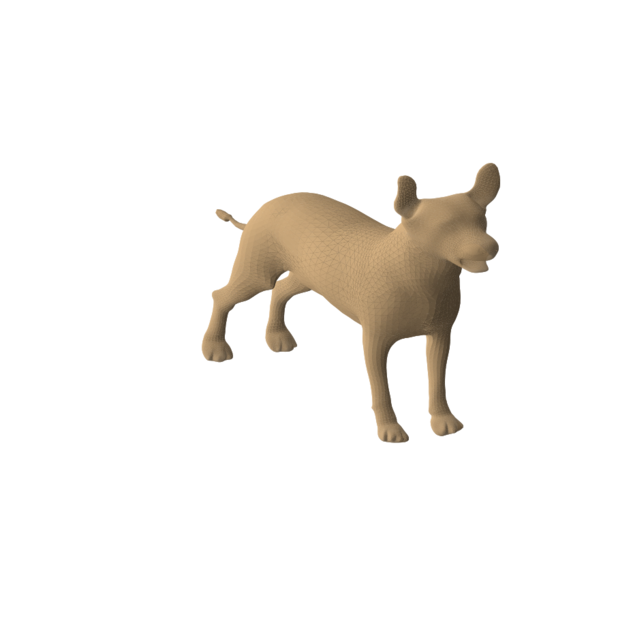}%
        \hspace{-4mm}%
        \includegraphics[width=0.125\textwidth]{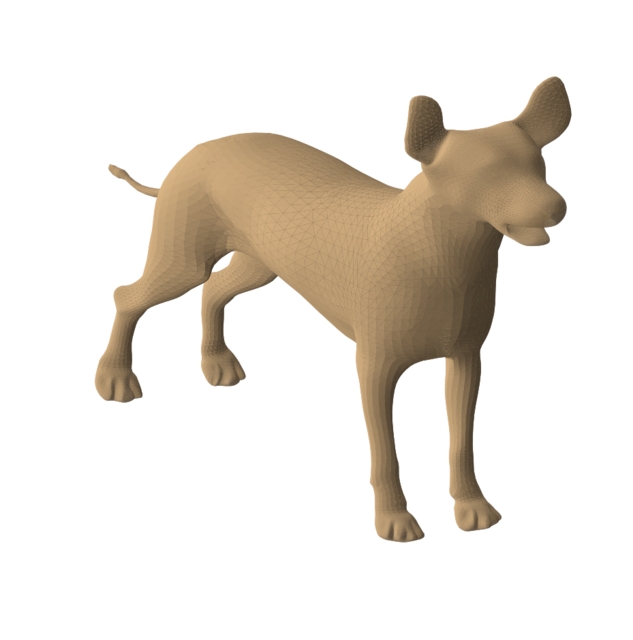}%
        \hspace{-4mm}%
        \includegraphics[width=0.125\textwidth]{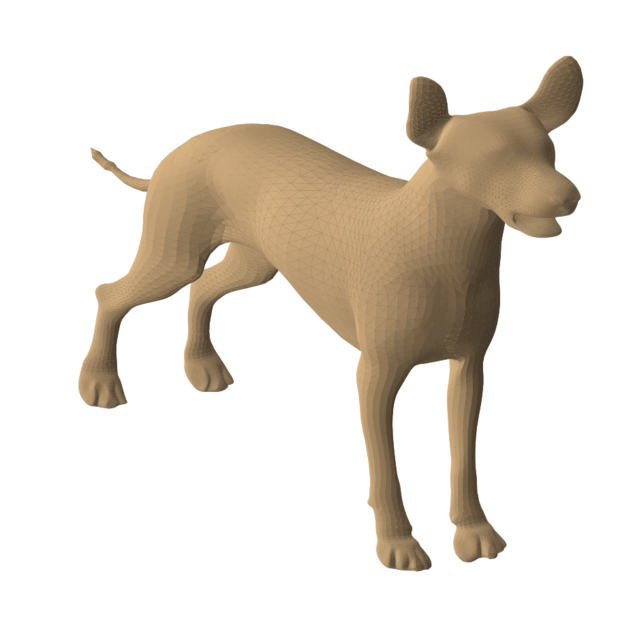}%
        \hspace{-4mm}%
        \includegraphics[width=0.125\textwidth]{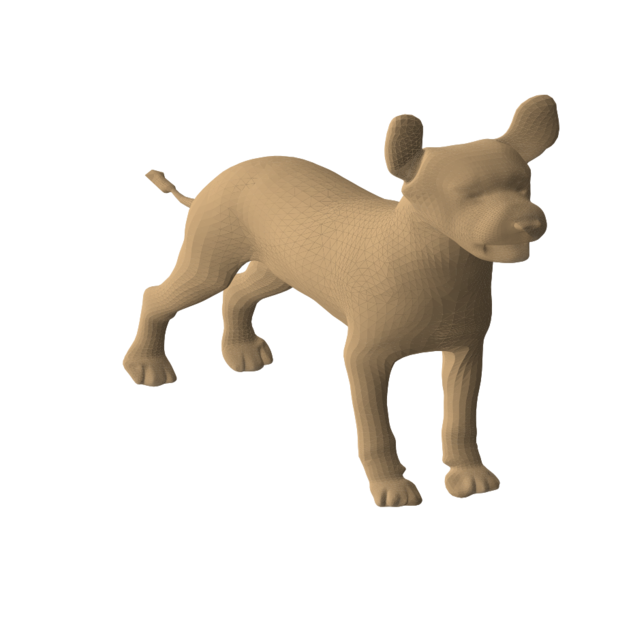}%
        \hspace{-4mm}%
        \includegraphics[width=0.125\textwidth]{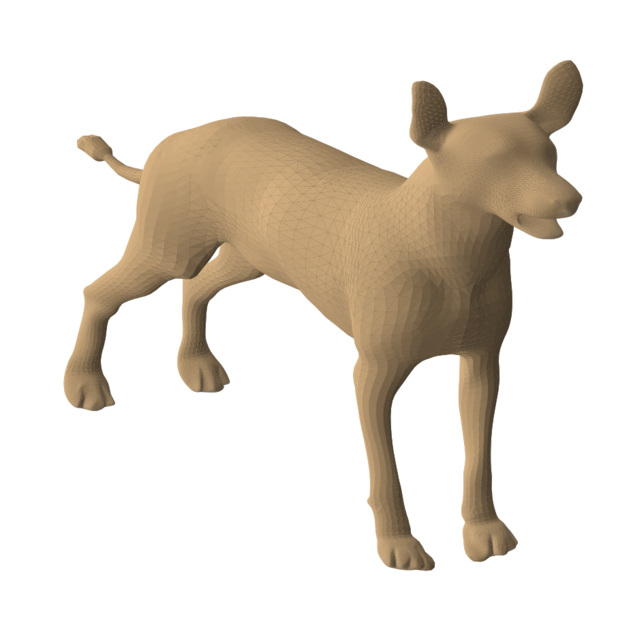}%
        \hspace{-4mm}%
        \includegraphics[width=0.125\textwidth]{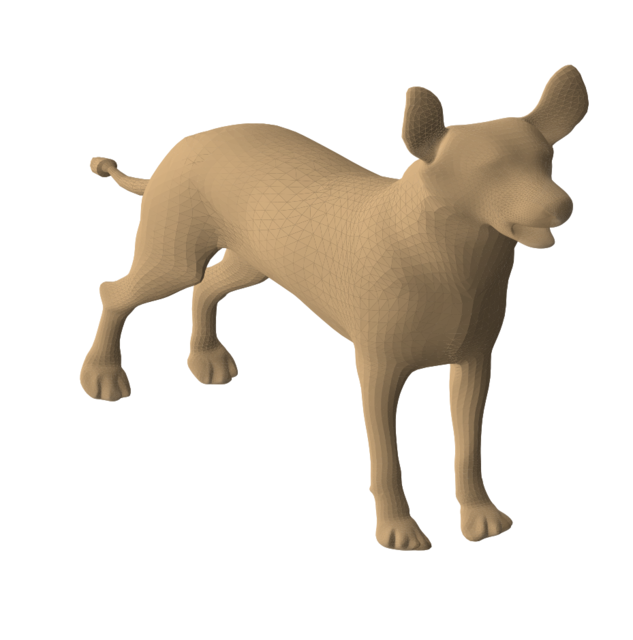}%
        \hspace{-4mm}%
        \includegraphics[width=0.125\textwidth]{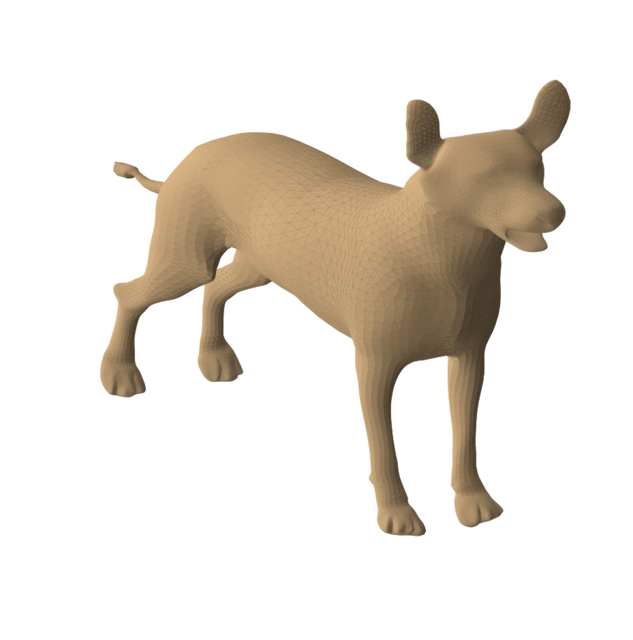}%
        \hspace{-4mm}%
        \includegraphics[width=0.125\textwidth]{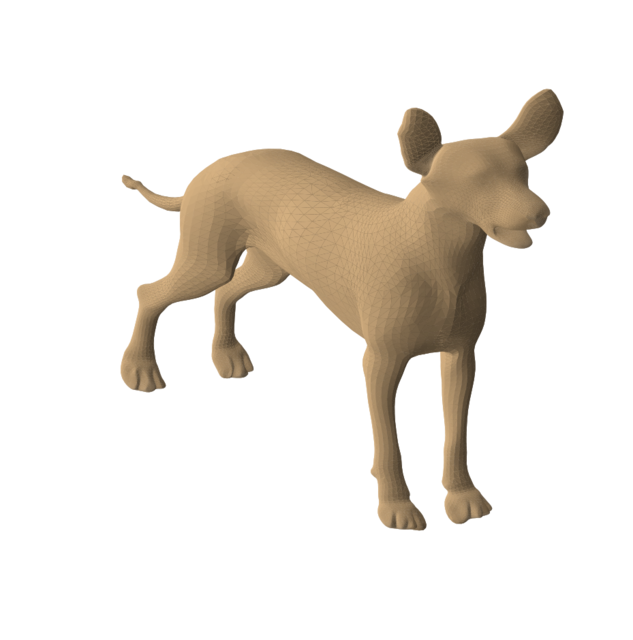}%
    }

    \vspace{-8mm}
    \resizebox{\textwidth}{!}{%
        \includegraphics[width=0.125\textwidth]{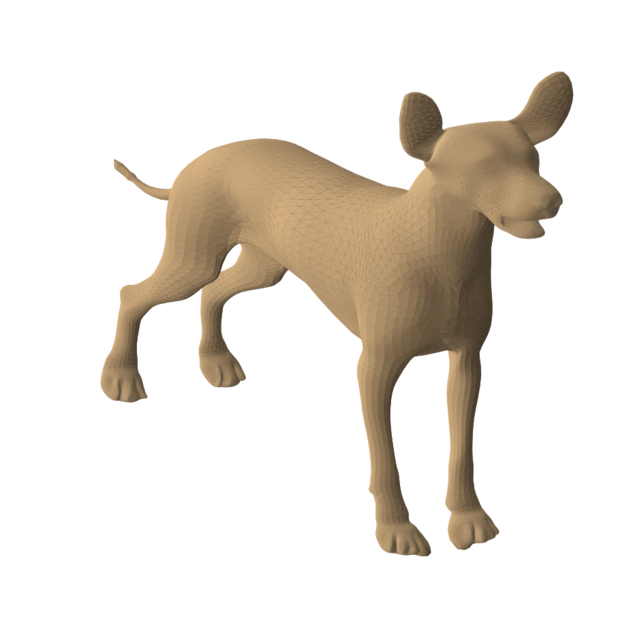}%
        \hspace{-4mm}%
        \includegraphics[width=0.125\textwidth]{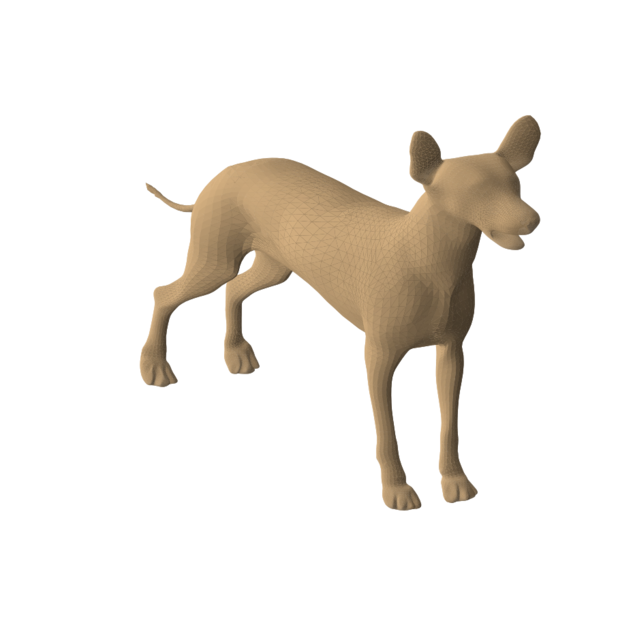}%
        \hspace{-4mm}%
        \includegraphics[width=0.125\textwidth]{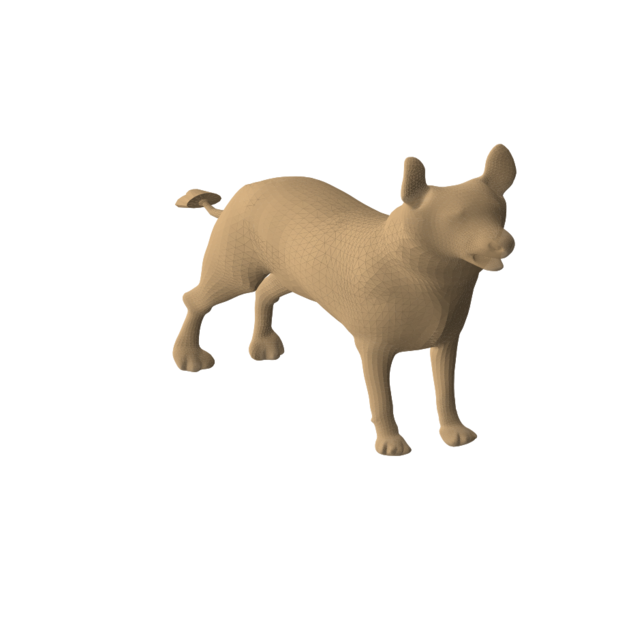}%
        \hspace{-4mm}%
        \includegraphics[width=0.125\textwidth]{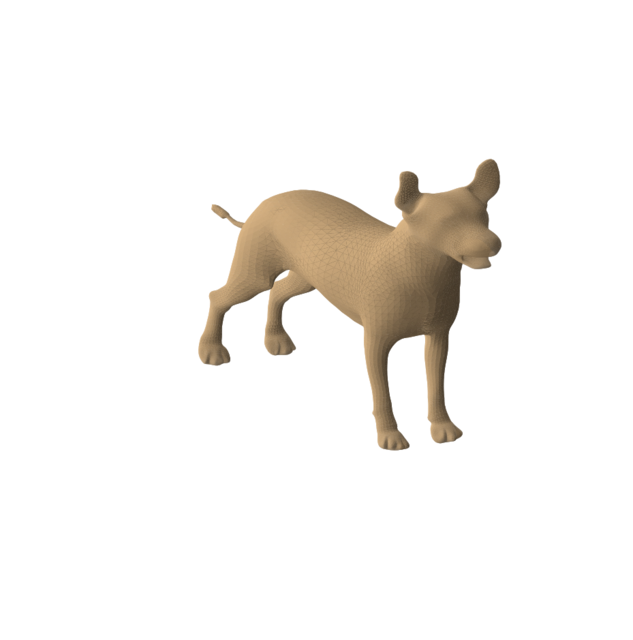}%
        \hspace{-4mm}%
        \includegraphics[width=0.125\textwidth]{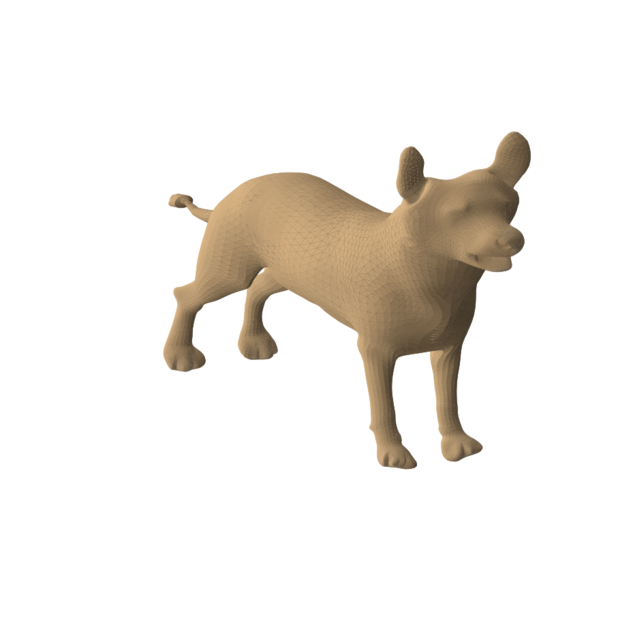}%
        \hspace{-4mm}%
        \includegraphics[width=0.125\textwidth]{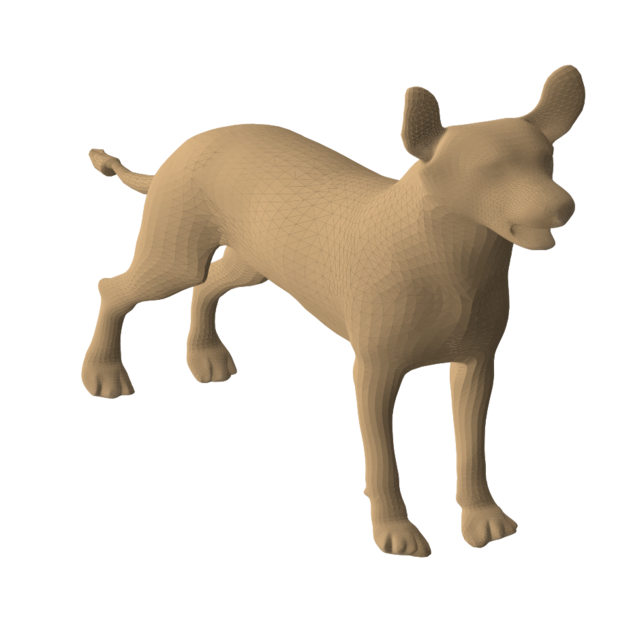}%
        \hspace{-4mm}%
        \includegraphics[width=0.125\textwidth]{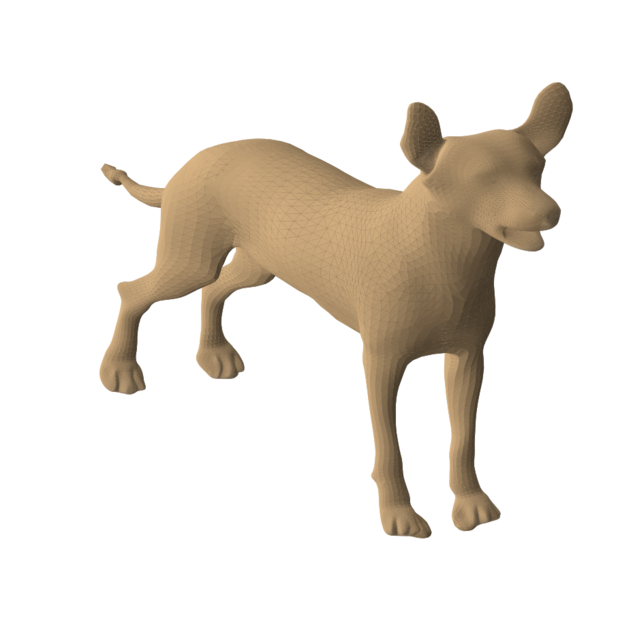}%
        \hspace{-4mm}%
        \includegraphics[width=0.125\textwidth]{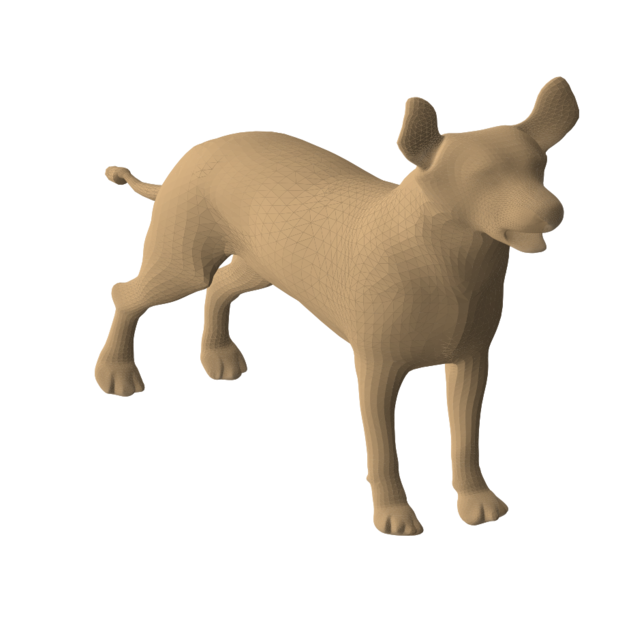}%
    }

    \vspace{-8mm}   
    \resizebox{\textwidth}{!}{%
        \includegraphics[width=0.125\textwidth]{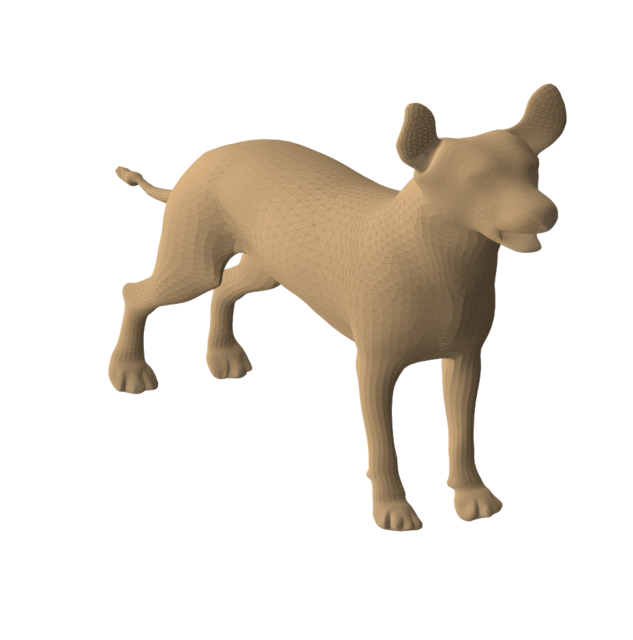}%
        \hspace{-4mm}%
        \includegraphics[width=0.125\textwidth]{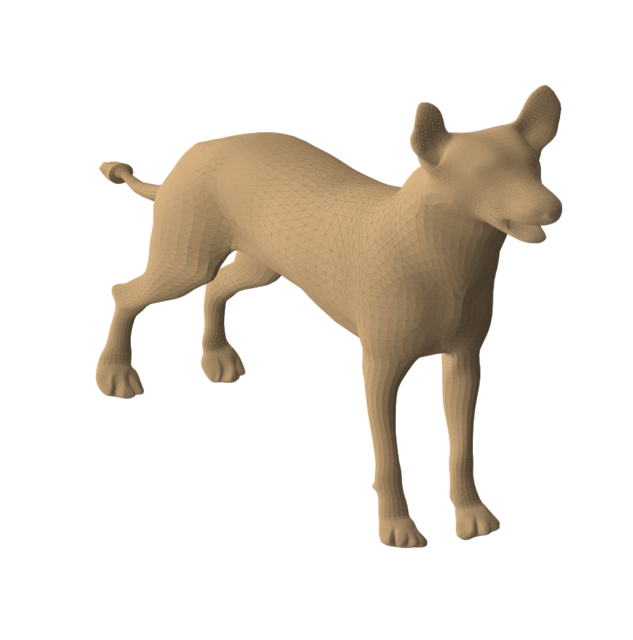}%
        \hspace{-4mm}%
        \includegraphics[width=0.125\textwidth]{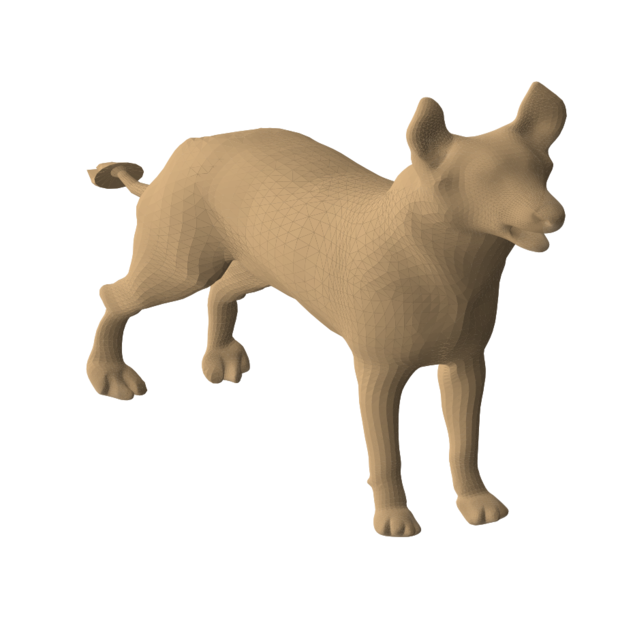}%
        \hspace{-4mm}%
        \includegraphics[width=0.125\textwidth]{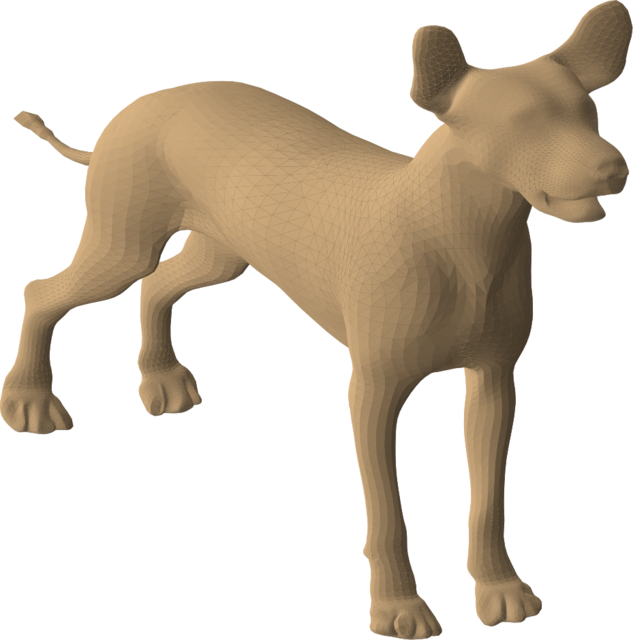}%
        \hspace{-4mm}%
        \includegraphics[width=0.125\textwidth]{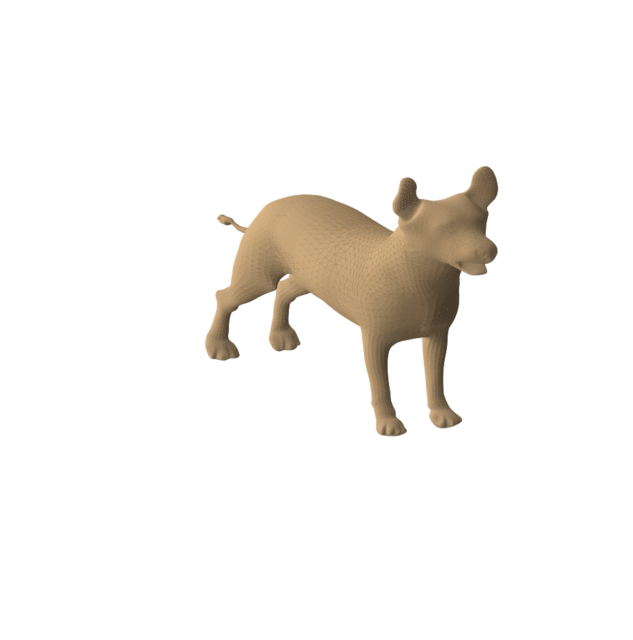}%
        \hspace{-4mm}%
        \includegraphics[width=0.125\textwidth]{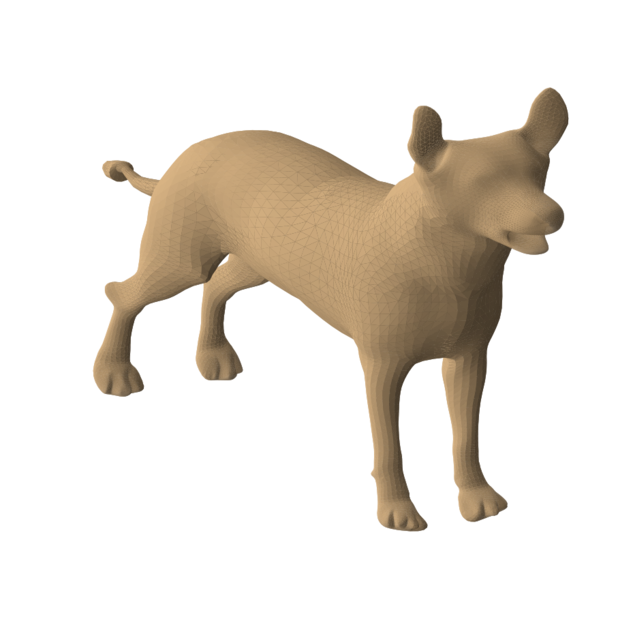}%
        \hspace{-4mm}%
        \includegraphics[width=0.125\textwidth]{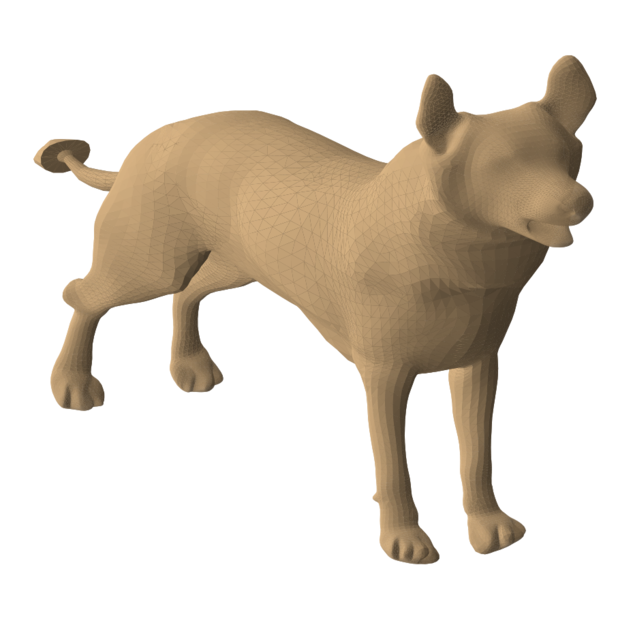}%
        \hspace{-4mm}%
        \includegraphics[width=0.125\textwidth]{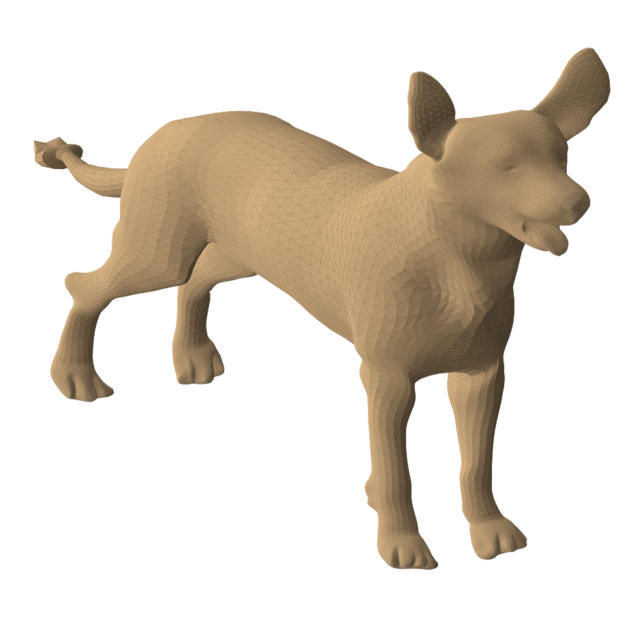}
    }

    \caption{Learned per-subject identity shapes. We show the recovered shape for each of the 31 individual dogs, as obtained through the registration pipeline. Each subject is visualized in its canonical pose.}
    \label{figure:subject_registrations_overview}
\end{figure*}

We present additional supplementary material in the appendix, covering implementation details and extended qualitative results.
First, we describe the registration pipeline in detail in \cref{section:appendix_registration}.
\cref{figure:subjects_overview} gives an overview of the recorded subjects, and \cref{figure:subject_registrations_overview} presents corresponding learned subject shapes from our registration pipeline.
In addition, to illustrate the range of our recorded dataset, covering both individual subject body shapes as well as pose variation, we present an overview of recorded scans for each subject in \cref{figure:per_subject_pose_variation_fittings}, with corresponding model fittings for BARC~\cite{BARC}, BITE~\cite{BITE}, and our model (LOOCV).
Finally, we describe the regression architecture in detail in \cref{section:appendix_regression}, and present supplementary qualitative results in \cref{figure:regression_results_overview_BITE,figure:regression_results_overview_StanfordExtra,figure:regression_results_overview_Animal3D} across different test datasets.

\section{Registration}
\label{section:appendix_registration}

\subsection{Implementation Details}
\begin{figure}[b]
    \begin{subfigure}{0.25\linewidth}
        \centering
        \includegraphics[width=\linewidth]{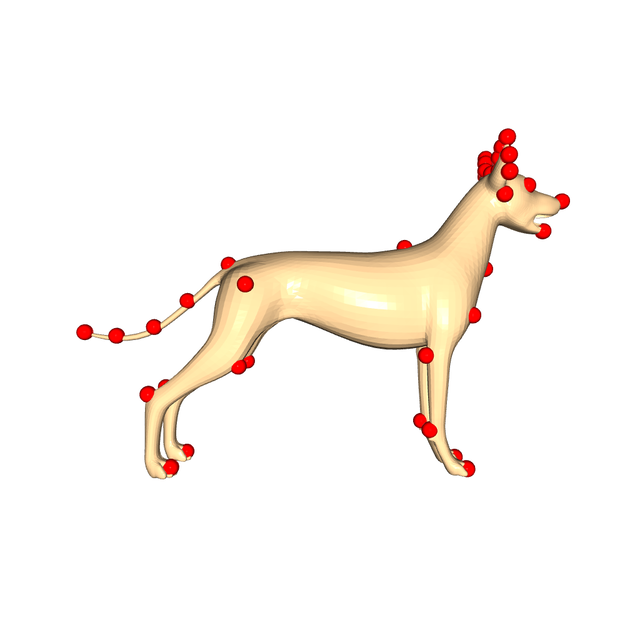}
        \vspace{-10mm}
    \end{subfigure}%
    \begin{subfigure}{0.25\linewidth}
        \centering
        \includegraphics[width=\linewidth]{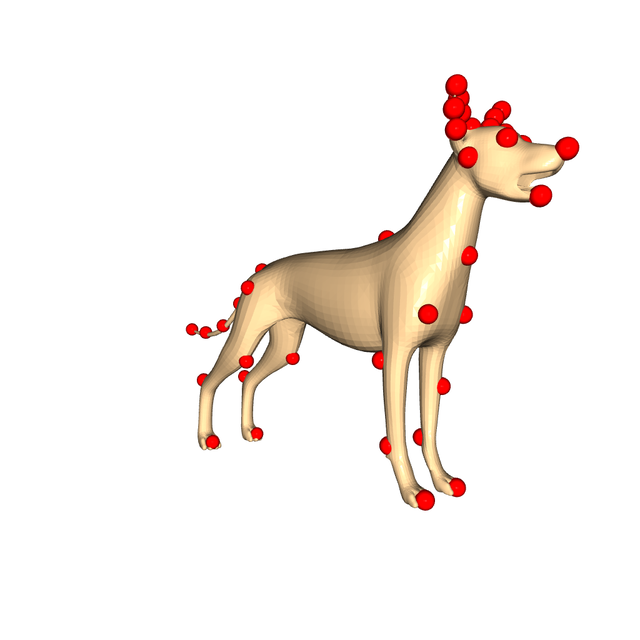}
        \vspace{-10mm}
    \end{subfigure}%
    \begin{subfigure}{0.25\linewidth}
        \centering
        \includegraphics[width=\linewidth]{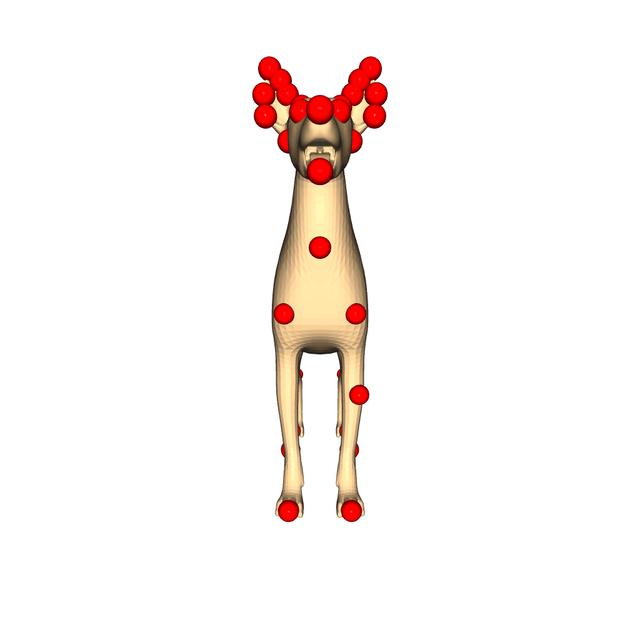}
        \vspace{-10mm}
    \end{subfigure}%
    \begin{subfigure}{0.25\linewidth}
        \centering
        \includegraphics[width=\linewidth]{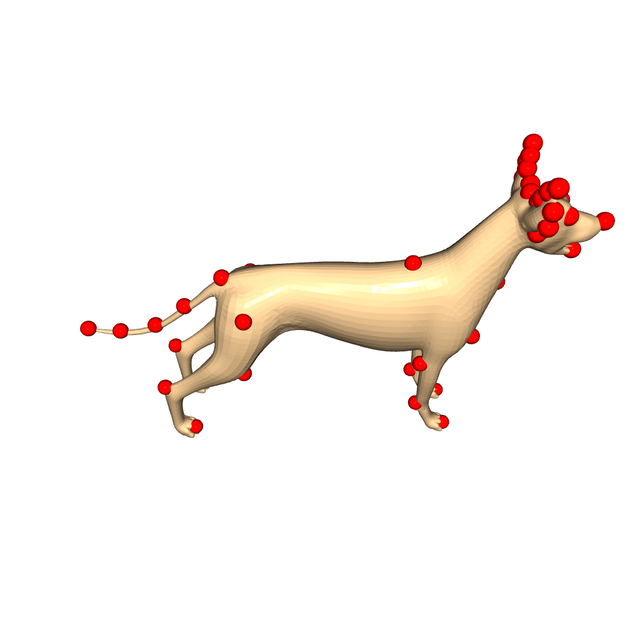}
        \vspace{-10mm}
    \end{subfigure}
    \caption{Overview of manually selected keypoint annotations on the PAWS template model from different views.}
    \label{figure:keypoint_annotations_paws}
\end{figure}

The registration pipeline is implemented in Python, using PyTorch for automatic differentiation. 
We optimize with Adam, using a fixed step-size of $5 \times 10^{-3}$ for initialization, pose and shape parameter optimization, and $10^{-4}$ for all subsequent refinement iterations.
Optimization terminates once the loss stabilizes.
Concretely, we monitor the last 128 iterations and declare convergence when the relative range falls below $5 \times 10^{-2}$.
Moreover, we sample with a density of $2^{14}$ points per square metre on both surfaces, when computing the scan-to-model and model-to-scan losses.
During optimization, we reduce the scale $\rho$ for the Geman-McClure~\cite{geman-mcclure-loss} iteratively from an initial 25cm to 2.5cm over the course of 12 steps, while alternating between pose and shape parameter optimization for four iterations at each scale.
After subject initialization, we include model parameter optimization, and alternate between pose, shape, and model optimization.
Finally, we refer to \cref{figure:keypoint_annotations_paws} for an overview of manually annotated keypoint locations on the PAWS model template that we rely on during registration.

\subsection{Loss Functions}
\label{subsection:appendix_registration_loss_functions}
In this section, we provide details on the implemented loss functions used during the registration.

\textbf{Keypoint Loss}:
The keypoint loss penalizes the squared Euclidean distance between each annotated keypoint and its corresponding landmark on the surface of the posed and shaped model instance.
In \cref{equation:appendix_keypoint_loss}, $N_{\text{kp}}$ denotes the number of keypoints, $v_j$ is an indicator of the visibility of keypoint $j$.
Analogously, $k_j$ refers to the 3D location of the target annotated keypoint and $M_{\text{Model}_{K_j}}$ refers to the corresponding location on the posed and shaped model instance.
\begin{align}
    \mathcal{L}_{\text{keypoint}} = \frac{1}{\sum_{j=1}^{N_{\text{kp}}} v_j} \sum_{j=1}^{N_{\text{kp}}} v_j \left\| M_{\text{Model}_{K_j}} - k_j\right\|^2_2
    \label{equation:appendix_keypoint_loss}
\end{align}

\textbf{Scan-to-Model Loss}:
The scan-to-model loss measures the distance from every sampled point on the surface of the scan to the closest point on the surface of the model.
In \cref{equation:appendix_scan_to_model_loss}, $\mathcal{S}$ denotes the surface of the corresponding scan and $M_{\text{Model}}$ denotes the posed and shaped model instance of the used parametric model.
\begin{align}
    \mathcal{L}_{\text{scan} \rightarrow \text{model}} &= \sum_{v_s \in \mathcal{S}} \rho \left( \min_{v_m \in M_{\text{PAWS}}} \| v_s - v_m \|_2^2 \right)
    \label{equation:appendix_scan_to_model_loss}
\end{align}

\textbf{Model-to-Scan Loss}:
The model-to-scan loss measures the distance from every sampled point on the surface of the model to the closest point on the surface of the scan.
In \cref{equation:appendix_model_to_scan_loss}, $\mathcal{S}$ denotes the surface of the corresponding scan and $M_{\text{Model}}$ denotes the posed and shaped model instance of the used parametric model.
We use the same sampling density to sample on the model as used before when computing the scan-to-model loss, to sample vertices on the surface of the model.
\begin{align}
    \mathcal{L}_{\text{model} \rightarrow \text{scan}} &= \sum_{v_m \in M_{\text{PAWS}}} \rho \left( \min_{v_s \in \mathcal{S}} \| v_s - v_m \|_2^2 \right)
    \label{equation:appendix_model_to_scan_loss}
\end{align}

\textbf{Joint Rotation Limits Loss}:
The joint rotation limits loss penalizes joint rotations that fall outside their anatomically plausible ranges.
For each joint, we measure the violation above the upper and below the lower limit, normalized by the joint's admissible range.
In \cref{equation:appendix_joint_limits_loss}, $N_j$ denotes the number of joints with $l_j$ and $u_j$ being a predefined lower and upper limit for joint $j$.     
\begin{align}
    \mathcal{L}_{\text{lim}} = \frac{1}{N_j} \sum_{i=1}^{N_j} \frac{\max(l_i-\theta_i, 0)^2 + \max(\theta_i-u_i, 0)^2}{(u_i-l_i)^2}
    \label{equation:appendix_joint_limits_loss}
\end{align}

\textbf{Anisotropic Offset}:
The anisotropic offset regularization penalizes the per-vertex offset, with a lower weight along the template's vertex normal.
We decompose each displacement $\Delta V^d_i$ into a normal (${\Delta V^d_i}^{\parallel}$) and tangential (${\Delta V^d_i}^{\perp}$) component based on the normal of the associated template vertex.
In \cref{equation:appendix_offset_regularization}, by setting $w_{\parallel}=0.25$ and $w_{\perp}=1.25$, we encourage the model to displace vertices along the surface normal rather than sliding within the surface of the template.
\begin{align}
    \mathcal{L}_{\text{aniso}} = \frac{1}{N_{\text{v}}} \sum_{i=1}^{N_{\text{v}}} w_{\perp} \left\| {\Delta V^d_i}^{\perp} \right\|_2^2 + w_{\parallel} \left\| {\Delta V^d_i}^{\parallel} \right\|_2^2
    \label{equation:appendix_offset_regularization}
\end{align}

\textbf{Laplacian Smoothing}:
The Laplacian smoothing term is used to encourage smooth deformations of the reconstructed model template by penalizing variations in the curvature of the template.
In \cref{equation:appendix_biharmonic_smoothing} $N_\text{v}$ denotes the number of vertices on the template mesh and $L$ is the cotangent Laplacian of the template mesh.
\begin{align}
    \mathcal{L}_{\text{lap}} = \frac{1}{N_{\text{v}}} \sum_{i=1}^{N_{\text{v}}} \left\| \left( L^2 V \right)_i \right\|_2^2
    \label{equation:appendix_biharmonic_smoothing}
\end{align}

\textbf{Normal Consistency}:
The normal consistency is used to encourage neighbouring vertices to have similar vertex normals.
In \cref{equation:appendix_normal_consistency} $N_e$ denotes the number of edges on the template mesh, with the sum indexing over edges with vertex endpoints $i$ and $j$.
Moreover, we use $n_i$ and $n_j$ to denote the normals of vertex $i$ and $j$.
\begin{align}
    \mathcal{L}_{\text{normal}} = \frac{1}{N_{\text{e}}} \sum_{(i, j)}^{N_{\text{e}}} \left( 1 - n_i \cdot n_j \right)
    \label{equation:appendix_normal_consistency}
\end{align}

\textbf{Skinning Weights Initialization}:
The skinning weights initialization term encourages the optimized skinning weights to remain close to their initialization.
In \cref{equation:appendix_skinning_init} $N_{\text{v}}$ denotes the number of vertices on the template mesh and $N_{\text{j}}$ denotes the number of joints on the kinematic tree. 
Moreover, $\mathcal{W}^{\text{init}}$ denotes the initialized skinning weights.
\begin{align}
    \mathcal{L}_{\mathcal{W}_{\text{init}}} = \frac{1}{N_{\text{v}} N_{\text{j}}} \sum_{i,j} \left\| \mathcal{W}_{ij} - \mathcal{W}^{\text{init}}_{ij} \right\|^2_2
    \label{equation:appendix_skinning_init}
\end{align}

\textbf{Skinning Weights Smoothness}:
The skinning weights smoothness term encourages skinning weights of neighbouring vertices to be similar. 
In \cref{equation:appendix_skinning_smoothness} $N_{\text{e}}$ denotes the number of edges on the template mesh, with the sum indexing over edges with vertex endpoints $i$ and $j$.
Moreover, $\mathcal{W}_i \in \mathbb{R}^{N_{\text{j}}}$ denotes the vector of skinning weights associated with vertex $i$.
\begin{align}
    \mathcal{L}_{\mathcal{W}_{\text{smooth}}} = \frac{1}{N_{\text{e}}} \sum_{(i,j)}^{N_{\text{e}}} \left\| \mathcal{W}_i - \mathcal{W}_j \right\|^2_2
    \label{equation:appendix_skinning_smoothness}
\end{align}

\textbf{Joint Regressor Initialization}:
We use the joint regressor initialization term to encourage the optimized joint regressor to regress joint locations that are close to the initialized anatomical joint locations.
In \cref{equation:appendix_joint_regressor_init} $N_{\text{j}}$ denotes the number of joints, and $\mathcal{J}_{\text{init}}$ the initialized joint regressor.
Moreover, we use $V$ to refer to the model vertices.
\begin{align}
    \mathcal{L}_{\mathcal{J}_{\text{init}}} = \frac{1}{N_{\text{j}}} \sum_{j=1}^{N_{\text{j}}} \left\| \left( \mathcal{J}_{\text{opt}} V \right)_j - \left( \mathcal{J}_{\text{init}} V \right)_j \right\|_2^2
    \label{equation:appendix_joint_regressor_init}
\end{align}

\textbf{Blend Pose Functions Smoothness}:
The blend pose functions smoothness term encourages the blend pose functions of neighbouring vertices to be similar.
In \cref{equation:appendix_blend_pose_smoothness} $N_{\text{e}}$ denotes the number of edges on the template mesh, with the sum indexing over edges with vertex endpoints $i$ and $j$.
Moreover, ${B_P}_i$ denotes the blend pose functions associated with vertex $i$.
\begin{align}
    \mathcal{L}_{{B_P}_{\text{smooth}}} = \frac{1}{N_{\text{e}}} \sum_{(i,j)}^{N_{\text{e}}} \left\| {B_P}_i - {B_P}_j \right\|^2_2
    \label{equation:appendix_blend_pose_smoothness}
\end{align}

\subsection{Evaluation}
We present additional supplementary qualitative evaluation results in \cref{figure:per_subject_pose_variation_fittings}. 
For each of the 31 individual dogs, we show five poses ordered by increasing pose difficulty.
We quantify pose difficulty as the average geodesic joint rotation distance from the identity pose, as estimated during the registration.

We additionally highlight selected poses with zoomed-in regions to illustrate cases where our model achieves a more accurate and robust fit to the reference scan. 
In particular, our model performs well in cases where scan ambiguity or holes in the scan provide no surface to constrain the poses naturally.
In summary, our model provides a more robust pose parameterization that limits the occurrence of artifacts such as joint twisting, which can be observed for BARC and BITE. 
Furthermore, by reducing the degrees of freedom to follow anatomical constraints, our approach avoids over-parameterized joint configurations and allows for stable and robust pose fitting to complex and challenging poses.
Finally, we also refer to \cref{figure:per_subject_pose_variation_fittings} for an overview of per-subject shape variation as well as pose variation captured in our dataset.

\label{section:appendix_registration_evaluation}

\newcommand{\posepage}[1]{
  \begin{adjustbox}{max width=\textwidth, max totalheight=0.92\textheight, center}
    \begin{varwidth}{2\textwidth}
      \foreach \subjA/\subjB in {#1}{%
        \posestripBARC{\subjA}\hspace{15mm}\posestripBARC{\subjB}%
        \par\vspace{-12mm}%
      }%
    \end{varwidth}
  \end{adjustbox}%
}

\newcommand{\posestrip}[1]{%
    \begin{adjustbox}{max width=0.5\textwidth}
        \begin{tabular}{@{}c@{\hspace{1mm}}c@{}}
            \rotatebox{90}{\makebox[0.1\textwidth][c]{\textbf{Scan}}} &
            \foreach \i in {0,...,4}{%
                \includegraphics[width=0.1\textwidth]{supplementary/figures/subject_pose_variation/#1/scan_render_000\i_view-045deg.png}%
                \hspace{-8mm}%
            }\\[-18mm]
            \rotatebox{90}{\makebox[0.1\textwidth][c]{\textbf{BARC}}} &
            \foreach \i in {0,...,4}{%
                \includegraphics[width=0.1\textwidth]{supplementary/figures/subject_pose_variation_BARC/#1/model_render_000\i_view-045deg.png}%
                \hspace{-8mm}%
            }\\[-18mm]
            \rotatebox{90}{\makebox[0.1\textwidth][c]{\textbf{BITE}}} &
            \foreach \i in {0,...,4}{%
                \includegraphics[width=0.1\textwidth]{supplementary/figures/subject_pose_variation_BITE/#1/model_render_000\i_view-045deg.png}%
                \hspace{-8mm}%
            }\\[-18mm]
            \rotatebox{90}{\makebox[0.1\textwidth][c]{\textbf{Ours}}} &
            \foreach \i in {0,...,4}{%
                \includegraphics[width=0.1\textwidth]{supplementary/figures/subject_pose_variation_PAWS_LOOCV/#1/model_render_000\i_view-045deg.png}%
                \hspace{-8mm}%
            }%
            \end{tabular}
    \end{adjustbox}%
}

\newcommand{\posestriphighlight}[4]{%
    \begin{adjustbox}{max width=0.5\textwidth}
        \begin{tabular}{@{}c@{\hspace{1mm}}c@{}}
            \rotatebox{90}{\makebox[0.1\textwidth][c]{\textbf{Scan}}} &
            \foreach \i in {0,...,4}{%
                \ifnum\i=#4
                    \begin{tikzpicture}[spy using outlines={rectangle, red!85!black, very thick, magnification=3, size=1.25cm, connect spies}]
                        \node[
                            anchor=south west,
                            inner sep=0pt
                        ] (im) at (0,0) {%
                            \includegraphics[width=0.1\textwidth]{supplementary/figures/subject_pose_variation/#1/scan_render_000\i_view-045deg.png}%
                        };

                        \coordinate (sp) at ($(im.south west)!#2!(im.south east)$);
                        \coordinate (sp) at ($(sp)!#3!(im.north west -| sp)$);
                        \spy on (sp) in node[anchor=north east] at ([yshift=-10mm]im.north east);
                    \end{tikzpicture}%
                \else
                    \includegraphics[width=0.1\textwidth]{supplementary/figures/subject_pose_variation/#1/scan_render_000\i_view-045deg.png}%
                \fi
                \hspace{-8mm}%
            }\\[-18mm]

            \rotatebox{90}{\makebox[0.1\textwidth][c]{\textbf{BARC}}} &
            \foreach \i in {0,...,4}{%
                \ifnum\i=#4
                    \begin{tikzpicture}[spy using outlines={rectangle, red!85!black, very thick, magnification=3, size=1.25cm, connect spies}]
                        \node[anchor=south west, inner sep=0pt] (im) at (0,0) {%
                            \includegraphics[width=0.1\textwidth]{supplementary/figures/subject_pose_variation_BARC/#1/model_render_000\i_view-045deg.png}%
                        };

                        \coordinate (sp) at ($(im.south west)!#2!(im.south east)$);
                        \coordinate (sp) at ($(sp)!#3!(im.north west -| sp)$);
                        \spy on (sp) in node[anchor=north east] at ([yshift=-10mm]im.north east);
                    \end{tikzpicture}%
                \else
                    \includegraphics[width=0.1\textwidth]{supplementary/figures/subject_pose_variation_BARC/#1/model_render_000\i_view-045deg.png}%
                \fi
                \hspace{-8mm}%
            }\\[-18mm]

            \rotatebox{90}{\makebox[0.1\textwidth][c]{\textbf{BITE}}} &
            \foreach \i in {0,...,4}{%
                \ifnum\i=#4
                    \begin{tikzpicture}[spy using outlines={rectangle, red!85!black, very thick, magnification=3, size=1.25cm, connect spies}]
                        \node[anchor=south west, inner sep=0pt] (im) at (0,0) {%
                            \includegraphics[width=0.1\textwidth]{supplementary/figures/subject_pose_variation_BITE/#1/model_render_000\i_view-045deg.png}%
                        };

                        \coordinate (sp) at ($(im.south west)!#2!(im.south east)$);
                        \coordinate (sp) at ($(sp)!#3!(im.north west -| sp)$);
                        \spy on (sp) in node[anchor=north east] at ([yshift=-10mm]im.north east);
                    \end{tikzpicture}%
                \else
                    \includegraphics[width=0.1\textwidth]{supplementary/figures/subject_pose_variation_BITE/#1/model_render_000\i_view-045deg.png}%
                \fi
                \hspace{-8mm}%
            }\\[-18mm]

            \rotatebox{90}{\makebox[0.1\textwidth][c]{\textbf{Ours}}} &
            \foreach \i in {0,...,4}{%
                \ifnum\i=#4
                    \begin{tikzpicture}[spy using outlines={rectangle, red!85!black, very thick, magnification=3, size=1.25cm, connect spies}]
                        \node[anchor=south west, inner sep=0pt] (im) at (0,0) {%
                            \includegraphics[width=0.1\textwidth]{supplementary/figures/subject_pose_variation_PAWS_LOOCV/#1/model_render_000\i_view-045deg.png}%
                        };

                        \coordinate (sp) at ($(im.south west)!#2!(im.south east)$);
                        \coordinate (sp) at ($(sp)!#3!(im.north west -| sp)$);
                        \spy on (sp) in node[anchor=north east] at ([yshift=-10mm]im.north east);
                    \end{tikzpicture}%
                \else
                    \includegraphics[width=0.1\textwidth]{supplementary/figures/subject_pose_variation_PAWS_LOOCV/#1/model_render_000\i_view-045deg.png}%
                \fi
                \hspace{-8mm}%
            }%
        \end{tabular}
    \end{adjustbox}%
}

\begin{figure*}[p]
    \centering
    \begin{adjustbox}{max width=\textwidth, max totalheight=0.9\textheight, center}
        \begin{varwidth}{2\textwidth}
            \posestrip{Alone}%
            \hspace{10mm}%
            \posestrip{Bess}%
            \par\vspace{10mm}%

            \posestriphighlight{Pixie}{0.5}{0.47}{4}%
            \hspace{10mm}%
            \posestriphighlight{Elkes}{0.625}{0.125}{4}
            \par\vspace{10mm}%

            \posestrip{doggy}
            \hspace{10mm}%
            \posestrip{black_dog}%
            \par\vspace{10mm}%

            \posestrip{Malin}
            \hspace{10mm}%
            \posestrip{001}%
            \par\vspace{10mm}%
        \end{varwidth}
    \end{adjustbox}
    \caption{Per-subject pose and shape variations. We present each subject in five different poses, ordered from left to right by increasing average geodesic joint rotation distance from the identity pose as estimated during the registration.} 
    \label{figure:per_subject_pose_variation_fittings}
\end{figure*}

\begin{figure*}[p]
    \ContinuedFloat
    \centering
    \begin{adjustbox}{max width=\textwidth, max totalheight=0.9\textheight, center}
        \begin{varwidth}{2\textwidth}
            \posestriphighlight{002}{0.25}{0.55}{0}
            \hspace{10mm}%
            \posestriphighlight{003}{0.25}{0.29}{4}%
            \par\vspace{10mm}%

            \posestriphighlight{004}{0.3}{0.55}{2}
            \hspace{10mm}%
            \posestriphighlight{005}{0.175}{0.475}{0}%
            \par\vspace{10mm}%

            \posestrip{006}
            \hspace{10mm}%
            \posestrip{007}%
            \par\vspace{10mm}%

            \posestriphighlight{008}{0.675}{0.175}{4}
            \hspace{10mm}%
            \posestriphighlight{009}{0.4}{0.6}{1}
            \par\vspace{10mm}%
        \end{varwidth}
    \end{adjustbox}
    \caption{Per-subject pose and shape variations. We present each subject in five different poses, ordered from left to right by increasing average geodesic joint rotation distance from the identity pose as estimated during the registration.} 
    \label{figure:per_subject_pose_variation_fittings}
\end{figure*}

\begin{figure*}[p]
    \ContinuedFloat
    \centering
    \begin{adjustbox}{max width=\textwidth, max totalheight=0.9\textheight, center}
        \begin{varwidth}{2\textwidth}
            \posestrip{010}
            \hspace{10mm}%
            \posestrip{011}
            \par\vspace{10mm}%

            \posestriphighlight{012}{0.45}{0.275}{0}
            \hspace{10mm}%
            \posestrip{013}%
            \par\vspace{10mm}%

            \posestrip{014}
            \hspace{10mm}%
            \posestriphighlight{015}{0.3}{0.275}{3}
            \par\vspace{10mm}%

            \posestriphighlight{016}{0.675}{0.325}{2}
            \hspace{10mm}%
            \posestriphighlight{017}{0.675}{0.3}{1}
            \par\vspace{10mm}%
        \end{varwidth}
    \end{adjustbox}
    \caption{Per-subject pose and shape variations. We present each subject in five different poses, ordered from left to right by increasing average geodesic joint rotation distance from the identity pose as estimated during the registration.} 
    \label{figure:per_subject_pose_variation_fittings}
\end{figure*}

\begin{figure*}[p]
    \ContinuedFloat
    \centering
    \begin{adjustbox}{max width=\textwidth, max totalheight=0.9\textheight, center}
        \begin{varwidth}{2\textwidth}
            \posestrip{018}
            \hspace{10mm}%
            \posestriphighlight{019}{0.375}{0.225}{4}%
            \par\vspace{10mm}%

            \posestrip{020}
            \hspace{10mm}%
            \posestrip{021}%
            \par\vspace{10mm}%

            \posestrip{022}
            \hspace{10mm}%
            \posestrip{023}%
            \par\vspace{10mm}%

            \posestrip{024}
        \end{varwidth}
    \end{adjustbox}
    \caption{Per-subject pose and shape variations. We present each subject in five different poses, ordered from left to right by increasing average geodesic joint rotation distance from the identity pose as estimated during the registration.} 
    \label{figure:per_subject_pose_variation_fittings}
\end{figure*}

\newpage
\section{Regression}
\label{section:appendix_regression}
\subsection{Implementation Details}
Our implementation follows the architecture as described in the original HMR paper~\cite{HMR}.
We use a ResNet-50 V1.5 architecture to encode images and extract features. 
The ResNet is initialized with pretrained weights from \texttt{torchvision} trained on ImageNet.
Moreover, the encoder produces 2048-dimensional feature vectors for each image.
From these features, the architecture predicts the pose $\theta$, shape $\beta$ and translation parameters.
Finally, we use $T=3$ for the iterative regressor, allowing the network to iteratively refine its prediction.
The shape and pose parameters are regularized at all $T$ steps during training.
We optimize parameters using the AdamW optimizer with a learning rate of $1 \times 10^{-4}$ and $5 \times 10^{-4}$ for the encoder and iterative regressor respectively, and a weight decay of $1 \times 10^{-4}$ for both.
We linearly warm up the learning rate over the first four epochs (from a factor of 0.1) and subsequently use cosine annealing decay to reduce the learning rate over 252 epochs, after the linear warmup.
Moreover, we use a batch size of 64 and draw 384 batches per epoch.
Given that we use a mixture of datasets during training, we sample with replacement using fixed per-dataset weights.
We use the following weights during sampling: PAWSDogs: $0.4$, StanfordExtra: $0.25$, AnimalPose: $0.15$, Animal3D: $0.1$, and CtrlAni3D: $0.1$.

Lastly, we use the following augmentations during training to increase the effective dataset size and improve the robustness of the model.
First, we take a square crop centred on the annotated bounding box, where the size of the square is equal to the longer side of the annotated bounding box. 
Moreover, we randomly scale the crop by a randomly drawn factor between $0.8$ and $1.75$ to account for different zoomed image crops, and jitter the centre of the crop by up to $25\%$.
In addition, we apply random horizontal flipping (with 50\%), random photometric distortion (with 50\%), and random grayscale mapping (with 5\%). 
To improve robustness to varying image resolution, we additionally downscale the crop to a random size between 256 and 768 pixels before upscaling it back to the original size (with 30\%).
We apply Gaussian blur (with 10\%), and apply JPEG compression (with 50\% and quality between 30\% and 90\%).
We normalize the images using the ImageNet mean and standard deviation, before applying random erasing (with 40\% and scale between 2\% and 15\%).
Lastly, we resize the crop to $224 \times 224$ pixels.

The training loss is made up of direct parametric model parameter supervision, including the pose, shape and translation parameters, as well as 2D keypoint and joint reprojection and reference aligned 3D losses.
For the model parameter terms we use mean-squared error terms between the predicted values and the ground-truth annotations, averaged over all $T$ refinement iterations of the iterative regressor.
The 2D reprojection terms utilize a smooth $L_1$ loss between the ground-truth and the corresponding predicted and reprojected model vertices.
Finally, the 3D terms utilize a smooth $L_1$ loss between the ground-truth annotations and corresponding predicted model vertices, after aligning both to a common coordinate frame by removing the global translation.
We weight the individual loss terms as follows: pose parameters $10$, shape parameters $1$, and translation parameters $1$. 
For the 2D reprojection losses (on keypoints and joints) we use $0.05$, whereas we use a weight of $1$ for the reference aligned 3D keypoint and joint losses.

\subsection{Datasets}
We use a collection of different datasets for training and evaluation of our regression architecture.
We focus on datasets of quadrupeds and dogs in particular.
Moreover, the datasets provide different sets of annotations for both real and synthetic input images.
Dataset annotations span parametric pose and shape annotations, translation parameters, as well as 2D and 3D keypoints, silhouette masks, and camera parameters.

\noindent\textbf{StanfordExtra}~\cite{WLDO}: StanfordExtra extends the StanfordDogs~\cite{StanfordDogs} dataset, totalling 6078 training images and 1553 test images.
The dataset contains a diverse range of dog breeds captured in natural environments and natural poses. 
We use the 24 annotated keypoints to supervise our regression architecture.

\noindent\textbf{Animal3D}~\cite{Animal3D}: From the Animal3D dataset we utilize 3034 images for training and reserve 320 images for testing.
The dataset provides pose and shape parameter annotations for the SMAL model, in addition to keypoint and joint location annotations.
We use the 26 annotated keypoints to supervise our regression architecture, combining images from different quadrupeds.

\noindent\textbf{CtrlAni3D}~\cite{AniMer}: The CtrlAni3D dataset is a synthetic dataset of different quadrupeds.
In total, we utilize a subset of 6621 images for training, while retaining 1434 images for testing.
Annotations for the dataset include pixel-aligned pose and shape parameters for the SMAL model, in addition to extracted keypoint and joint locations.
From the dataset we use the keypoint and joint annotations to supervise our regression architecture.

\textbf{AnimalPose}~\cite{AnimalPoseDataset}: From the AnimalPose dataset we utilize a subset of 4893 images for training, while retaining 612 images for testing.
We supervise our architecture using the keypoint annotations.

\textbf{PAWSDogs}: From our synthetic renders of natural dog appearance we use 11059 images for training, retaining the remaining 1229 for validation and testing.
Given the aligned ground-truth pose and shape parameters, we supervise our model with the parametric model parameters directly.

\subsection{Evaluation}
\begin{figure*}
    \centering

    \newcommand{\rowimgWIDTH}{0.125\textwidth}
    \newcommand{\resultrowBITEAppendix}[1]{%
        \includegraphics[width=0.125\textwidth]{supplementary/figures/AMR_regression_BITE/reference_images/input_#1.jpg}%
        \includegraphics[width=0.125\textwidth]{supplementary/figures/AMR_regression_BITE/WLDO/overlay_#1.jpg}%
        \includegraphics[width=0.125\textwidth]{supplementary/figures/AMR_regression_BITE/BARC/overlay_#1.jpg}%
        \includegraphics[width=0.125\textwidth]{supplementary/figures/AMR_regression_BITE/BITE/overlay_#1.jpg}%
        \includegraphics[width=0.125\textwidth]{supplementary/figures/AMR_regression_BITE/GenZoo/overlay_#1.jpg}%
        \includegraphics[width=0.125\textwidth]{supplementary/figures/AMR_regression_BITE/AniMer/overlay_#1.jpg}%
        \includegraphics[width=0.125\textwidth]{supplementary/figures/AMR_regression_BITE/AniMerPlus/overlay_#1.jpg}%
        \includegraphics[width=0.125\textwidth]{supplementary/figures/AMR_regression_BITE/PAWS/overlay_#1.jpg}%
    }

    \begin{minipage}{0.125\linewidth}\centering\textbf{Image}\end{minipage}%
    \begin{minipage}{0.125\linewidth}\centering\textbf{WLDO}\end{minipage}%
    \begin{minipage}{0.125\linewidth}\centering\textbf{BARC}\end{minipage}%
    \begin{minipage}{0.125\linewidth}\centering\textbf{BITE}\end{minipage}%
    \begin{minipage}{0.125\linewidth}\centering\textbf{GenZoo}\end{minipage}%
    \begin{minipage}{0.125\linewidth}\centering\textbf{AniMer}\end{minipage}%
    \begin{minipage}{0.125\linewidth}\centering\textbf{AniMer+}\end{minipage}%
    \begin{minipage}{0.125\linewidth}\centering\textbf{Ours}\end{minipage}%

    \resultrowBITEAppendix{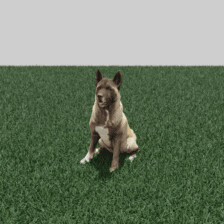}\vspace{-1pt}
    \resultrowBITEAppendix{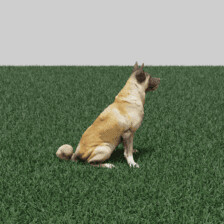}\vspace{-1pt}
    \resultrowBITEAppendix{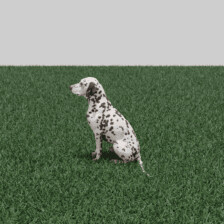}\vspace{-1pt}
    \resultrowBITEAppendix{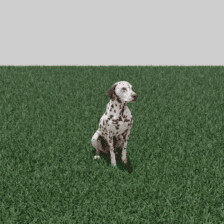}\vspace{-1pt}
    \resultrowBITEAppendix{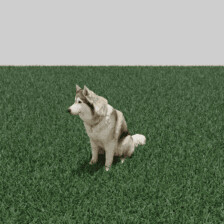}\vspace{-1pt}
    \resultrowBITEAppendix{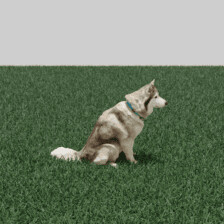}\vspace{-3pt}
    \resultrowBITEAppendix{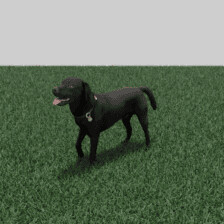}\vspace{-5mm}
    \resultrowBITEAppendix{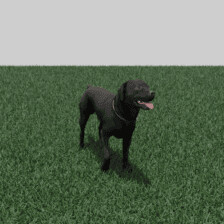}\vspace{-5mm}
    \resultrowBITEAppendix{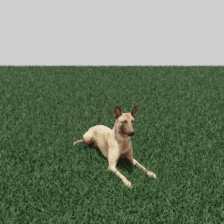}\vspace{-5mm}
    \resultrowBITEAppendix{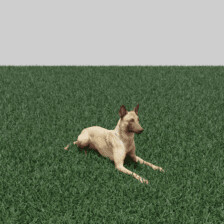}

    \caption{Qualitative comparison of monocular 3D dog reconstruction methods. Each column shows the reconstruction overlaid on the input image produced by a different approach. From left to right, WLDO~\cite{WLDO}, BARC~\cite{BARC}, BITE~\cite{BITE}, GenZoo~\cite{GenZoo}, AniMer~\cite{AniMer}, AniMer+~\cite{AniMerPlus}, and ours. The input image is shown in the first column. Images are cropped around the dog in the centre before being passed into each regression architecture. Images sourced from the set of synthetic dog renders from BITE~\cite{BITE}. We provide results on two views for each of the presented dogs, where the dog's pose is identical but viewed from a different viewpoint.}
    \label{figure:regression_results_overview_BITE}
\end{figure*}

\begin{figure*}
    \centering

    \newcommand{\resultrowStanfordExtraAppendix}[1]{%
        \includegraphics[width=0.125\textwidth]{supplementary/figures/AMR_regression_StanfordExtra/reference_images/full_input_#1.jpg}%
        \includegraphics[width=0.125\textwidth]{supplementary/figures/AMR_regression_StanfordExtra/WLDO/full_overlay_#1.jpg}%
        \includegraphics[width=0.125\textwidth]{supplementary/figures/AMR_regression_StanfordExtra/BARC/full_overlay_#1.jpg}%
        \includegraphics[width=0.125\textwidth]{supplementary/figures/AMR_regression_StanfordExtra/BITE/full_overlay_#1.jpg}%
        \includegraphics[width=0.125\textwidth]{supplementary/figures/AMR_regression_StanfordExtra/GenZoo/full_overlay_#1.jpg}%
        \includegraphics[width=0.125\textwidth]{supplementary/figures/AMR_regression_StanfordExtra/AniMer/full_overlay_#1.jpg}%
        \includegraphics[width=0.125\textwidth]{supplementary/figures/AMR_regression_StanfordExtra/AniMerPlus/full_overlay_#1.jpg}%
        \includegraphics[width=0.125\textwidth]{supplementary/figures/AMR_regression_StanfordExtra/PAWS/full_overlay_#1.jpg}%
    }

    \begin{minipage}{0.125\linewidth}\centering\textbf{Image}\end{minipage}%
    \begin{minipage}{0.125\linewidth}\centering\textbf{WLDO}\end{minipage}%
    \begin{minipage}{0.125\linewidth}\centering\textbf{BARC}\end{minipage}%
    \begin{minipage}{0.125\linewidth}\centering\textbf{BITE}\end{minipage}%
    \begin{minipage}{0.125\linewidth}\centering\textbf{GenZoo}\end{minipage}%
    \begin{minipage}{0.125\linewidth}\centering\textbf{AniMer}\end{minipage}%
    \begin{minipage}{0.125\linewidth}\centering\textbf{AniMer+}\end{minipage}%
    \begin{minipage}{0.125\linewidth}\centering\textbf{Ours}\end{minipage}%

    \resultrowStanfordExtraAppendix{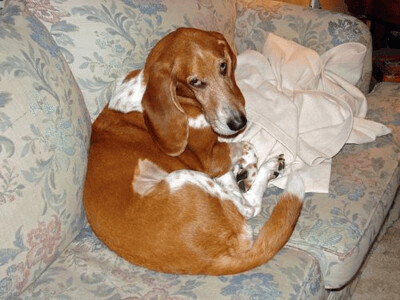}\vspace{-1pt}
    \resultrowStanfordExtraAppendix{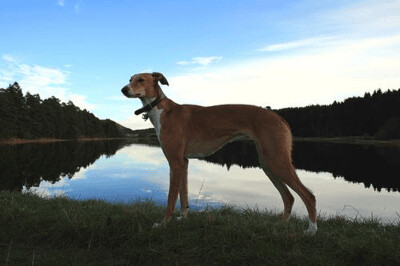}\vspace{-1pt}
    \resultrowStanfordExtraAppendix{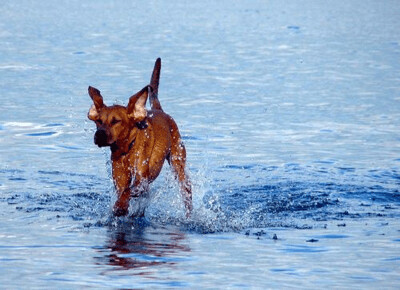}\vspace{-1pt}
    \resultrowStanfordExtraAppendix{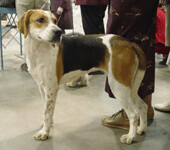}\vspace{-1pt}
    \resultrowStanfordExtraAppendix{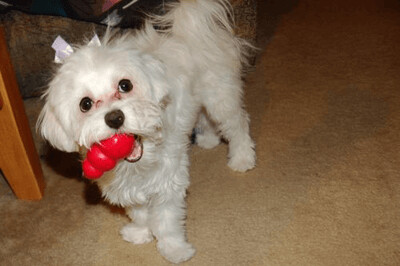}\vspace{-1pt}
    \resultrowStanfordExtraAppendix{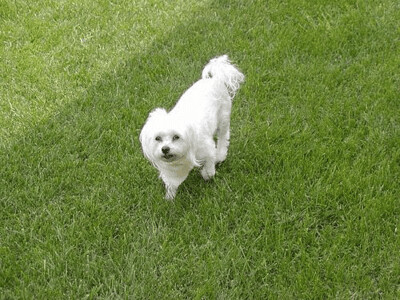}\vspace{-1pt}
    \resultrowStanfordExtraAppendix{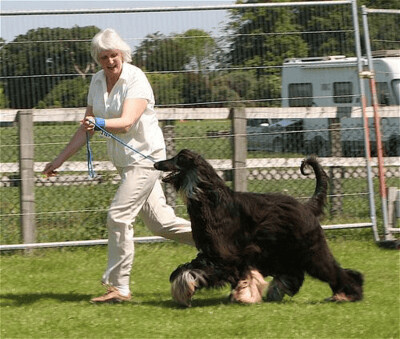}\vspace{-1pt}
    \resultrowStanfordExtraAppendix{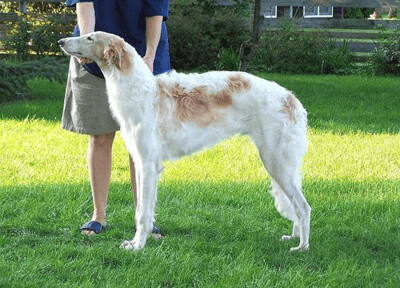}\vspace{-1pt}
    \resultrowStanfordExtraAppendix{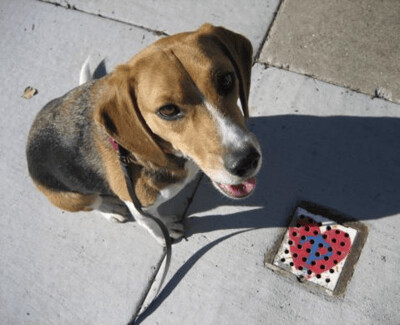}\vspace{-1pt}
    \resultrowStanfordExtraAppendix{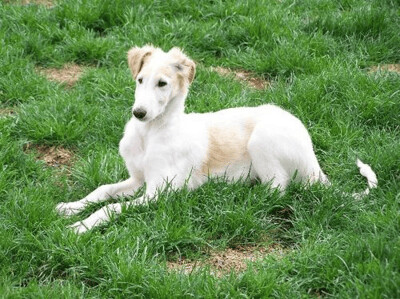}\vspace{-1pt}
    \resultrowStanfordExtraAppendix{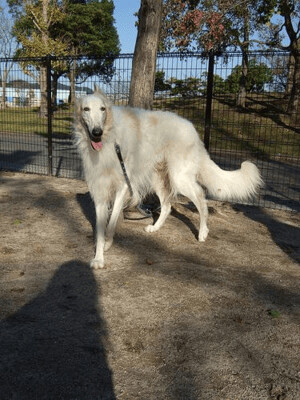}

    \caption{Qualitative comparison of monocular 3D dog reconstruction methods. Each column shows the reconstruction overlaid on the input image produced by a different approach. From left to right, WLDO~\cite{WLDO}, BARC~\cite{BARC}, BITE~\cite{BITE}, GenZoo~\cite{GenZoo}, AniMer~\cite{AniMer}, AniMer+~\cite{AniMerPlus}, and ours. The input image is shown in the first column. All methods regress from a bounding-box crop around the dog with the reconstruction transformed onto the full image. Images sourced from StanfordExtra~\cite{WLDO}.}
    \label{figure:regression_results_overview_StanfordExtra}
\end{figure*}

\begin{figure*}
    \centering

    \newcommand{\resultrowAn}[1]{%
        \includegraphics[width=0.125\textwidth]{supplementary/figures/AMR_regression_Animal3D/reference_images/full_input_#1.jpg}%
        \includegraphics[width=0.125\textwidth]{supplementary/figures/AMR_regression_Animal3D/WLDO/full_overlay_#1.jpg}%
        \includegraphics[width=0.125\textwidth]{supplementary/figures/AMR_regression_Animal3D/BARC/full_overlay_#1.jpg}%
        \includegraphics[width=0.125\textwidth]{supplementary/figures/AMR_regression_Animal3D/BITE/full_overlay_#1.jpg}%
        \includegraphics[width=0.125\textwidth]{supplementary/figures/AMR_regression_Animal3D/GenZoo/full_overlay_#1.jpg}%
        \includegraphics[width=0.125\textwidth]{supplementary/figures/AMR_regression_Animal3D/AniMer/full_overlay_#1.jpg}%
        \includegraphics[width=0.125\textwidth]{supplementary/figures/AMR_regression_Animal3D/AniMerPlus/full_overlay_#1.jpg}%
        \includegraphics[width=0.125\textwidth]{supplementary/figures/AMR_regression_Animal3D/PAWS/full_overlay_#1.jpg}%
    }

    \begin{minipage}{0.125\linewidth}\centering\textbf{Image}\end{minipage}%
    \begin{minipage}{0.125\linewidth}\centering\textbf{WLDO}\end{minipage}%
    \begin{minipage}{0.125\linewidth}\centering\textbf{BARC}\end{minipage}%
    \begin{minipage}{0.125\linewidth}\centering\textbf{BITE}\end{minipage}%
    \begin{minipage}{0.125\linewidth}\centering\textbf{GenZoo}\end{minipage}%
    \begin{minipage}{0.125\linewidth}\centering\textbf{AniMer}\end{minipage}%
    \begin{minipage}{0.125\linewidth}\centering\textbf{AniMer+}\end{minipage}%
    \begin{minipage}{0.125\linewidth}\centering\textbf{Ours}\end{minipage}%

    \resultrowAn{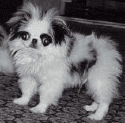}\vspace{-1pt}
    \resultrowAn{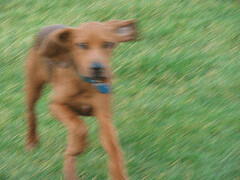}\vspace{-1pt}
    \resultrowAn{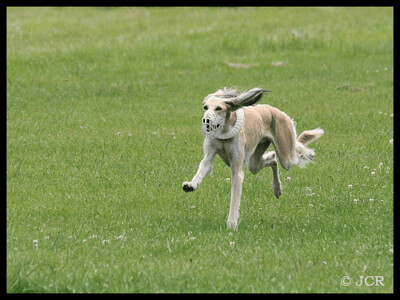}\vspace{-1pt}
    \resultrowAn{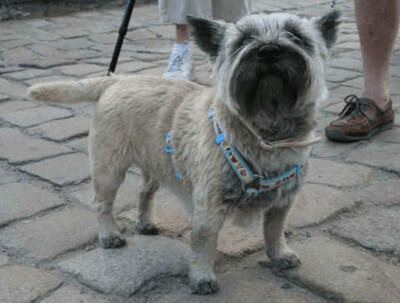}\vspace{-1pt}
    \resultrowAn{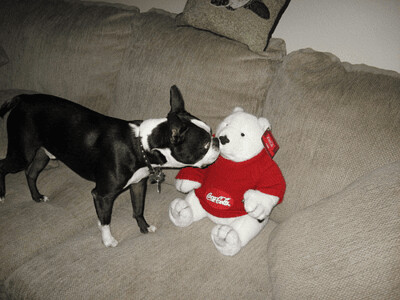}\vspace{-1pt}
    \resultrowAn{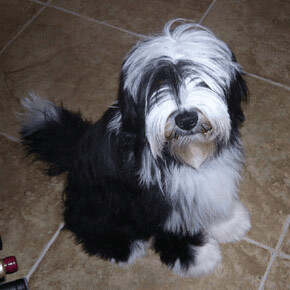}\vspace{-1pt}
    \resultrowAn{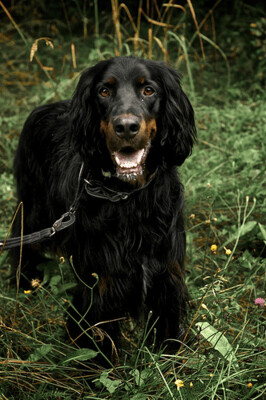}\vspace{-3pt}
    \resultrowAn{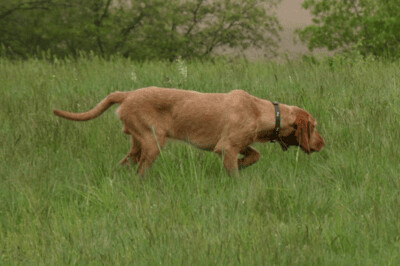}\vspace{-5mm}
    \resultrowAn{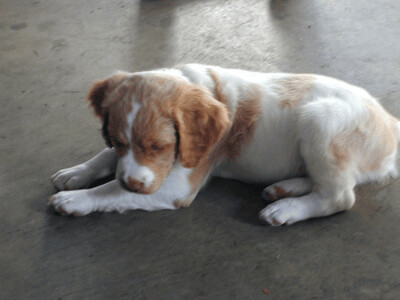}

    \caption{Qualitative comparison of monocular 3D dog reconstruction methods. Each column shows the reconstruction overlaid on the input image produced by a different approach. From left to right, WLDO~\cite{WLDO}, BARC~\cite{BARC}, BITE~\cite{BITE}, GenZoo~\cite{GenZoo}, AniMer~\cite{AniMer}, AniMer+~\cite{AniMerPlus}, and ours. The input image is shown in the first column. All methods regress from a bounding-box crop around the dog with the reconstruction transformed onto the full image. Images sourced from Animal3D~\cite{Animal3D}.}
    \label{figure:regression_results_overview_Animal3D}
\end{figure*}

To extend our quantitative evaluation, we provide additional qualitative results from our learned regression architecture in \cref{figure:regression_results_overview_BITE,figure:regression_results_overview_StanfordExtra,figure:regression_results_overview_Animal3D}, evaluated on BITE~\cite{BITE}, StanfordExtra~\cite{WLDO}, and Animal3D~\cite{Animal3D}, respectively.
For each test dataset we show the input image next to the recovered posed and shaped mesh overlaid on the image.
In our evaluation we compare qualitatively against WLDO~\cite{WLDO}, BARC~\cite{BARC}, BITE~\cite{BITE}, GenZoo~\cite{GenZoo}, AniMer~\cite{AniMer}, AniMer+~\cite{AniMerPlus}, and ours.

\cref{figure:regression_results_overview_BITE} highlights the robustness of our pose reconstruction from single-view images under difficult self-occlusions and challenging poses.
For each synthetically rendered dog from the dataset introduced by R\"uegg et al.~\cite{BITE}, we present reconstructions of the same pose from two different views. 
Overall, our model robustly reconstructs the pose faithfully from different observations, exhibiting a high pose similarity between different views. 
Next, we present extended reconstruction results on two real datasets, StanfordExtra~\cite{WLDO} and Animal3D~\cite{Animal3D}, in \cref{figure:regression_results_overview_StanfordExtra} and \cref{figure:regression_results_overview_Animal3D}, respectively.
In addition to the reconstructions on the synthetic dataset, our model is able to accurately reconstruct the observed pose on real-world datasets, even under severe occlusions, such as when only a part of the dog is observed.


\end{document}